\documentclass[10pt,twocolumn,letterpaper]{article}

\usepackage[pagenumbers]{cvpr} % To force page numbers, e.g. for an arXiv version

\usepackage[dvipsnames]{xcolor}

\usepackage{xcolor}
\usepackage{adjustbox}
\definecolor{bg}{RGB}{255,249,227}

\definecolor{cvprblue}{rgb}{0.21,0.49,0.74}
\usepackage[pagebackref,breaklinks,colorlinks,citecolor=cvprblue]{hyperref}
\usepackage{booktabs}
\usepackage{multirow}
\usepackage{makecell}
\usepackage{comment}
\usepackage{colortbl}
\usepackage[accsupp]{axessibility} % Improves PDF readability for those with visual impairments.

\newcommand{\Tref}[1]{Table~\ref{#1}}

\newcommand{\fref}[1]{Fig.~\ref{#1}}

\definecolor{firstcolor}{RGB}{199, 85, 93}
\definecolor{secondcolor}{RGB}{63, 217, 167}
\definecolor{thirdcolor}{RGB}{39, 87, 146}

\definecolor{firstcolor}{RGB}{252, 235, 193}%{236 168 169}
\definecolor{secondcolor}{RGB}{241, 233, 223} %{211 226 183}
\definecolor{thirdcolor}{RGB}{219, 200, 189} %{116 174 212}

\definecolor{thirdcolor}{RGB}{255, 255, 255}

\newcommand{\markfirst}[1]{\cellcolor{firstcolor}{\textbf{#1}}}
\newcommand{\marksecond}[1]{\cellcolor{secondcolor}{#1}}
\newcommand{\markthird}[1]{\cellcolor{thirdcolor}{#1}}
\newcommand{\colorfirsttext}[1]{\colorbox{firstcolor}{\textbf{#1}}}
\newcommand{\colorsecondtext}[1]{\colorbox{secondcolor}{#1}}

\def\paperID{8906} % *** Enter the Paper ID here
\def\confName{CVPR}
\def\confYear{2024}

\title{Photo-SLAM: Real-time Simultaneous Localization and Photorealistic Mapping for Monocular, Stereo, and RGB-D Cameras}

\author{Huajian Huang$^1$ \quad Longwei Li$^2$ \quad Hui Cheng$^2$ \quad Sai-Kit Yeung$^1$\\
$^1$The Hong Kong University of Science and Technology \quad
$^2$Sun Yat-sen University \\
{\tt\small hhuangbg@connect.ust.hk},
{\tt\small lilw23@mail2.sysu.edu.cn},
{\tt\small chengh9@mail.sysu.edu.cn},
{\tt\small saikit@ust.hk}
}

\begin{document}
\maketitle
\begin{abstract}
The integration of neural rendering and the SLAM system recently showed promising results in joint localization and photorealistic view reconstruction. However, existing methods, fully relying on implicit representations, are so resource-hungry that they cannot run on portable devices, which deviates from the original intention of SLAM. In this paper, we present Photo-SLAM, a novel SLAM framework with a hyper primitives map. Specifically, we simultaneously exploit explicit geometric features for localization and learn implicit photometric features to represent the texture information of the observed environment. In addition to actively densifying hyper primitives based on geometric features, we further introduce a Gaussian-Pyramid-based training method to progressively learn multi-level features, enhancing photorealistic mapping performance. The extensive experiments with monocular, stereo, and RGB-D datasets prove that our proposed system Photo-SLAM significantly outperforms current state-of-the-art SLAM systems for online photorealistic mapping, e.g., PSNR is 30\% higher and rendering speed is hundreds of times faster in the Replica dataset. Moreover, the Photo-SLAM can run at real-time speed using an embedded platform such as Jetson AGX Orin, showing the potential of robotics applications. Project Page and code: \url{https://huajianup.github.io/research/Photo-SLAM/}.

% achieves state-of-the-art performance. in terms of localization efficiency, photorealistic mapping quality, and rendering speed. 
%The geometry mapping is optimized by minimizing reprojection residuals while the corresponding implicit features are learned by minimizing photometric loss. 
%As the system is free of depth supervision, it can support monocular, stereo, and RGB-D cameras. 
%\vspace{-0.2cm}
\end{abstract}    
\section{Introduction}\label{sec:intro}

%------------------------------------------------------------------------
\begin{figure}
    \centering
    \includegraphics[width=\linewidth]{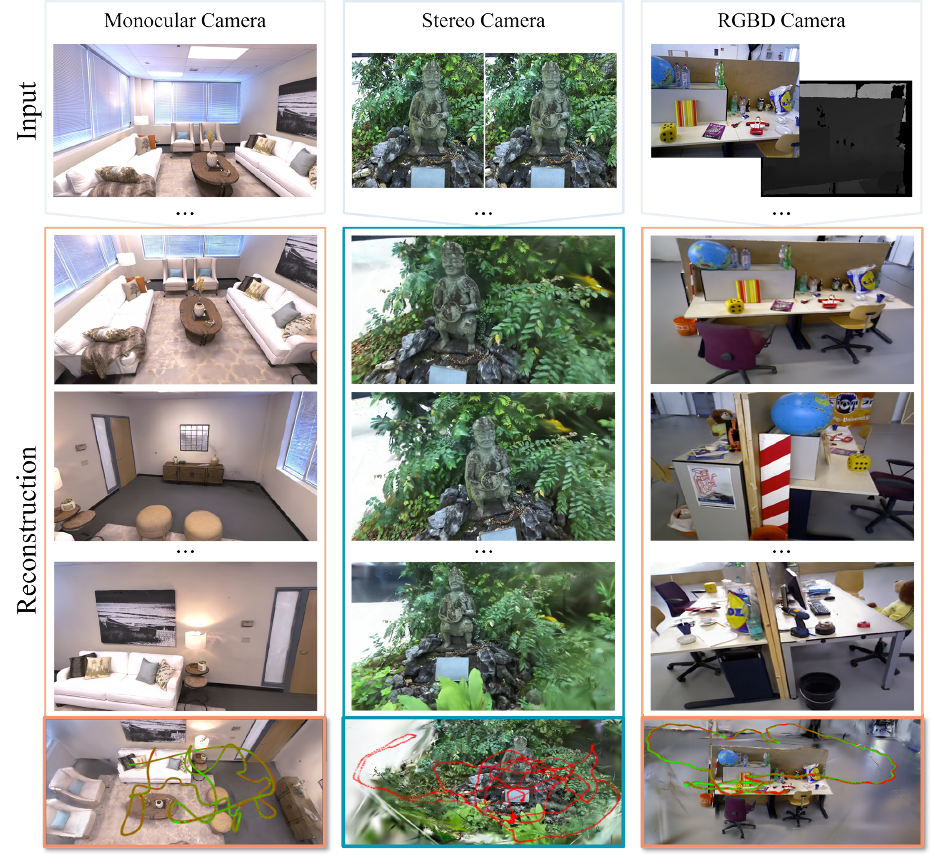}
    \vskip -0.1cm
    \caption{Rendering and trajectory results. Photo-SLAM can reconstruct high-fidelity views of scenes using monocular, stereo, and RGB-D cameras while render speed is up to 1000 FPS.} % at high speed,  for simultaneous localization and photorealistic mapping in real-time supporting monocular, stereo, and RGB-D cameras. It
    \label{fig:teaser}
    %\vskip -0.3cm
\end{figure}

Simultaneous Localization and Mapping (SLAM) using cameras is a fundamental problem in both computer vision and robotics, seeking to enable autonomous systems to navigate and comprehend their surroundings. 
Traditional SLAM systems~\cite{mur2015orb,svo,lsd-slam,dso} primarily focus on geometric mapping, providing accurate but visually simplistic representations of the environment. However, recent developments in neural rendering~\cite{advancesNueralRendering, neuralrendering} have demonstrated the potential of integrating photorealistic view reconstruction into the SLAM pipeline, enhancing the perception capabilities of robotic systems.

Despite the promising results achieved through the integration of neural rendering and SLAM, existing methods simply and heavily rely on implicit representations, making them computationally intensive and unsuitable for deployment on resource-constrained devices. For example, Nice-SLAM~\cite{niceslam} leverages a hierarchical grid~\cite{plenoctrees} to store learnable features representing the environment while ESLAM~\cite{eslam} utilizes multi-scale compact tensor components~\cite{TensoRF}. They then jointly estimate the camera poses and optimize features by minimizing the reconstruction loss of a batch of ray sampling~\cite{nerf}. Such an optimization process is time-consuming. 
Consequently, it is indispensable for them to incorporate corresponding depth information
obtained from various sources such as RGB-D cameras, dense optical flow estimators~\cite{raft}, or monocular depth estimators~\cite{monodepth} to ensure efficient convergence.
Additionally, since the implicit features are decoded by the multi-layer perceptrons (MLPs),
it is typically necessary to carefully define a bounding area to normalize ray sampling for optimal performance, as discussed in~\cite{360roam}. 
It essentially limits the scalability of the system. These limitations imply that they cannot provide real-time exploration and mapping capabilities in the unknown environment using portable platforms, which is one of the main objectives of SLAM.

%Consequently, corresponding depth information, obtained from RGB-D cameras, dense optical flow~\cite{raft}, or monocular depth estimators~\cite{monodepth}, is indispensable to ensure prompt convergence. In addition, since the implicit features are decoded by the multi-layer perceptron (MLP), a well-defined bounding area is generally necessary to normalize ray sampling for optimal performance~\cite{360roam}.  
% stand in contrast to the 
%as discussed in 
%and also violated the original
%In order to alleviate catastrophic forgetting of neural networks, 
%This limitation stands in contrast to the original objective of SLAM, which aims to provide real-time navigation and mapping capabilities on portable platforms.

%In this paper, we propose an innovative SLAM framework, denoted as Photo-SLAM, which addresses the scalability and computational resource constraints associated with existing methods while simultaneously achieving precise localization and photorealistic mapping. 
In this paper, we propose Photo-SLAM, an innovative framework that addresses the scalability and computational resource constraints of existing methods, while achieving precise localization and online photorealistic mapping.
We maintain a hyper primitives map which is composed of point clouds storing ORB features~\cite{ORB}, rotation, scaling, density, and spherical harmonic (SH) coefficients~\cite{nex, fridovich2022plenoxels}. 
The hyper primitives map allows the system to efficiently optimize tracking using a factor graph solver and learn the corresponding mapping by backpropagating the loss between the original images and rendering images. The images are rendered by 3D Gaussian splatting~\cite{kerbl20233dgaussiansplatting} rather than ray sampling. %a sophisticated 
Although the introduction of a 3D Gaussian splatting renderer can reduce view reconstruction costs, it does not enable the generation of high-fidelity rendering for online incremental mapping, in particular in monocular scenarios. 
To achieve high-quality mapping without reliance on dense depth information, we further propose a geometry-based densification strategy and a Gaussian-Pyramid-based (GP) learning method. Importantly, GP learning facilitates the progressive acquisition of multi-level features which effectively enhances the mapping performance of our system.

To evaluate the efficacy of our proposed approach, we conduct extensive experiments employing diverse datasets captured by monocular, stereo, and RGB-D cameras. These experiment results unequivocally demonstrate that Photo-SLAM attains state-of-the-art performance in terms of localization efficiency, photorealistic mapping quality, and rendering speed. Furthermore, the real-time execution of the Photo-SLAM system on the embedded devices showcases its potential for practical robotics applications.  The schematic overview of Photo-SLAM is demonstrated in \fref{fig:teaser} and \fref{fig:overview}.

%Jetson AGX Orin developer toolkit 
%thereby enabling the generation of high-fidelity photorealistic reconstructions.

In summary, the main contributions of this work include:
\begin{itemize}%[topsep=0pt, itemsep=0pt, leftmargin=*]
    %\item We introduced a temporal and spatial co-visibility graph to enable correcting the drift of hybrid primitive maps, which is essential for incremental learning in large-scale scenes.
    \item We developed the first simultaneous localization and photorealistic mapping system based on hyper primitives map. The novel framework supports monocular, stereo, and RGB-D cameras in indoor and outdoor environments.
    \item We proposed Gaussian-Pyramid-based learning allowing the model to efficiently and effectively learn multi-level features realizing high-fidelity mapping.
    \item The system, fully implemented in C++ and CUDA, achieves start-of-the-art performance and can run at real-time speed even on embedded platforms. 
    %The code will be publicly available.
\end{itemize}

\section{Related Work}\label{sec:relatedwork}

%-------------------------------------------------------------------------
\begin{figure}
    \centering
    \subfloat[Taxonomy]{\label{fig:taxonomy}\includegraphics[width=0.55\linewidth]{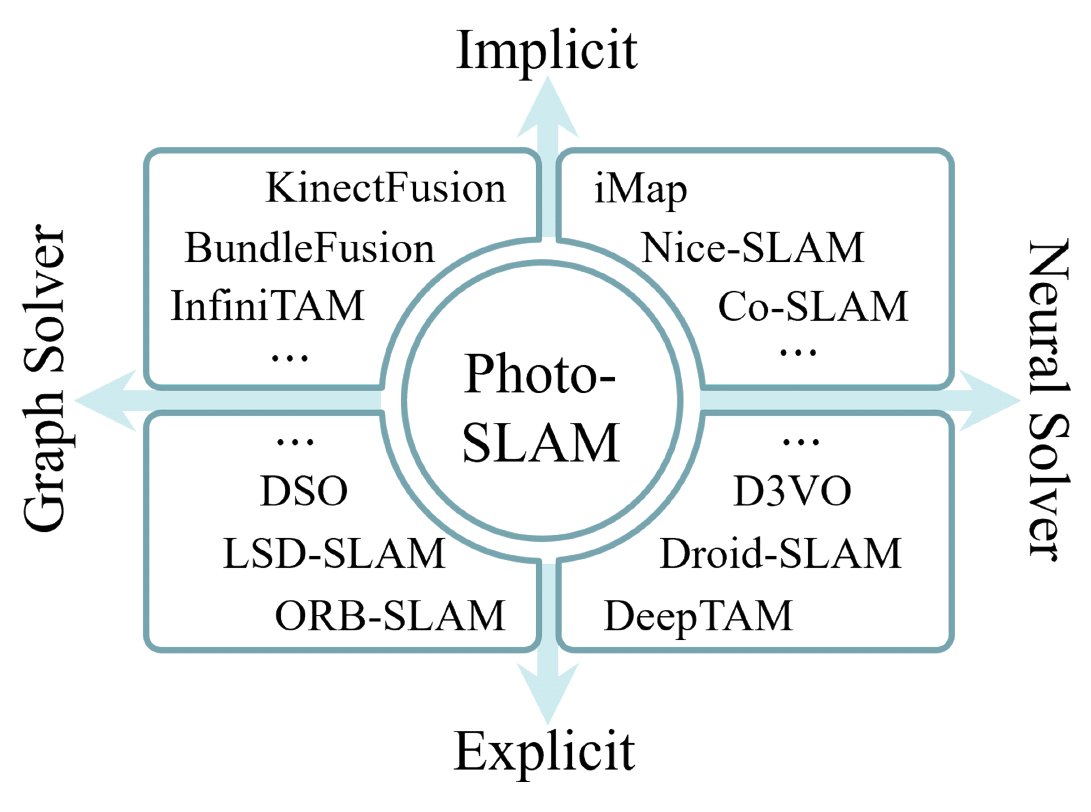}}\,
    \subfloat[Overview of Photo-SLAM]{\label{fig:overview}\includegraphics[width=0.42\linewidth]{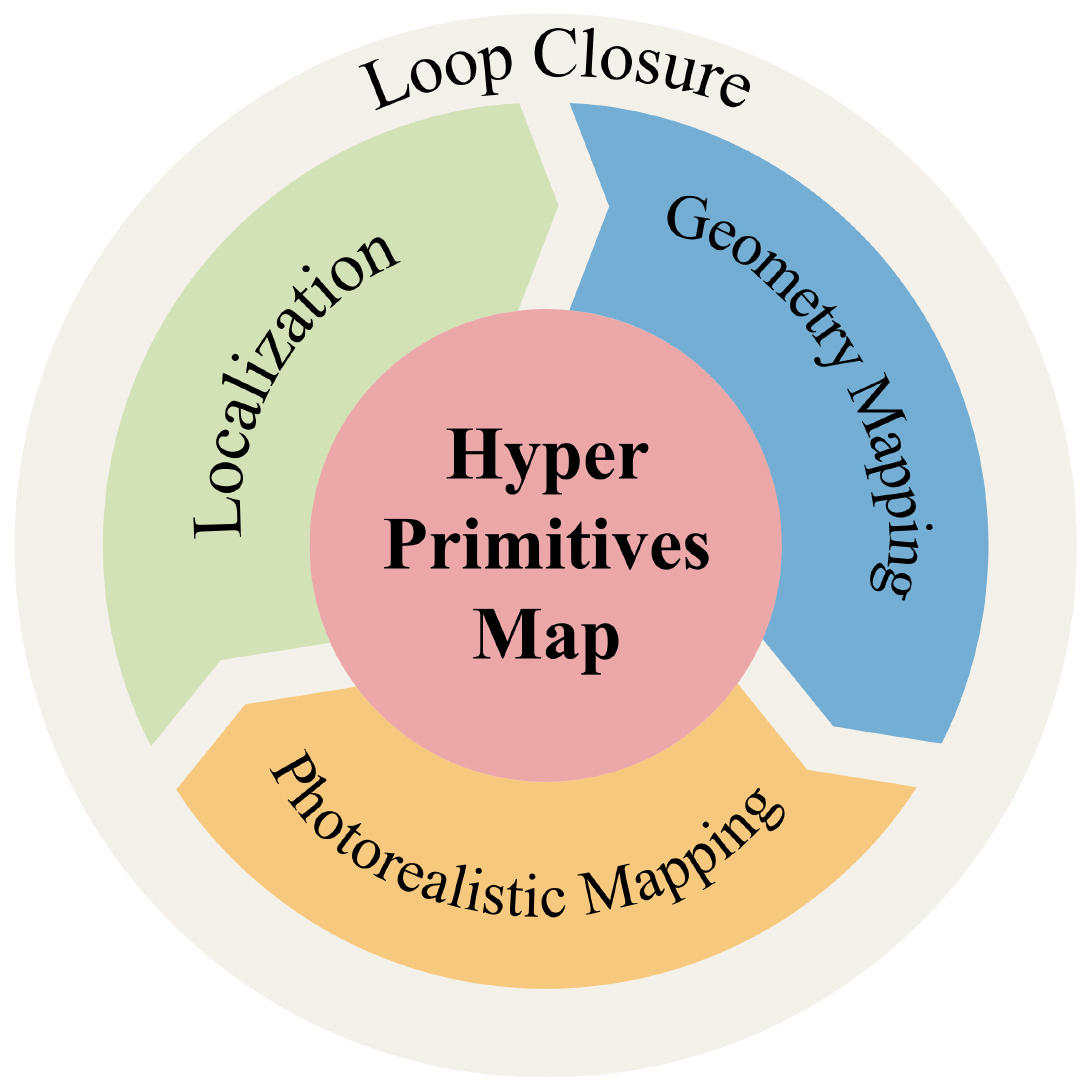}}
    \vskip -0.2cm
    \caption{The Photo-SLAM contains four main components, including localization, explicit geometry mapping, implicit photorealistic mapping, and loop closure components, while maintaining a map with hyper primitives. }
    \label{fig:slams}
    \vskip -0.2cm
    %\vskip -0.4cm schematic overview
\end{figure}

%\subsection{Visual Localization and Mapping}
Visual localization and mapping is a problem that aims to build a proper representation of an unknown environment via cameras while estimating their poses within that environment. In contrast to SfM techniques, visual SLAM techniques typically pursue a better trade-off between accuracy and real-time performance. In this section, we focus on visual SLAM and conduct a brief review. %that is mostly relevant to our work  denoted as Structure-from-motion (SFM) and visual SLAM. 

\noindent\textbf{Graph Solver vs Neural Solver}.
Classical SLAM methods widely adopt factor graphs to model complex optimization problems between variables (i.e., poses and landmarks) and measurements (i.e., observations and constraints). To achieve real-time performance, SLAM methods incrementally propagate their pose estimations while avoiding expensive operations. For example, ORB-SLAM series methods ~\cite{mur2015orb, mur2017orb2, campos2021orb3} rely on extracting and tracking lightweight geometric features across consecutive frames, which perform bundle adjustment locally instead of globally. Moreover, direct SLAMs like LSD-SLAM~\cite{lsd-slam} and DSO~\cite{dso} operate on raw image intensities, without the cost of geometric feature extractions. They maintain a sparse or semi-dense map represented by point clouds online, even on the resource-constraint system. Benefiting from the success of deep-learning models, learnable parameters and models are introduced into SLAM making the pipeline differentiable. Some methods such as DeepTAM~\cite{deeptam} predict camera poses by the neural network~\cite{posenet} end-to-end, while the accuracy is limited. 
To enhance performance, some methods, e.g., D3VO~\cite{d3vo} and Droid-SLAM ~\cite{droid}, introduce monocular depth estimation~\cite{monodepth} or dense optical flow estimation~\cite{raft} models into the SLAM pipeline as supervision signals. Therefore, they can generate depth maps that explicitly represent the scene geometry. With the large-scale synthetic SLAM dataset, TartanAir~\cite{tartanair}, available for training, Droid-SLAM building upon RAFT~\cite{raft} achieves state-of-the-art performance. However, the pure neural-based solver is computationally expensive and their performance would significantly degrade on the unseen scenes.
%. However, these learning-based methods lack generality and . With the large-scale synthetic SLAM dataset available for training, TartanAir~\cite{}, Droid-SLAM building up RAFT~\cite{} achieve the state-of-art performance. However,

\noindent\textbf{Explicit Representation vs Implicit Representation}.
In order to obtain dense reconstruction, some methods including KinectFusion~\cite{kinectfusion}, BundleFusion~\cite{bundlefusion}, and InfiniTAM~\cite{infinitam} utilize the implicit representation, Truncated Signed Distance Function (TSDF)~\cite{tsdf}, to integrate the incoming RGB-D images and reconstruct a continuous surface, which can run in real time on GPU. Although they can obtain dense reconstruction, view rendering quality is limited. Recently, neural rendering techniques represented by neural radiance field (NeRF)~\cite{nerf} have achieved breathtaking novel view synthesis.
Given camera poses, NeRF implicitly models the scene geometry and color by multi-layer perceptrons (MLP). The MLP is optimized by minimizing the loss of rendering images and training views. iMAP~\cite{imap} then adapts NeRF for incremental mapping, optimizing not only MLP but also camera poses. The following work Nice-SLAM~\cite{niceslam} introduces multi-resolution grids~\cite{plenoctrees} to store features reducing the cost of deep MLP query. Co-SLAM~\cite{coslam} and ESLAM~\cite{eslam} explore Instant-NGP~\cite{muller2022instant} and TensoRF~\cite{TensoRF} respectively to further accelerate the mapping speed. However, implicitly joint optimization of camera poses and geometry representation is still ill-conditioned. Inevitably, they rely on explicit depth information from RGB-D cameras or additional model predictions for fast convergence of the radiance field.  %by volume rendering techniques is

Our proposed Photo-SLAM seeks to recover a concise representation of the observed environment for immersive exploration rather than reconstructing a dense mesh.
%We take advantage of explicit geometric features for efficient localization while learning implicit color features for photorealistic mapping.  
It maintains a map with hyper primitives online which capitalizes on explicit geometric feature points for accurate and efficient localization while leveraging implicit representations to capture and model the texture information. Please refer to \fref{fig:taxonomy} for the taxonomy of existing systems.
%The optimization of the Photo-SLAM framework encompasses a two-step process. Initially, we minimize reprojection residuals to refine the sparse geometry map, ensuring precise localization. Subsequently, we learn and densify the corresponding implicit features by minimizing photometric loss. The photorealistic views are rendered by feature splatting~\cite{kerbl20233dgaussiansplatting} instead of ray sampling.
Since Photo-SLAM achieves high-quality mapping without reliance on dense depth information, it can support RGB-D cameras as well as monocular and stereo cameras. 
\section{Photo-SLAM}\label{sec:photoslam}

% \begin{figure}
%     \centering
%     \includegraphics[width=0.6\linewidth]{figure/system.png}
%     %\vskip -0.1cm
%     \caption{The Photo-SLAM contains four main components, including localization, geometry mapping, neural feature mapping, and loop closure components, while maintaining a map with hyper primitives.}
%     \label{fig:system}
%     %\vskip -0.4cm
% \end{figure}

    %The main components of our method is shown in Fig.~\ref{fig:slams}. Our proposed Photo-SLAM system takes a set of sequential frames as input, which can be denoted as ${\{I_i\}}^{M}_{i=1}$ for monocular, ${\{I^L_i, I^R_i\}}^{M}_{i=1}$ for stereo, and ${\{I_i, D_i\}}^{M}_{i=1}$ for RGB-D sensor type, where $M$ denotes the total number of images in sequence. In the geometric localization and mapping component described in Sec.~\ref{subsec:photoslam-localization}, we estimate Six-Degree-of-Freedom (6-DoF) camera poses $\{R_i|t_i\}^{M}_{i=1}$ and obtain a sparse geometric map $\{P_j\}^N_{j=1}$ consists of $N$ colored 3D points. Then, as described in Sec.~\ref{subsec:photoslam-primitives-mapping}, we continuously send the frame images and poses together with the sparse point cloud to the hyper primitives mapping component to incrementally generate a dense representation of the observed environment. The optimization process, which minimizes both reprojection residuals and photometric loss, is also introduced in detail. Specifically, Sec.~\ref{subsec:photoslam-pyramid} shows our Gaussian-Pyramid-based training strategy for multi-level feature learning, which boosts quality of our result models.
Photo-SLAM contains four main components, including localization, geometry mapping, photorealistic mapping, and loop closure, shown in \fref{fig:overview}. Each component runs in a parallel thread and jointly maintains a hyper primitives map.

\begin{figure*}
    \centering
    \def\imgw{0.182}
    \subfloat[e.g., NSVF~\cite{nsvf}]{\label{fig:pt1}\includegraphics[width=\imgw\linewidth]{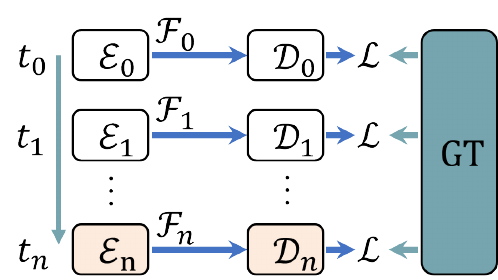}}\,
    \subfloat[e.g., NGLoD~\cite{nglod}]{\label{fig:pt2}\includegraphics[width=\imgw\linewidth]{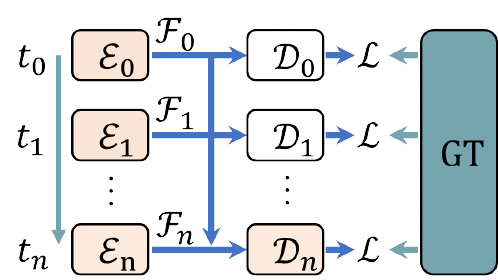}}\,
    \subfloat[e.g., Neuralangelo~\cite{neuralangelo}]{\label{fig:pt3}\includegraphics[width=\imgw\linewidth]{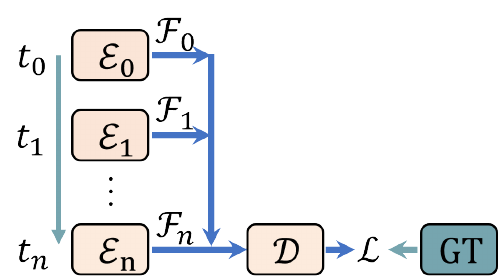}}\,
    \subfloat[e.g., BungeeNeRF~\cite{bungeenerf}]{\label{fig:pt4}\includegraphics[width=\imgw\linewidth]{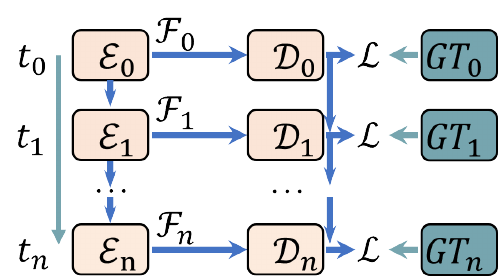}}\,
    \subfloat[Ours]{\label{fig:pt_ours}\includegraphics[width=0.22\linewidth]{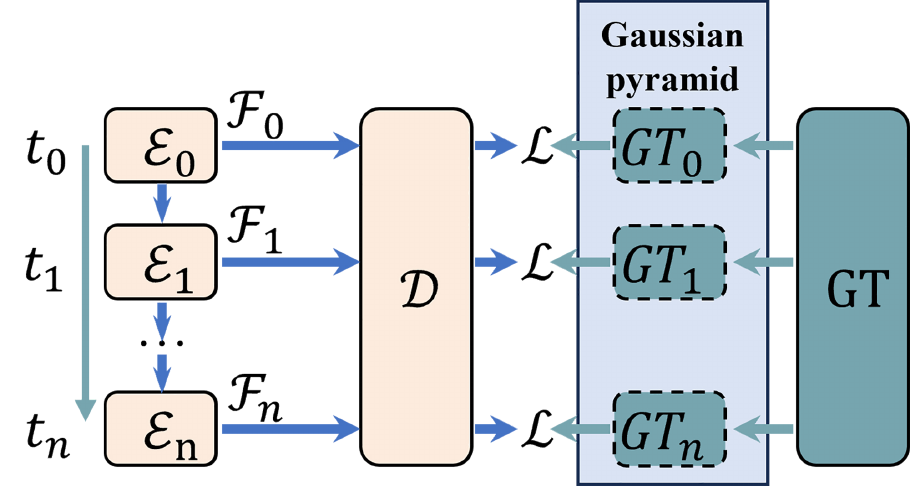}}\\
    \vspace{0.1em}
    \includegraphics[width=0.9\linewidth]{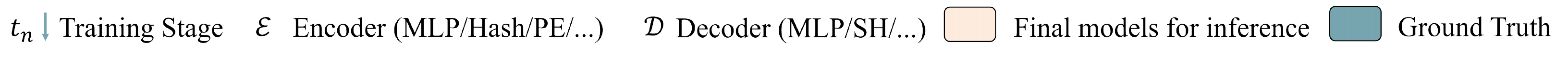}
    \vspace{-0.2cm}
    \caption{Comparison of different progressive training methods. The encoder $\mathcal{E}_n$ here represents a structure to regress features $\mathcal{F}_n$ which can be an MLP, voxel grid, hash table, positional encoding, etc. The decoder $\mathcal{D}_n$ here represents a structure converting $\mathcal{F}_n$ into density, color, or other information. We proposed a new method based on the Gaussian pyramid to efficiently learn multi-level features.}\label{fig:pt}
\end{figure*}

\subsection{Hyper Primitives Map}
In our system, hyper primitives are defined as a set of point clouds $\mathbf{P}\in \mathbb{R}^3$ associated with ORB features~\cite{ORB} $\mathbf{O}\in \mathbb{R}^{256}$, rotation $\mathbf{r}\in SO(3)$, scaling $\mathbf{s}\in \mathbb{R}^3$, density $\sigma \in \mathbb{R}^1$, and spherical harmonic coefficients $\mathbf{SH}\in \mathbb{R}^{16}$. ORB features extracted from image frames take responsibility for establishing 2D-to-2D and 2D-to-3D correspondences. 
%The system starts pose tracking once it successfully estimates the transformation matrix with sufficient 2D-to-2D correspondences between two adjacent frames. Meanwhile, the hyper primitives map will be initialized via triangulation. 
Once the system successfully estimates the transformation matrix based on sufficient 2D-to-2D correspondences between adjacent frames, the hyper primitives map is initialized via triangulation, and pose tracking gets started.
During tracking, the localization component processes the incoming images and makes use of 2D-to-3D correspondence to calculate current camera poses. 
In addition, the geometry mapping component will incrementally create and initialize sparse hyper primitives.  Finally, the photorealistic component progressively optimizes and densifies hyper primitives. %, especially rotation, scaling, density, and spherical harmonic coefficients by minimizing the photometric loss.
%create hyper primitives
%In addition, the geometry mapping component will incrementally create hyper primitives as a coarse stage. Finally, the photorealistic component progressively optimizes hyper primitives, especially normal, scaling, density, and spherical harmonic coefficients by minimizing the photometric loss.

\subsection{Localization and Geometry Mapping}\label{subsec:photoslam-localization}
    The localization and geometry mapping components provide not only efficient 6-DoF camera pose estimations of the input images, but also sparse 3D points. The optimization problem is formulated as a factor graph solved by the Levenberg–Marquardt (LM) algorithm.

    %The simultaneous localization and sparse mapping component of Photo-SLAM is built on the explicit-geometric-feature-based ORB-SLAM3~\cite{campos2021orb3}, which is modified to meet the demand of providing primary data for dense hyper primitives mapping. The system consists of three parallel threads: 1) the tracking thread who finds geometric feature matches between every current frame and the local map then applies motion-only bundle adjustment (BA) in order to localize the camera; 2) the local mapping thread who creates new colored 3D points, manages the local map and optimize it by performing local BA; 3) the loop closing thread who detects large loops and performs pose-graph optimizations to correct accumulated drift.

    %The geometric localization and mapping component provides not only robust and accurate 6-DoF camera poses of the input images, but also a sparse 3D point cloud map. Both of them are needed by the hyper primitives mapping component. To be specific, we obtain keyframes and sparse map points after local BAs and pose-graph optimizaitons, and feed them continuously to the hyper primitives mapping thread. This makes them serve as a incremental training set of our densification process.

%\subsubsection{Geometric Reprojection Residuals}\label{subsubsec:photoslam-opt-reproj}

    %We first perform BA in the geometry localization and mapping component to optimize camera poses and the position of geometry features in world coordinates.

    %We use the g2o~\cite{grisetti2011g2o} implemented Levenberg–Marquardt (LM) method to minimize geometric residuals. 

    In the localization thread, we use a motion-only bundle adjustment to optimize the camera orientation $\mathbf{R}\in SO(3)$ and position $\mathbf{t}\in \mathbb{R}^3$ in order to minimize the reprojection error between matched 2D geometric keypoint $\mathbf{p}_i$ of the frame and 3D point $\mathbf{P}_i$. Let $i\in \mathcal{X}$ be the index of set of matches $\mathcal{X}$, what we are trying to optimize with LM is
    \begin{equation}
        \{\mathbf{R},\mathbf{t}\} = \mathop{\arg\min}\limits_{\mathbf{R},\mathbf{t}}
        \sum_{i\in\mathcal{X}} \rho \left( \lVert \mathbf{p}_i - \pi (\mathbf{RP}_i + \mathbf{t}) \rVert ^2_{\Sigma_g} \right),
    \end{equation}
    where $\Sigma_g$ is the scale-associated covariance matrix of the keypoint, $\pi(\cdot)$ is the 3D-to-2D projection function, and $\rho$ denotes the robust Huber cost function. 

\begin{comment}
    The projection function $\pi_{(\cdot)}$ applied on $\mathbf{P}=[X,Y,Z]^\mathsf{T}$ is different for monocular $(\pi_m)$ and stereo $(\pi_s)$:
    \begin{equation}
        \pi_m(\mathbf{P}) = \left[ 
            \begin{matrix}
                f_x \frac{X}{Z} + c_x \\[2pt]
                f_y \frac{Y}{Z} + c_y
            \end{matrix}
        \right],
        \pi_s(\mathbf{P}) = \left[ 
            \begin{matrix}
                f_x \frac{X}{Z} + c_x \\[2pt]
                f_y \frac{Y}{Z} + c_y \\[2pt]
                f_x \frac{X-b}{Z} + c_x
            \end{matrix}
        \right]
    \end{equation}
    where focal length $(f_x,f_y)$, principal point $(c_x,c_y)$ and baseline $b$ are calibrated intrinsic parameters of the camera.
\end{comment}

    In the geometry mapping thread, we perform a local bundle adjustment on a set of covisible points $\mathcal{P}_L$ and keyframes $\mathcal{K}_L$. The keyframes are selected frames from the input camera sequence and provide good visual information. We construct a factor graph where each keyframe is a node, and the edges represent constraints between keyframes and matched 3D points. We iteratively minimize the reprojection residual by refining the keyframe poses and 3D points using the first-order derivatives of the error function. We fix the poses of keyframes $\mathcal{K}_F$ which are also observing $\mathcal{P}_L$ but not in $\mathcal{K}_L$. Let $\mathcal{K} = \mathcal{K}_L \cup \mathcal{K}_F$, and $\mathcal{X}_k$ be the set of matches between 2D keypoints in a keyframe $k$ and 3D points in $\mathcal{P}_L$. The optimization process aims to reduce the geometric inconsistency between $\mathcal{K}$ and $\mathcal{P}_L$, and is defined as 
    \begin{equation}
        \{ \mathbf{P}_i,\mathbf{R}_l,\mathbf{t}_l | i\in\mathcal{P}_L,l\in\mathcal{K}_L \} =
        \mathop{\arg\min}\limits_{\mathbf{P}_i,\mathbf{R}_l,\mathbf{t}_l} \sum_{k\in\mathcal{K}} \sum_{j\in\mathcal{X}_k} \rho(E(k,j)),
    \end{equation}
    %\begin{equation}
    %    \{ \mathbf{P}_i,\mathbf{R}_l,\mathbf{t}_l | i\in\mathbf{P}_L,l\in\mathcal{K}_L \} =
    %    \mathop{\arg\min}\limits_{\mathbf{P}_i,\mathbf{R}_l,\mathbf{t}_l} \sum_{k\in\mathcal{K}} \sum_{j\in\mathcal{X}_k} \rho(E(k,j)),
    %\end{equation}
    with reprojection residual 
    \begin{equation}\notag
        E(k,j)= \lVert \mathbf{p}_j - \pi (\mathbf{R}_k\mathbf{P}_j +\mathbf{t}_k) \rVert ^2_{\Sigma_g}.
    \end{equation}
    
    %optimizing not only a set of covisible keyframes $\mathcal{K}_L$, but also all points $\mathbf{P}_L$ observed by the keyframes.

\subsection{Photorealisitc Mapping}\label{subsec:photoslam-photorealisitc-mapping}
    The photorealistic mapping thread is responsible for optimizing hyper primitives that are incrementally created by the geometry mapping thread. The hyper primitives can be rasterized by a tile-based renderer to synthesize corresponding images with keyframe poses. The rendering process is formulated as 
    \vspace{-0.3cm}
    \begin{equation}
        C (\mathbf{R, t}) = \sum_{i \in N}\mathbf{c}_i \alpha_i \prod_{j=1}^{i-1} (1-\alpha_i),
    \end{equation}
    where $N$ is the number of hyper primitives, $\mathbf{c}_i$ denotes the color converted from $\mathbf{SH}\in \mathbb{R}^{16}$, and $\alpha_i$ is equal to $\sigma_i \cdot \mathcal{G}(\mathbf{R, t}, \mathbf{P}_i, \mathbf{r}_i, \mathbf{s}_i)$, $\mathcal{G}$ denotes 3D Gaussian splatting algorithm~\cite{kerbl20233dgaussiansplatting}.
    The optimization in terms of position $\mathbf{P}$, rotation $\mathbf{r}$, scaling $\mathbf{s}$, density $\sigma$, and spherical harmonic coefficients $\mathbf{SH}$ is performed by minimizing the photometric loss $\mathcal{L}$ between rendering image ${I_\text{r}}$ and ground truth image $I_\text{gt}$, denoted as
    %The $\mathcal{L}_1$ loss between images $I_1$ and $I_2$ is defined as:
    \begin{equation}\label{equ:loss}
        \mathcal{L} = (1-\lambda)\left|{ I_\text{r} - I_\text{gt} }\right|_1 + \lambda{(1-\text{SSIM}(I_\text{r}, I_\text{gt}))},
    \end{equation} 
    where $\text{SSIM}(I_\text{r}, I_\text{gt})$ denotes structural similarity between two images and $\lambda$ is a weight factor for balance.

\subsubsection{Geometry-based Densification}\label{subsubsec:photoslam-opt-densification}
    If we consider photorealistic mapping as a regression model of the scene, denser hyper primitives, i.e., more parameters, generally can better model the complexity of the scene for higher rendering quality. To meet the demand for real-time mapping, the geometry mapping component only establishes sparse hyper primitives.
    %the localization and geometry mapping component only establish a sparse correspondence between 3D and 2D geometric feature points of frames. 
    Therefore, the coarse hyper primitives created by the geometry mapping need to be densified during the optimization of photorealistic mapping. Apart from splitting or cloning hyper primitives with large loss gradients similar to~\cite{kerbl20233dgaussiansplatting}, we introduce an additional geometry-based densification strategy. 
   
\begin{figure}
        \centering
        \def\imw{0.85}
        \includegraphics[width=\imw\linewidth]{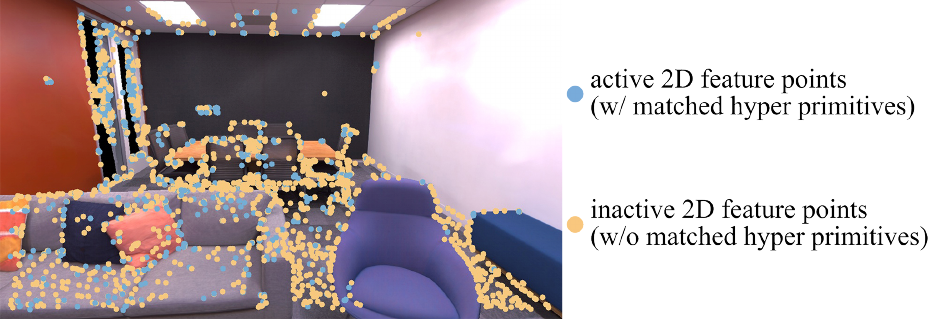}
        %\subfloat[]{\includegraphics[width=\imw\linewidth]{figure/GD/1725.jpg}}
        \vspace{-0.1cm}
        \caption{We make use of initial geometric information to densify hyper primitives.}
        \vspace{-0.1cm}
        
        \label{fig:GD}
    \end{figure}

    Experimentally, less than $30\%$ of 2D geometric feature points of frames are active and have corresponding 3D points, especially for non-RGB-D scenarios, as shown in \fref{fig:GD}. 
    We argue that 2D geometric feature points spatially distributed in the frames essentially represent the region with a complex texture that requires more hyper primitives.
    Therefore, we actively create additional temporary hyper primitives based on the inactive 2D feature points once the keyframe is created for photorealistic mapping. When we use RGB-D cameras, we can directly project the inactive 2D feature points with depth to create temporary hyper primitives. As for monocular scenarios, we estimate the depth of inactive 2D feature points by interpreting the depth of their nearest neighborhood's active 2D feature points. In stereo scenarios, we rely on a stereo-matching algorithm to estimate the depth of inactive 2D feature points.

\subsection{Gaussian-Pyramid-Based Learning}\label{subsec:photoslam-pyramid}
     %of photorealistic mapping, we propose Gaussian-Pyramid-based (GP) learning, a new progressive training method.
    Progressive training is a widely used technology in neural rendering to accelerate the optimization process. Some methods have been proposed to reduce training time while achieving better rendering quality. A basic method is to progressively increase the structure resolution and the number of model parameters. For example, NSVF~\cite{nsvf} and DVGO~\cite{dvgo} progressively increase the feature grid resolution during training which significantly improves training efficiency compared to previous work. The lower-resolution model is used to initialize the higher-resolution model but is not retained for final inference, as shown in \fref{fig:pt1}. To enhance performance with multi-resolution features, NGLoD~\cite{nglod} progressively trains multiple MLPs as encoders and decoders, while only retaining the final decoder to decode integrated multi-resolution features, as shown in \fref{fig:pt2}. Furthermore, Neuralangelo~\cite{neuralangelo} only maintains a single MLP during training, as shown in \fref{fig:pt3}. It progressively activates different levels of hash tables~\cite{muller2022instant} achieving better performance in large-scale scene reconstruction. Similarly, 3D Gaussian Splatting~\cite{kerbl20233dgaussiansplatting} progressively densifies 3D Gaussian achieving top performance on radiance field rendering. Training different level models in these methods is supervised by the same training images. Conversely, the fourth method (\fref{fig:pt4}) used in BungeeNeRF~\cite{bungeenerf} is to apply different models to tackle different-resolution images. BungeeNeRF demonstrates the efficiency of explicitly grouping multi-resolution training images for models to learn multi-level features.
    %BungeeNeRF shows that it is more efficient for models to learn multi-level features after explicitly grouping the training images.
    However, such a method is not universal since multi-resolution images are not available for most scenarios. 
    %Therefore explicitly grouping training images based on resolution of the scene and  
    %In short, at the early stage after a new keyframe $k_{\text{new}}$ joins the hyper primitive mapping process, we use downsized ground truth image and rendered image to compute and backpropagate the loss for $k_{\text{new}}$. As the times that $k_{\text{new}}$ is used as the training camera view increases, we use larger and larger image size for it. At last, after a total usage of $\tau_u$ times, we use the original image size for $k_{\text{new}}$ from then on.
    %strategy designed for the demand of real-time hyper primitives mapping, which is named 
    %advantages
    
\begin{figure}
    \centering
    \includegraphics[width=0.6\linewidth]{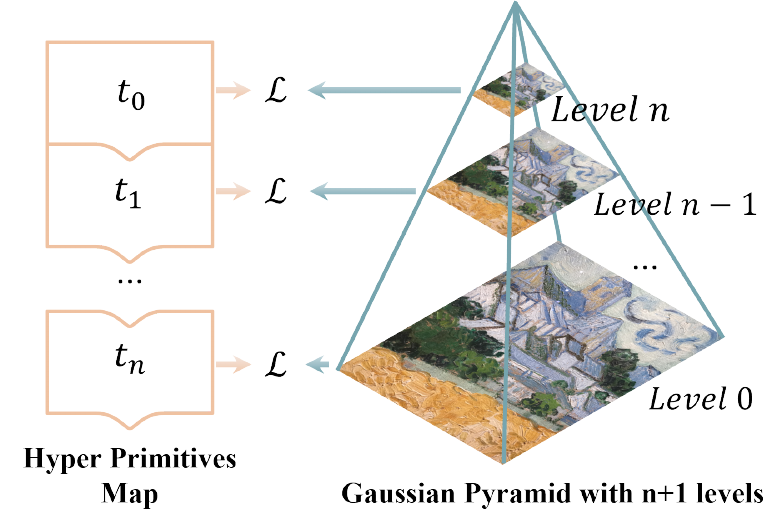}
    \vskip -0.2cm
    \caption{Training process based on the Gaussian pyramid.}
    \label{fig:gp}
    \vskip -0.2cm
\end{figure}

    To make full use of various merits, we propose Gaussian-Pyramid-based (GP) learning (\fref{fig:pt_ours}), a new progressive training method. As illustrated in \fref{fig:gp}, a Gaussian pyramid is a multi-scale representation of an image containing different levels of detail. It is constructed by repeatedly applying Gaussian smoothing and downsampling operations to the original image. At the beginning training step, the hyper primitives are supervised by the highest level of the pyramid, \ie level $n$. As training iteration increases, we not only densify hyper primitives as described in Sec.~\ref{subsubsec:photoslam-opt-densification} but also reduce the pyramid level and obtain a new ground truth until reaching the bottom of the Gaussian pyramid. The optimization process using a Gaussian pyramid with n+1 levels can be denoted as
    \begin{equation}
        \begin{split}
            t_0 &: \mathop{\arg\min} \mathcal{L}\left(I^n_\text{r},  \text{GP}^n(I_\text{gt})\right), \\
            t_1 &:  \mathop{\arg\min}\mathcal{L}\left(I^{n-1}_\text{r}, \text{GP}^{n-1}(I_\text{gt})\right), \\
            &\dots \\
            t_n &:  \mathop{\arg\min}\mathcal{L}\left(I^0_\text{r}, \text{GP}^{0}(I_\text{gt})\right),
        \end{split}
    \end{equation}
    where $\mathcal{L}(I_\text{r},\text{GP}(I_\text{gt}))$ is Eq.~\ref{equ:loss}, while $\text{GP}^n(I_\text{gt})$ denotes the ground image in the level n of the Gaussian pyramid. %This process is also illustrated in \fref{fig:gp}
    In the experiment, we prove that GP learning significantly improves the performance of photorealistic mapping particularly for monocular cameras.
    %Each level $i$ will be used in training for a certain number of times before the next level (i.e. the $(i-1)$-th level) is used.

    %The resulting Gaussian pyramid consists of multiple levels, with each level representing a different scale of the image. The higher levels of the pyramid correspond to lower-resolution versions of the original image, while the lower levels contain finer details.
    
    %To be specific, let $\mathcal{D}^i$ be the domain specific operator that downsamples the ground truth image $I_\text{gt}$ by $2^i$ times, and $I^i_\text{render}$ be the image rendered at $(1/2^i)$ of the original frame image size. As shown in Fig~\ref{fig:pyramid}, a Gaussian Pyramid computes the loss $\mathcal{L}^i$ of its $i$-th level as:
    %The importance of our Gaussian-Pyramid-based multi-level feature learning strategy is apparent under the circumstances where the sensor type is constrained. Though showing little effect in RGB-D and stereo scenes, this strategy plays an essentially important role for successful reconstruction in monocular scenes by speeding up the optimization process, as shown in Sec.~\ref{subsec:exp-ablation}.

\subsection{Loop Closure}\label{subsec:photoslam-loop}
    Loop Closure~\cite{GalvezTRO2012DBOW2} is crucial in SLAM because it helps address the problem of accumulated errors and drift that can occur during the localization and geometry mapping process. After detecting a closing loop, we can correct local keyframes and hyper primitives by similarity transformation. With corrected camera poses, the photorealistic mapping component can further get rid of the ghosting caused by odometry drifts and improve the mapping quality.

    %In our geometric component, a new keyframe $k_\text{new}$ from the local mapping thread will be used to try detecting and closing loops. To be specific, with DBoW2~\cite{GalvezTRO2012DBOW2} methods, we first use the ORB features of $k_\text{new}$ to compute the bags of words vector of it in the local mapping thread. Then, in the loop closing thread, we compare this vector with bag of words vectors of covisibility graph neighbors of $k_\text{new}$, and retain the lowest score $s_\text{min}$. We query the keyframe database and get rid of all keyframes with a score lower than $s_\text{min}$, and mark the remainders as candidates. We accept a loop candidate if three consistent loop candidates are detected. After detecting a loop closure, we first perform RANSAC similarity optimization between $k_\text{new}$ and the loop candidates, in order to optimize the camera pose of $k_\text{new}$, as well as correct local keyframes and map points connected to it. Then, a pose-graph optimization~\cite{strasdat2010scale} is performed on the essential graph~\cite{mur2015orb} to correct all keyframe camera poses. The photorealistic mapping process benefits a lot from loop closures. With corrected camera poses, it is able to adjust the 3D Gaussians by moving them, or lowering their density and pruning them. As a result, we can get rid of the ghosting caused by odometry drifts and improve the mapping quality.
\section{Experiment}\label{sec:experiment}
    In this section, we compare Photo-SLAM to other state-of-the-art (SOTA) SLAM and real-time 3D reconstruction systems in various scenarios encapsulating monocular, stereo, RGB-D cameras, and indoor and outdoor environments. In addition, we evaluate Photo-SLAM performance on various hardware configurations to demonstrate its efficiency. Finally, we conduct an ablation study to verify the effectiveness of the proposed algorithms.

\begin{table*}[t]
    \centering
    %\tiny
    \footnotesize
    \tabcolsep=0.15cm 
    \resizebox{0.88\textwidth}{!}{
    \begin{tabular}{cc|ccccccccc}
    \toprule
	\multicolumn{2}{c}{On Replica Dataset}& \multicolumn{2}{c}{Localization (cm)}& \multicolumn{3}{c}{Mapping}& \multicolumn{4}{c}{Resources}\\\cmidrule(lr){1-2} \cmidrule(lr){3-4} \cmidrule(lr){5-7} \cmidrule(lr){8-11}
	Cam&Method&\makecell{RMSE $\downarrow$} &STD $\downarrow$& { PSNR $\uparrow$} &SSIM $\uparrow$ & LPIPS $\downarrow$ & \makecell{Operation Time} $\downarrow$ & \makecell{Tracking FPS} $\uparrow$  & \makecell{Rendering FPS} $\uparrow$ & \makecell{GPU Memory Usage} $\downarrow$ \\
	\midrule      
    \multirow{8}{*}{{\rotatebox{90}{Mono}}}& ORB-SLAM3~\cite{campos2021orb3} &3.942	&3.115 &-&-&- 
    &$<$1 mins &\markfirst{58.749} &- & 0\\
    &DROID-SLAM~\cite{droid}&\marksecond{0.725} &\markfirst{0.308}&-&-&- 
    &$<$2 mins &35.473 &- & 11 GB \\
    &Nice-SLAM*~\cite{niceslam}&99.9415&35.336 &16.311 &0.720 &0.439
    &$>$10 mins &2.384 &0.944& 12 GB\\
    &Orbeez-SLAM~\cite{chung2023orbeez}&-&- & 23.246	&0.790 &0.336 
    &$<$5 mins &\marksecond{49.200}&1.030& \markthird{6 GB}\\ 
    %&Go-SLAM~\cite{go-slam}&71.036 &24.594 &21.207 &0.647	&0.442 &21.499	&0.827&22 GB \\ 
    &Go-SLAM~\cite{go-slam}&71.054&24.593 &21.172 &0.703 &0.421 
    &$<$5 mins &25.366 &0.821 &22 GB\\
    &\textbf{Ours (Jetson)}&1.235&\markthird{0.756} &\markthird{29.284} &\markthird{0.883} &\markthird{0.139}
    &$<$5 mins &18.315 &\markthird{95.057}& \markfirst{4 GB}\\ 
    &\textbf{Ours (Laptop)}&\markfirst{0.713}&\marksecond{0.524}&\marksecond{33.049} &\markfirst{0.926} &\marksecond{0.086} 
    &$<$5 mins &19.974&\marksecond{353.504} & \markfirst{4 GB}\\ 
    %&\textbf{Ours}&1.576&0.995&31.648 &0.907 &0.106 &45.927 & &4 GB\\ 
    &\textbf{Ours}&\markthird{1.091} &0.892&\markfirst{33.302} &\markfirst{0.926} &\markfirst{0.078}
    &$<$2 mins &\markthird{41.646}&\markfirst{911.262}&\markthird{6 GB}\\ 
    \midrule\midrule
    \multirow{11}{*}{{\rotatebox{90}{RGB-D}}}& 
    ORB-SLAM3~\cite{campos2021orb3} &1.833 &1.478 &-&-&- 
    &$<$1 mins &\markfirst{52.209}&- &0\\ 
    &DROID-SLAM~\cite{droid}&0.634&\marksecond{0.248}	&-&-&- 
    &$<$2 mins &36.452&- & 11 GB\\ 
    &BundleFusion~\cite{bundlefusion}&1.606 &0.969 &23.839	&0.822 &0.197 
    &$<$5 mins &8.630 & - & 5 GB \\
    &Nice-SLAM~\cite{niceslam}&2.350&1.590 &26.158 &0.832 &0.232
    &$>$10 mins &2.331 &0.611 & 12 GB\\ 
    &Orbeez-SLAM~\cite{chung2023orbeez}&0.888&0.562 &\markthird{32.516}&\markthird{0.916} &0.112
    &$<$5 mins &\markthird{41.333}&1.401& 6 GB\\ 
    &ESLAM~\cite{eslam}&\markfirst{0.568} &\markthird{0.274} &30.594 &0.866 &0.162
    &$<$5 mins &6.687	&2.626& 21 GB \\ 
    %&Co-SLAM~\cite{}&1.182&0.644 &30.183 &0.866	&0.174&14.551&3.714& 4 GB\\ 
    &Co-SLAM~\cite{coslam}&1.158&0.602&30.246 &0.864	&0.175
    &$<$5 mins &14.575 &3.745& \markfirst{4 GB}\\ 
    %&Go-SLAM~\cite{go-slam}&\marksecond{0.570}&\markfirst{0.218} &25.230 &0.785 &0.336&19.697&0.447& 24 GB\\ 
    &Go-SLAM~\cite{go-slam}&\marksecond{0.571} &\markfirst{0.218} &24.158 &0.766 &0.352 
    &$<$5 mins &19.437 &0.444 &24 GB\\ 
    &Point-SLAM~\cite{point-slam}&0.596 &0.249 &34.632 &0.927 &0.083 &$>$2 hrs &0.345 &0.510 &24 GB\\ 
    &\textbf{Ours (Jetson)}&\markthird{0.581}&0.289 &31.978 &\markthird{0.916} &\markthird{0.101}
    &$<$5 mins &17.926&\markthird{116.395} & \markfirst{4 GB}\\ 
    &\textbf{Ours (Laptop)}&0.590&0.289 &\marksecond{34.853} &\markfirst{0.944} &\marksecond{0.062}
    &$<$5 mins &20.597&\marksecond{396.082} & \markfirst{4 GB}\\ 
    %&\textbf{Ours}&0.582&0.283 &34.010 &0.935 &0.069 &44.846& & 4 GB\\ 
    &\textbf{Ours}&0.604&0.298&\markfirst{34.958}&\marksecond{0.942} &\markfirst{0.059}
    &$<$2 mins &\marksecond{42.485} &\markfirst{1084.017}&5 GB\\ 
    \bottomrule
    \end{tabular}}
    \vspace{-0.5em}
    \caption{Quantitative results on the Replica dataset. We mark the best two results with \colorfirsttext{first} and \colorsecondtext{second}. Nice-SLAM* means the depth supervision is disabled. ``-'' denotes the system does not support view rendering or fails to track camera poses. The results of Photo-SLAM running on the laptop and Jetson platform are denoted as ``Ours (Laptop)'' and  ``Ours (Jetson)'' respectively.} %The results show that, in both monocular and RGB-D scenes, our Photo-SLAM achieves the best mapping quality, rendering speed and GPU memory usage while preserving competitive localization accuracy and tracking FPS. 
    \label{tab:replica}
\end{table*}

\begin{comment}
\begin{figure*}
    \centering
    \def\imw{0.19}
     \includegraphics[width=\imw\linewidth]{figure/compare/replica_rgbd/00420_bundle.jpg}
     \includegraphics[width=\imw\linewidth]{figure/compare/replica_rgbd/00420_nice.jpg}
     \includegraphics[width=\imw\linewidth]{figure/compare/replica_rgbd/00420_eslam.jpg}
     \includegraphics[width=\imw\linewidth]{figure/compare/replica_rgbd/00420_co.jpg}
     \includegraphics[width=\imw\linewidth]{figure/compare/replica_rgbd/00420_ours.jpg}
    \caption{RGBD}
    \label{fig:enter-label}
\end{figure*}
\end{comment}

\begin{figure*}
    \centering
    \def\imw{0.19}
    \subfloat[BundleFusion~\cite{bundlefusion}]{
    \begin{minipage}{\imw\linewidth}
        \includegraphics[width=\linewidth]{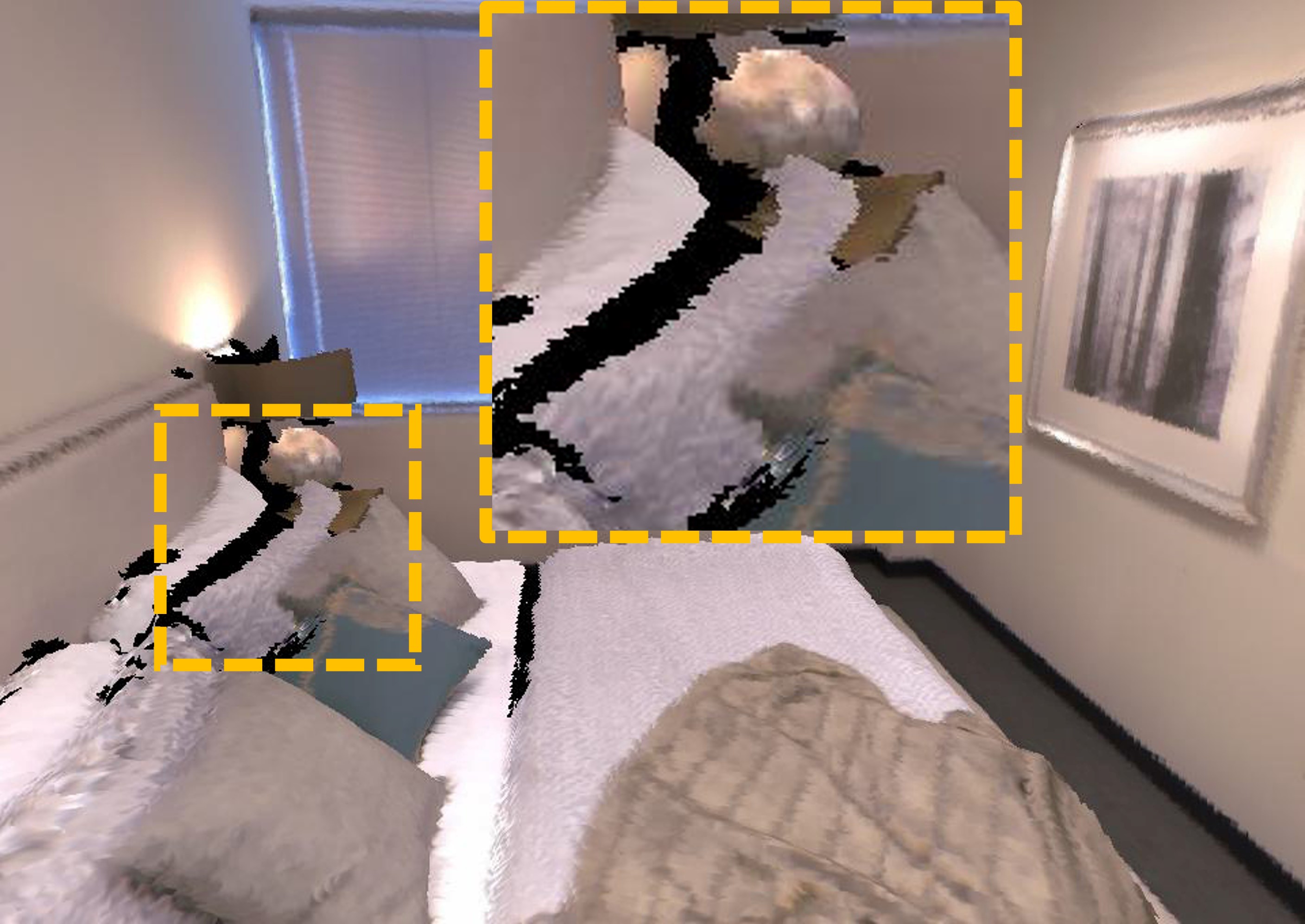}
        \includegraphics[width=\linewidth]{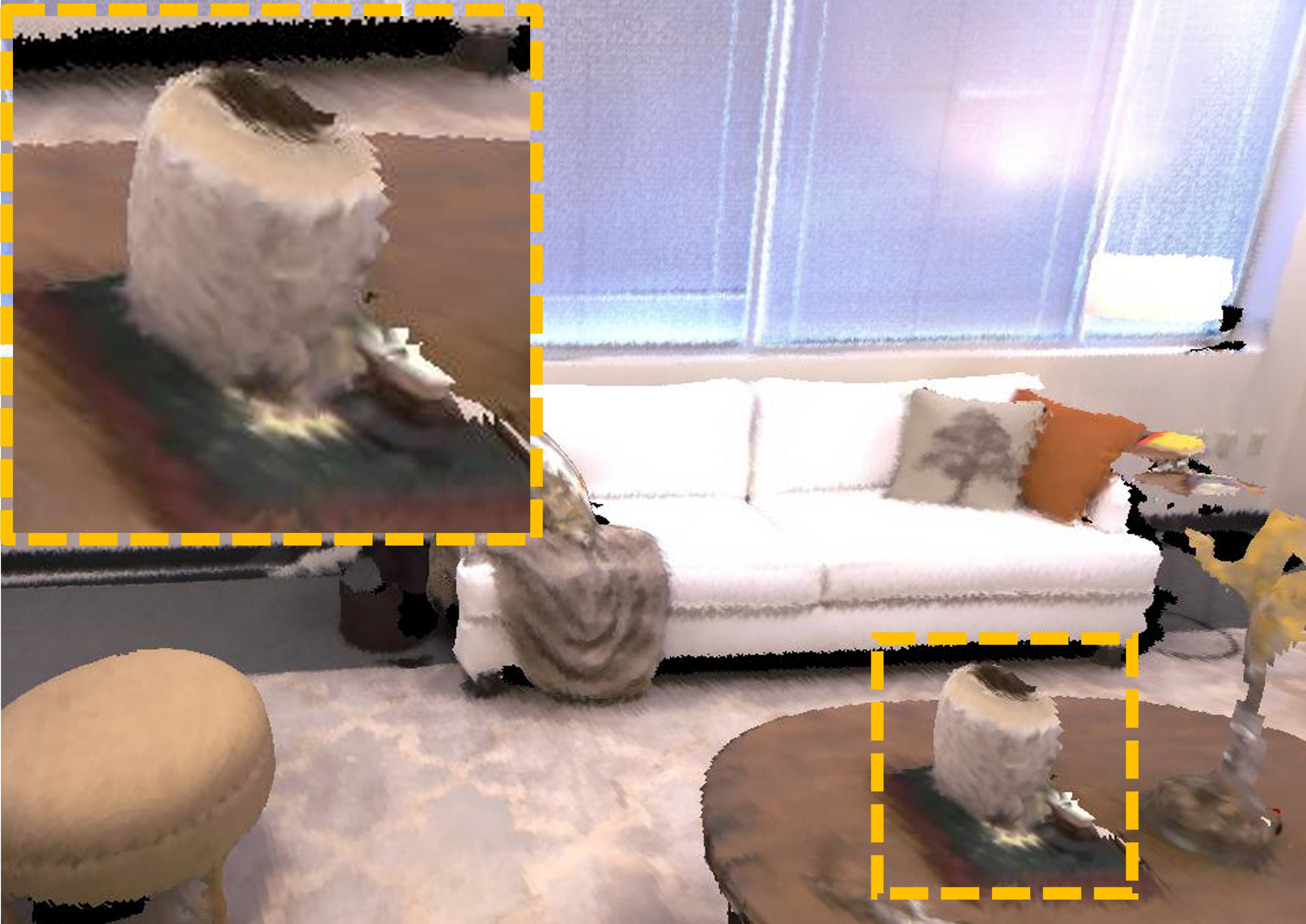}
    \end{minipage}
    }
    \subfloat[Nice-SLAM~\cite{niceslam}]{
    \begin{minipage}{\imw\linewidth}
        \includegraphics[width=\linewidth]{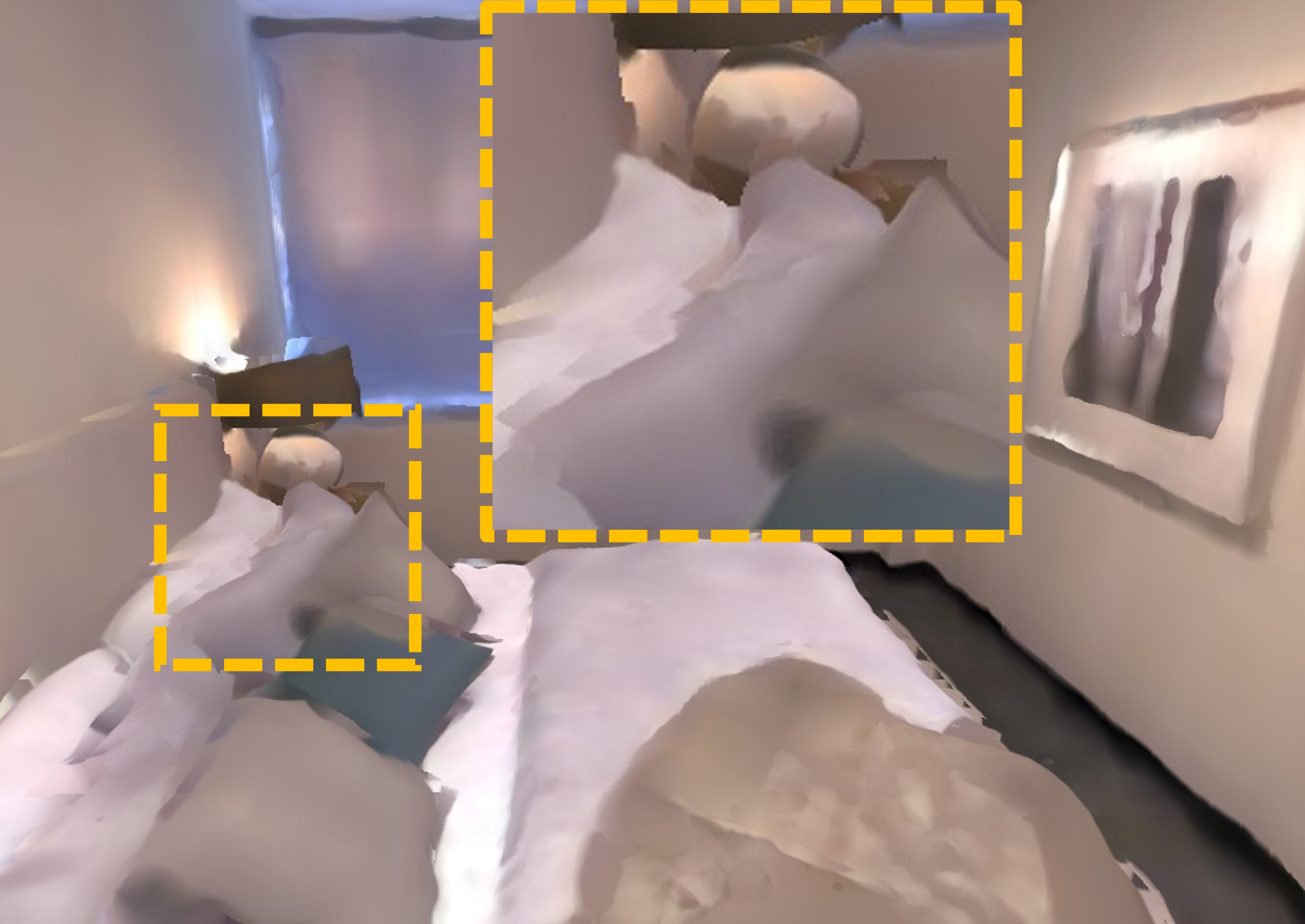}
        \includegraphics[width=\linewidth]{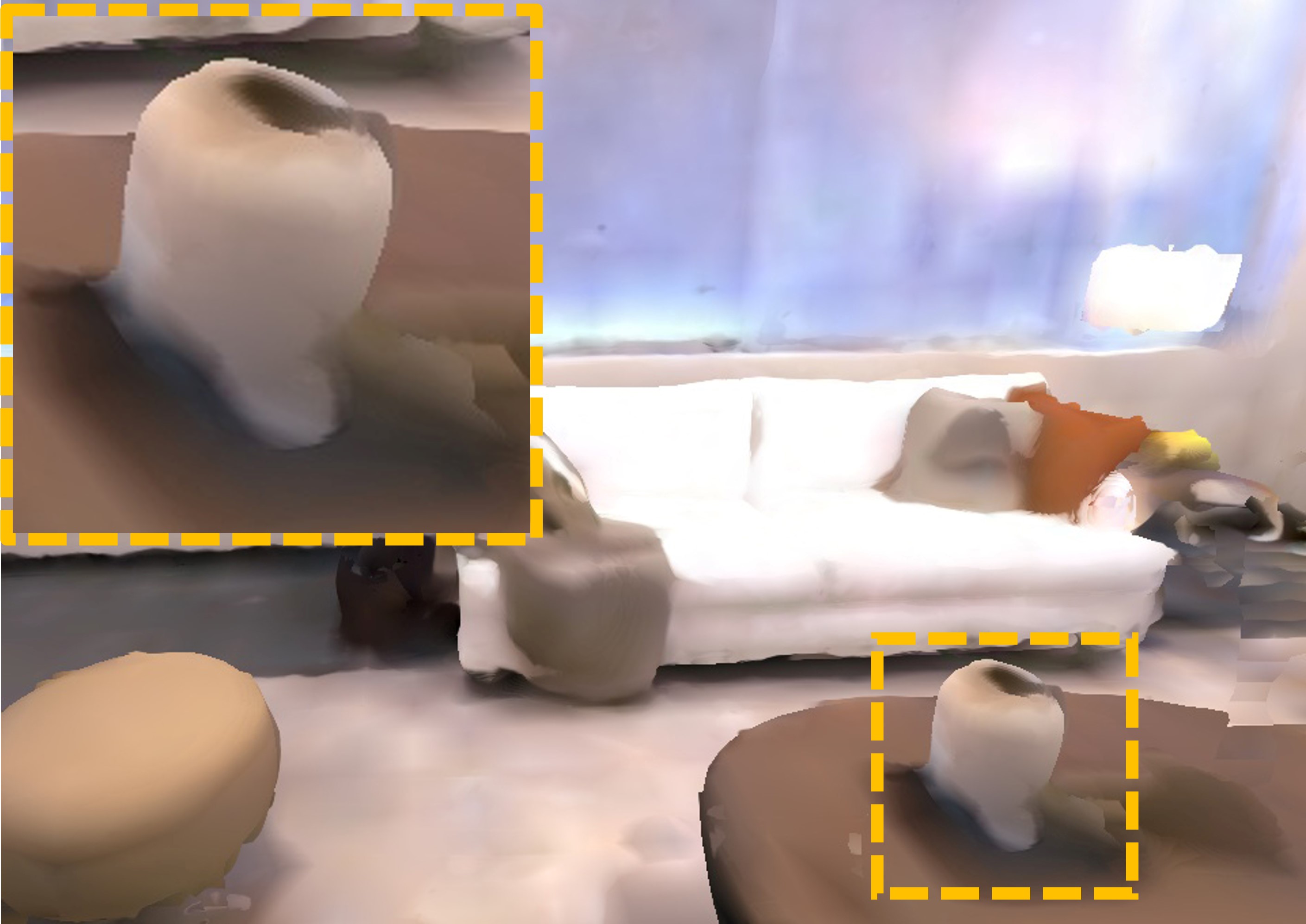}
    \end{minipage}
    }  
    \subfloat[ESLAM~\cite{eslam}]{
    \begin{minipage}{\imw\linewidth}
        \includegraphics[width=\linewidth]{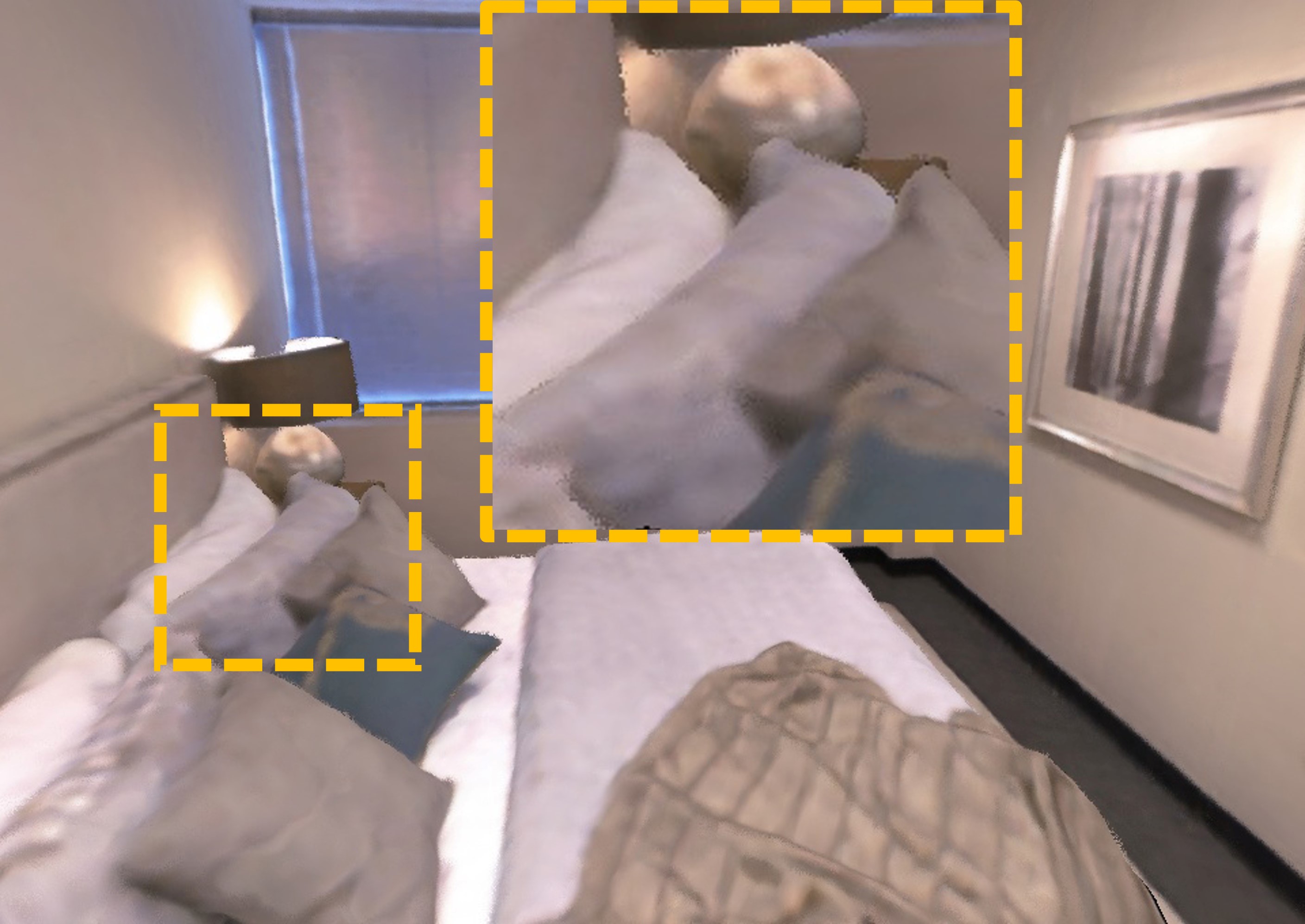}
        \includegraphics[width=\linewidth]{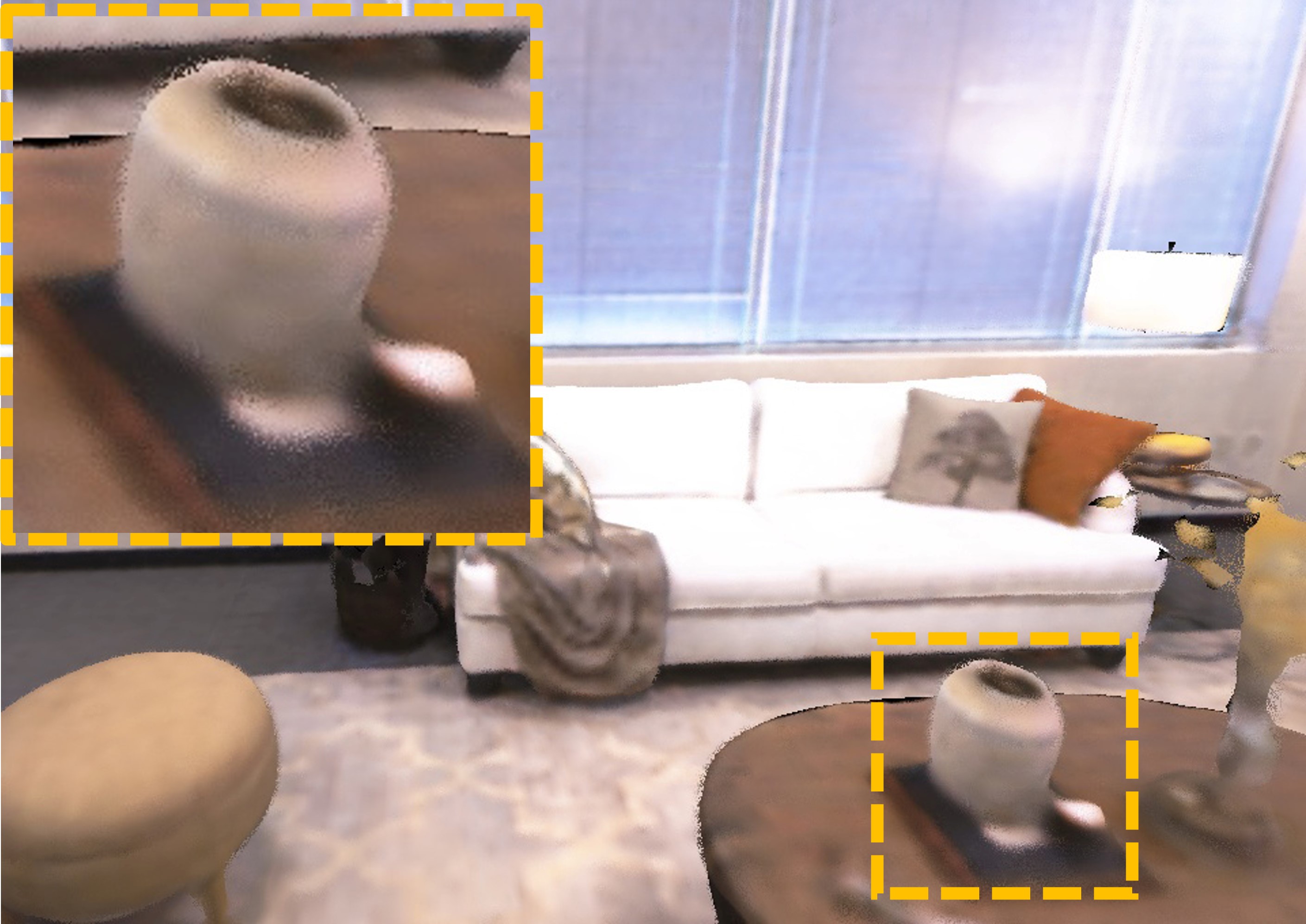}
    \end{minipage}
    }  
    \subfloat[Co-SLAM~\cite{coslam}]{
    \begin{minipage}{\imw\linewidth}
        \includegraphics[width=\linewidth]{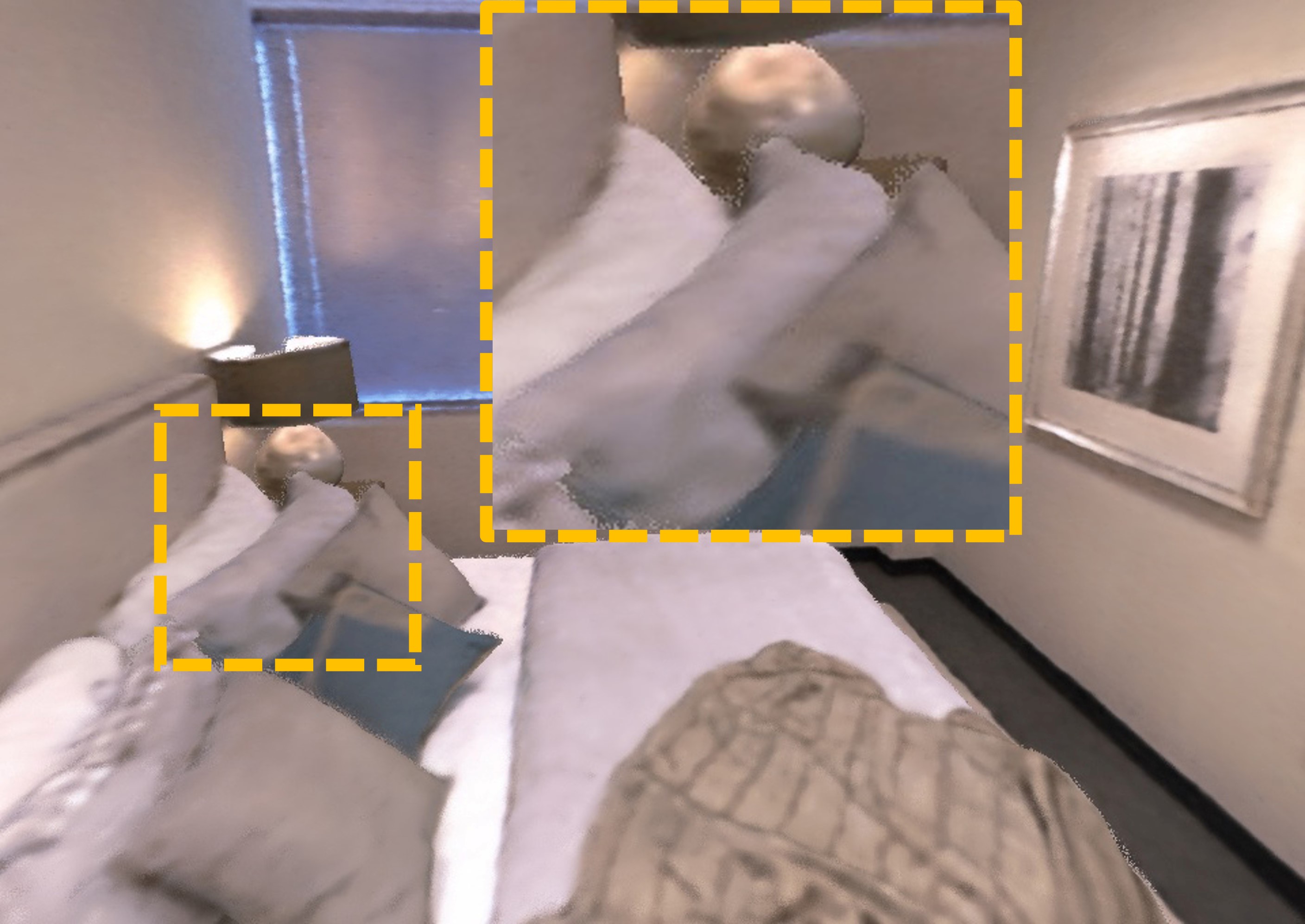}
        \includegraphics[width=\linewidth]{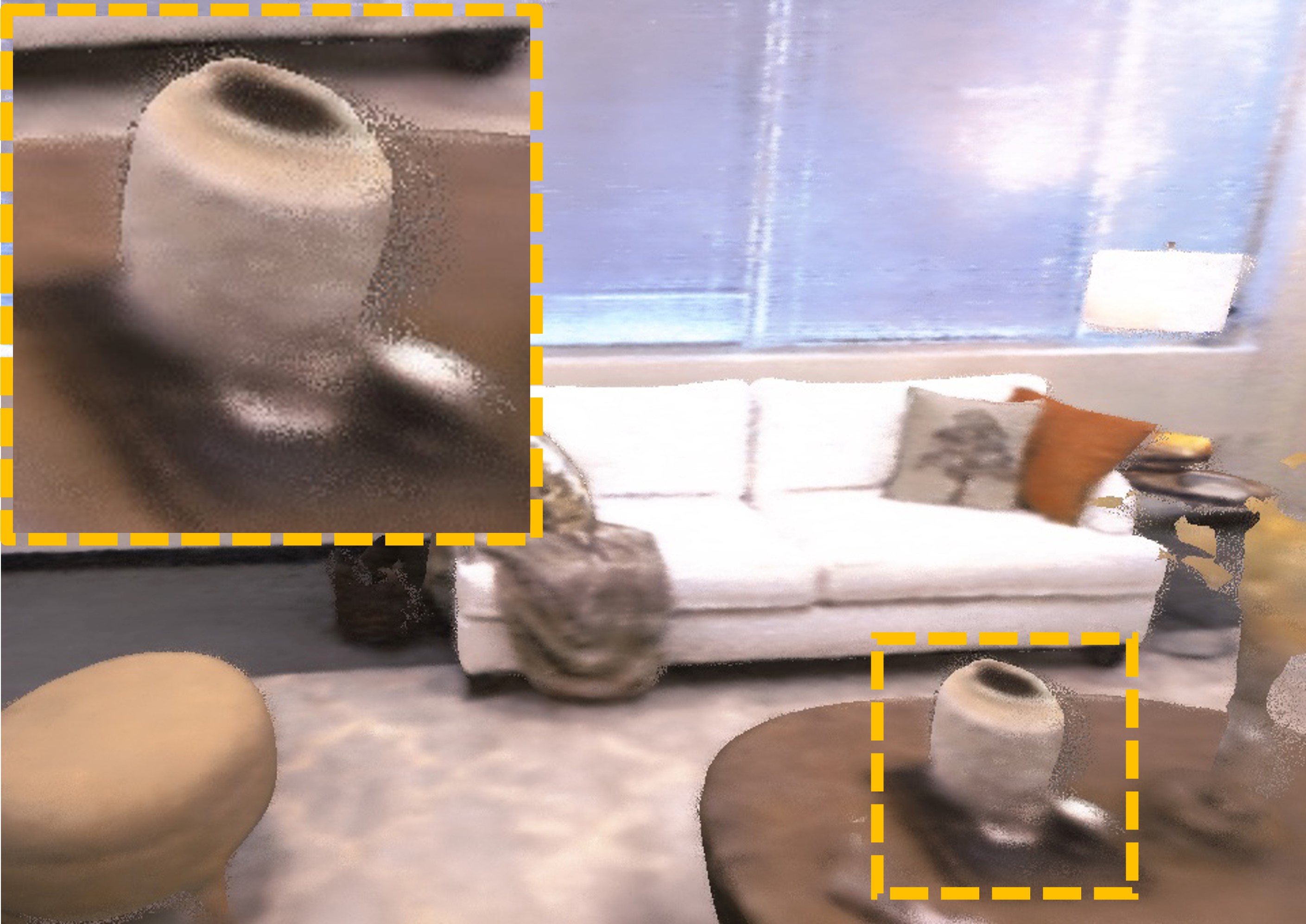}
    \end{minipage}
    }  
    \subfloat[Ours]{
    \begin{minipage}{\imw\linewidth}
        \includegraphics[width=\linewidth]{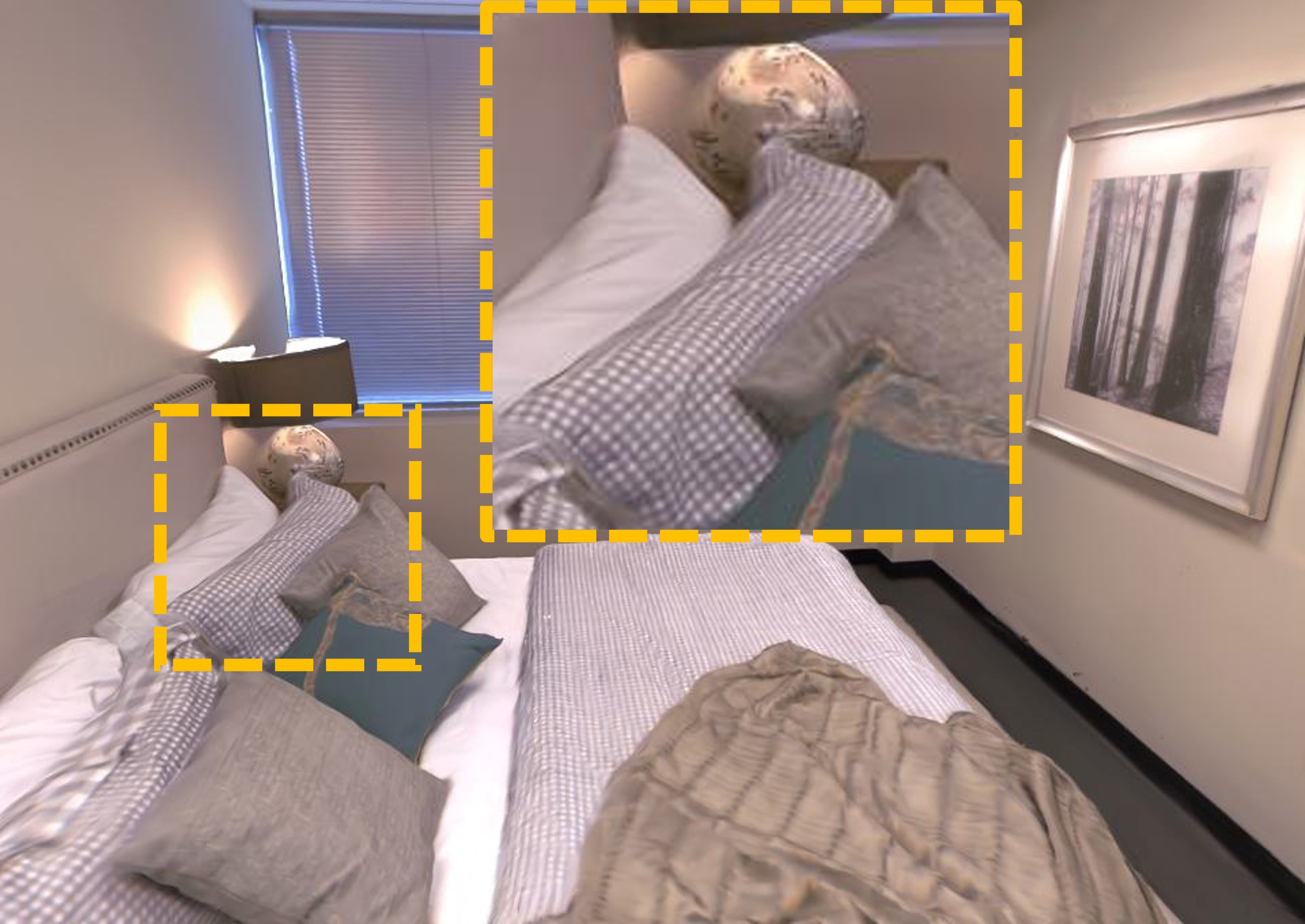}
        \includegraphics[width=\linewidth]{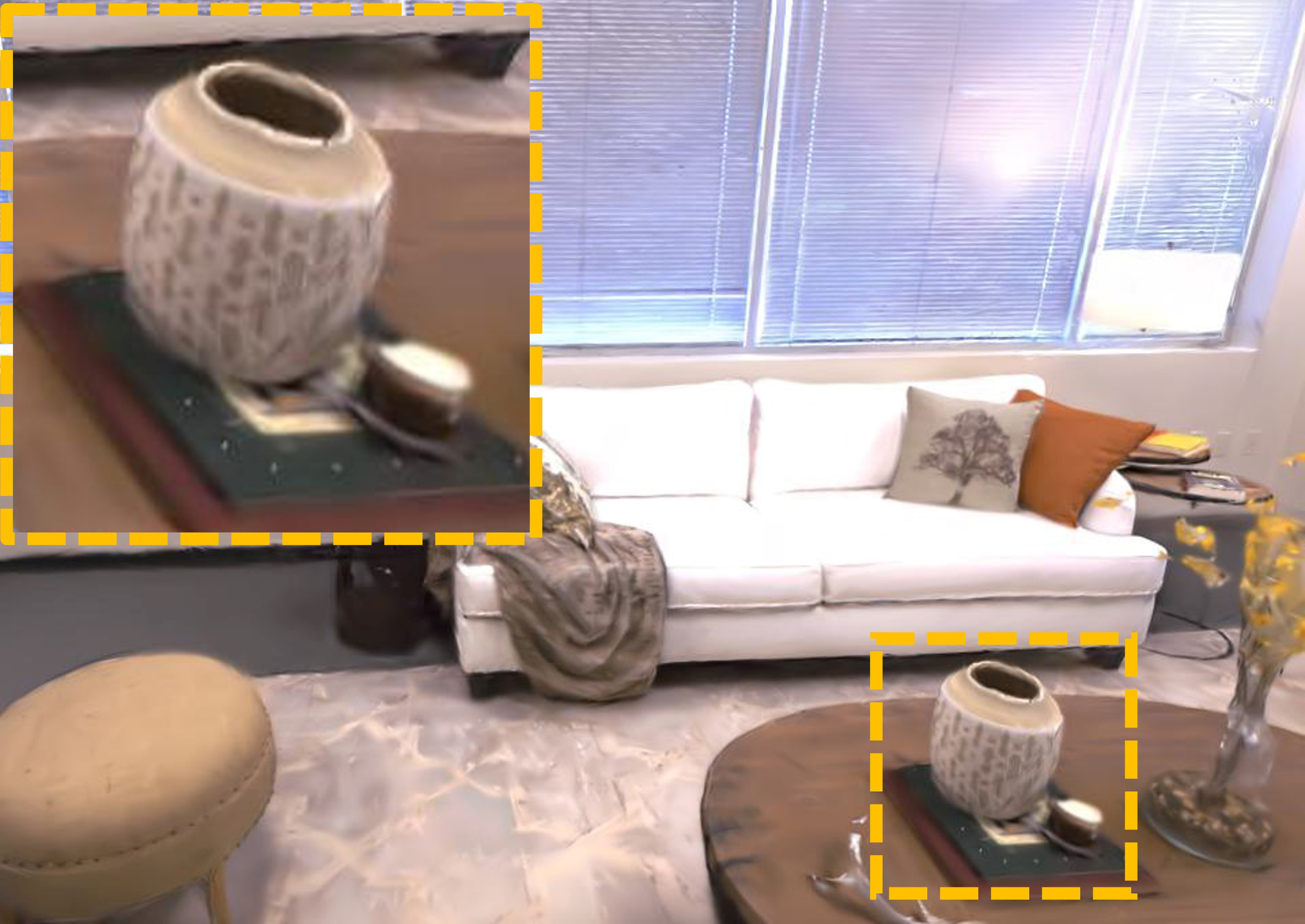}
    \end{minipage}
    }  
    \vspace{-0.5em}
    \caption{Qualitative comparison of diverse systems using RGB-D images from dataset Replica. Photo-SLAM can reconstruct high-fidelity scenes while others are over-smooth and have obvious artifacts. Zoom in for better views.}
    %\vspace{-0.1em}
    \label{fig:replica-rgbd}
\end{figure*}

\begin{figure}
    \centering
    \def\imgw{0.48}
    %\subfloat[Ground Truth]{
     %\includegraphics[width=\imgw\linewidth]{figure/compare/replica_rgbd/frame001441.jpg}\,
     %\includegraphics[width=\imgw\linewidth]{figure/compare/replica_rgbd/frame001618.jpg}
     %}\\
     \subfloat[Go-SLAM~\cite{go-slam}]{
     \includegraphics[width=\imgw\linewidth]{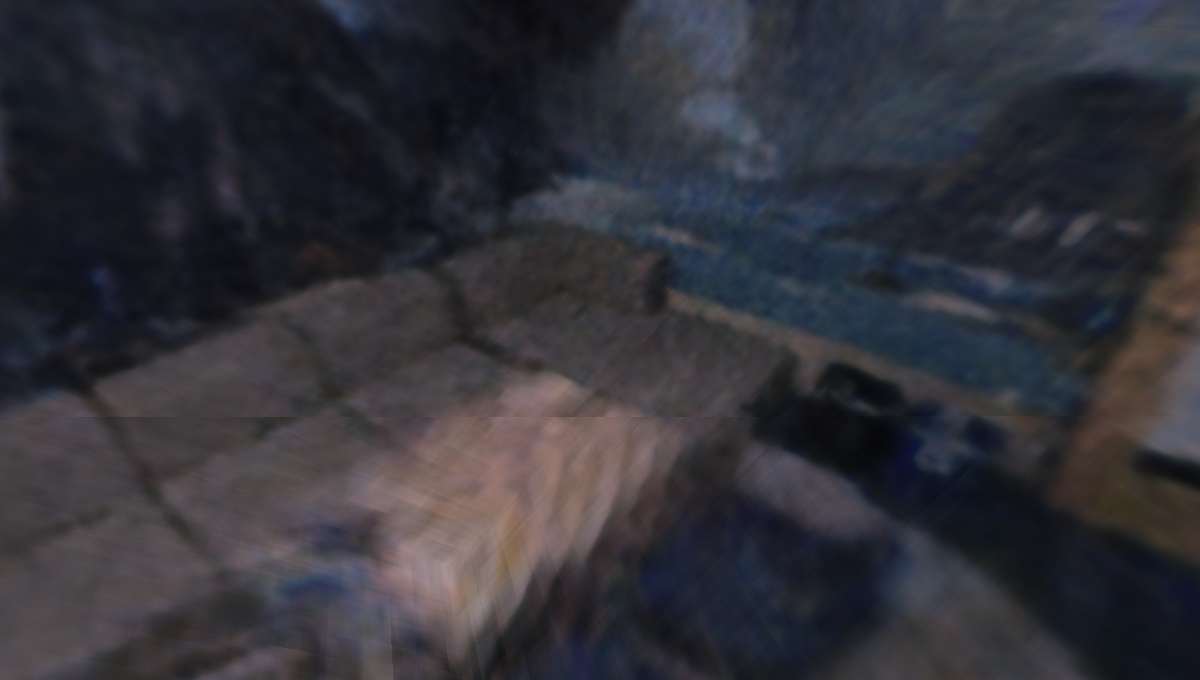}\,
     \includegraphics[width=\imgw\linewidth]{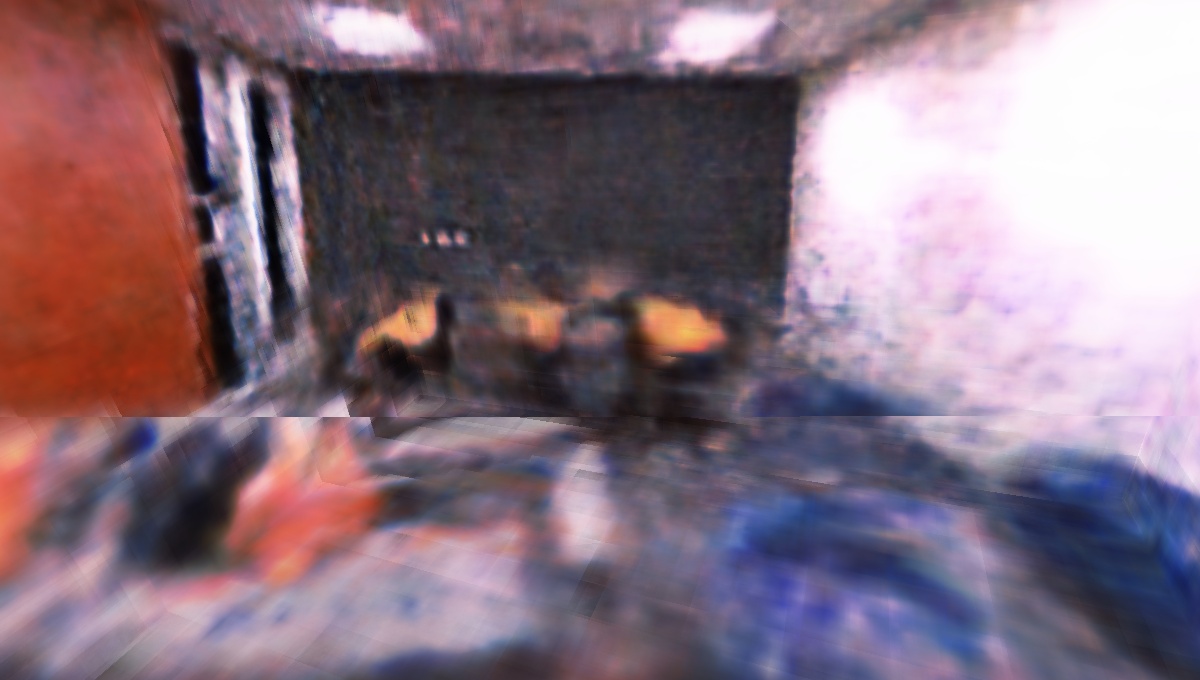}}\\
     \subfloat[Orbeez-SLAM~\cite{chung2023orbeez}]{
     \includegraphics[width=\imgw\linewidth]{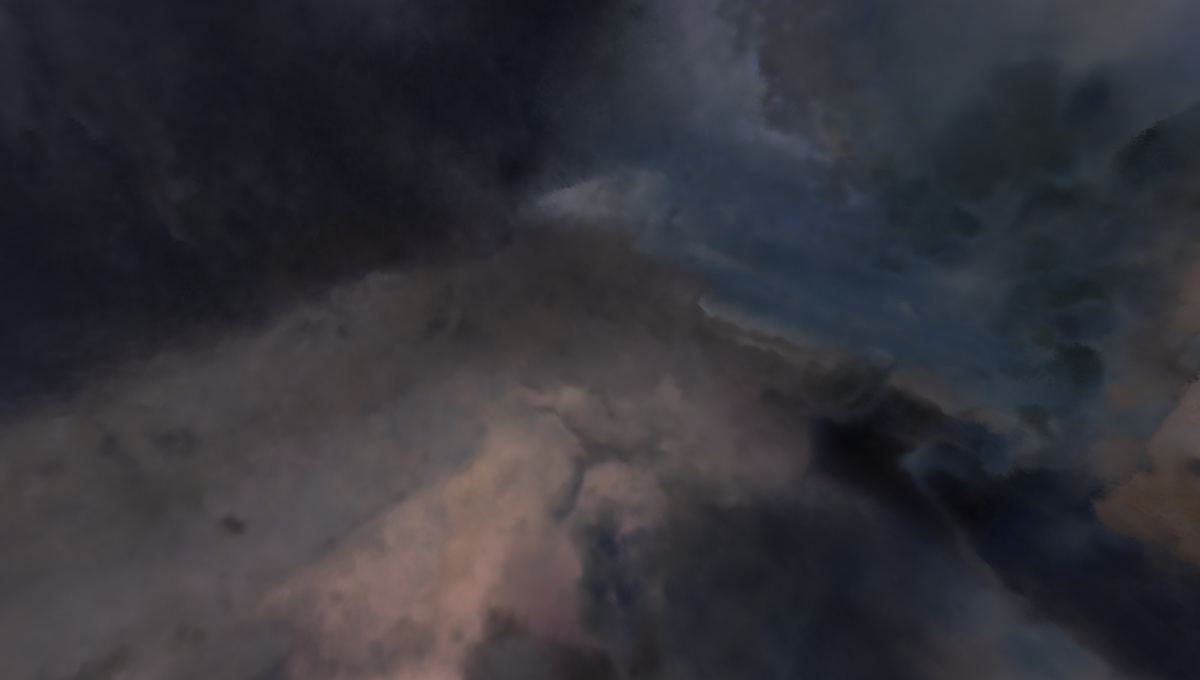}\,
     \includegraphics[width=\imgw\linewidth]{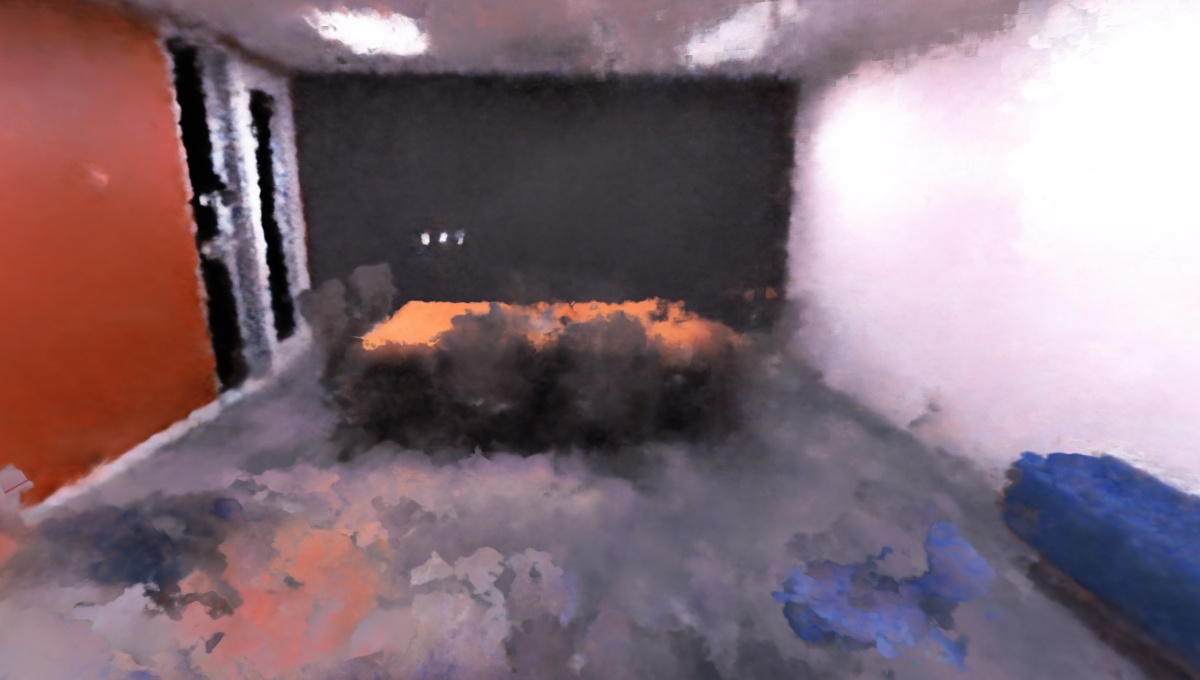}}\\
     \subfloat[Ours]{
     \includegraphics[width=\imgw\linewidth]{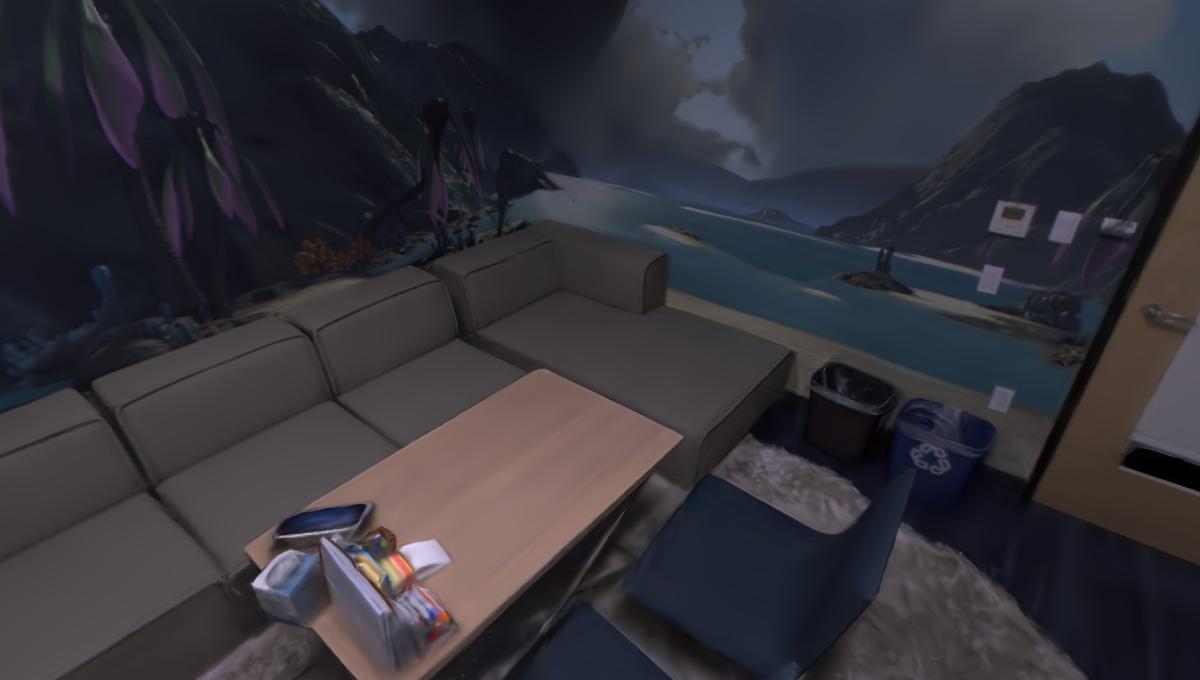}\,
     \label{fig:replica-mono-ours}\includegraphics[width=\imgw\linewidth]{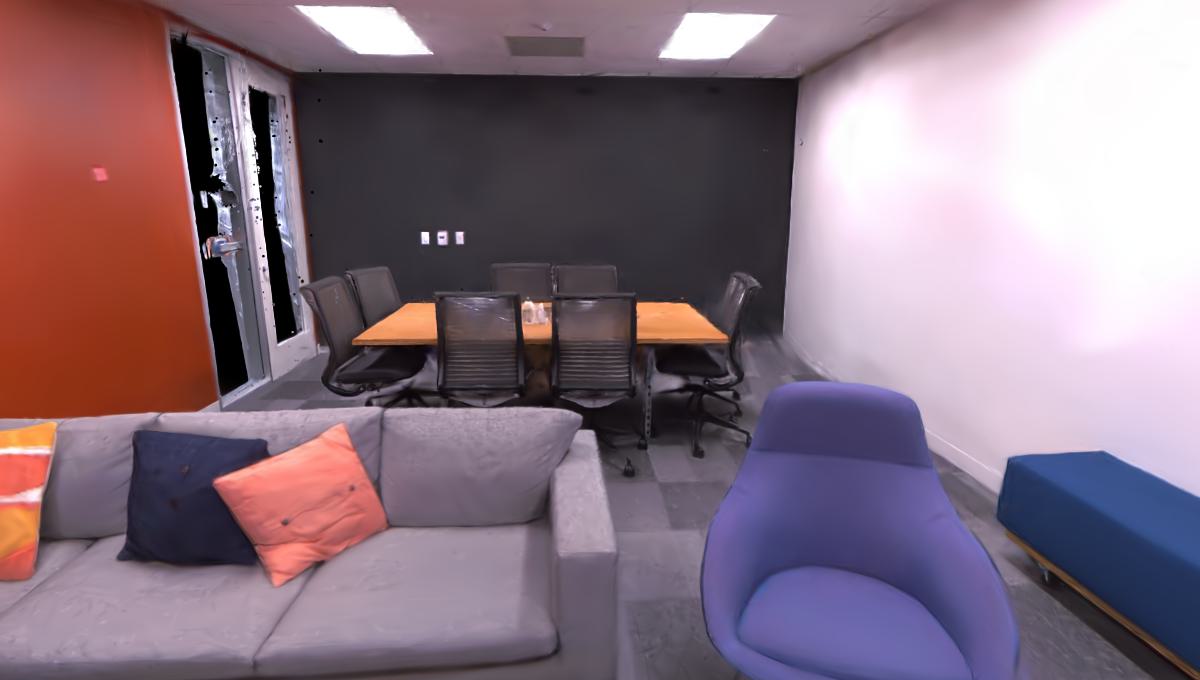}}
     \vskip -0.2cm
    \caption{Mapping comparison with other monocular camera systems on \textit{Replica office3} scene.}
    \label{fig:replica-mono}
    %\vskip -0.2cm
\end{figure}

\begin{table*}%[t]
    \centering
    %\tiny
    \footnotesize
    \tabcolsep=0.08cm 
    \resizebox{0.8\textwidth}{!}{
    \begin{tabular}{cc|cccccccccccc}
    \toprule
	\multicolumn{2}{c}{On TUM Dataset}&  \multicolumn{4}{c}{fr1-desk}& \multicolumn{4}{c}{fr2-xyz}& \multicolumn{4}{c}{fr3-office}\\\cmidrule(lr){1-2} \cmidrule(lr){3-6} \cmidrule(lr){7-10}\cmidrule(lr){11-14}
	{\scriptsize Cam}&{\scriptsize Method}&{\scriptsize RMSE (cm) $\downarrow$} &{\scriptsize PSNR $\uparrow$} &{\scriptsize SSIM $\uparrow$} & {\scriptsize LPIPS $\downarrow$}&{\scriptsize RMSE (cm) $\downarrow$}& {\scriptsize PSNR $\uparrow$} &{\scriptsize SSIM $\uparrow$} & {\scriptsize LPIPS $\downarrow$}&{\scriptsize RMSE (cm) $\downarrow$}& {\scriptsize PSNR $\uparrow$} &{\scriptsize SSIM $\uparrow$ }& {\scriptsize LPIPS $\downarrow$} \\
	\midrule      
    \multirow{6}{*}{{\rotatebox{90}{Mono}}}& ORB-SLAM3~\cite{campos2021orb3} &\markfirst{1.534}&-&-&-&\marksecond{0.720}&-&-&-&\marksecond{1.400}&-&-&- \\
    &DROID-SLAM~\cite{droid}&78.245&-&-&-&36.050&-&-&-&154.383&-&-&- \\
    &Go-SLAM~\cite{go-slam}&33.122&11.705&0.406&0.614&28.584&14.807&0.443&0.572&105.755&13.572&0.480&0.643\\
    %&Orbeez-SLAM~\cite{chung2023orbeez}&-&-&-&-&\markfirst{0.237}&20.555&\markfirst{0.754}&0.243&2.171&17.888&0.603&0.507\\ 
    &\textbf{Ours (Jetson)}&1.757&\markthird{18.811}&\markthird{0.681}&\markthird{0.329}&\markfirst{0.558}&\marksecond{21.347}&\marksecond{0.727}&\markthird{0.187}&1.687&\markthird{18.884}&\markthird{0.672}&\markthird{0.289}\\ 
    &\textbf{Ours (Laptop)}&\markthird{1.549}&\marksecond{20.515}&\marksecond{0.733}&\marksecond{0.241}&0.852&\markfirst{21.575}&\markfirst{0.739}&\markfirst{0.157}&\markthird{1.542}&\marksecond{19.138}&\marksecond{0.680}&\marksecond{0.259}\\
    &\textbf{Ours}&\marksecond{1.539}&\markfirst{20.972}&\markfirst{0.743}&\markfirst{0.228}&0.984&\markthird{21.072}&0.726&\marksecond{0.166}&\markfirst{1.257}&\markfirst{19.591}&\markfirst{0.692}&\markfirst{0.239}\\ 
    %%%%% RGBD %%%%%
    \midrule\midrule
    \multirow{9}{*}{{\rotatebox{90}{RGB-D}}}& ORB-SLAM3~\cite{campos2021orb3} &\markfirst{1.724}&-&-&-&0.385&-&-&-&1.698&-&-&-\\ 
    &DROID-SLAM~\cite{droid}&91.985&-&-&-&41.833&-&-&-&160.141&-&-&-\\
    %&Orbeez-SLAM~\cite{chung2023orbeez}&\markfirst{1.456}&17.459&0.625&0.439&0.381&\markfirst{24.424}&\markfirst{0.809}&\marksecond{0.154}&\marksecond{1.060}&\markfirst{24.206}&\markfirst{0.808}&\marksecond{0.155}\\ 
    &Nice-SLAM~\cite{niceslam}&19.317&12.003&0.417&0.510&36.103&18.200&0.603&0.313&25.309&16.341&0.548&0.386\\ 
    &ESLAM~\cite{eslam}&3.359&17.497&0.561&0.484&31.448&22.225&0.727&0.233&25.808&19.113&0.616&0.359\\ 
    &Co-SLAM~\cite{coslam}&3.094&16.419&0.482&0.591&31.347&19.176&0.595&0.374&25.374&17.863&0.547&0.452\\ 
    &Go-SLAM~\cite{go-slam}&2.119&15.794&0.531&0.538&31.788&16.118&0.534&0.419&26.802&16.499&0.566&0.569\\ 
    &\textbf{Ours (Jetson)}&4.571&\markthird{18.273}&\markthird{0.663}&\markthird{0.338}&\marksecond{0.360}&\markfirst{23.127}&\markfirst{0.780}&\markfirst{0.149}&1.874&19.781&0.701&0.235\\ 
    &\textbf{Ours (Laptop)}&\marksecond{1.891}&\marksecond{20.403}&\marksecond{0.728}&\marksecond{0.251}&\markthird{0.361}&\markthird{22.570}&\marksecond{0.777}&\marksecond{0.158}&\marksecond{1.315}&\marksecond{21.569}&\marksecond{0.749}&\marksecond{0.184}\\ 
    &\textbf{Ours}&2.603&\markfirst{20.870}&\markfirst{0.743}&\markfirst{0.239}&\markfirst{0.346}&\marksecond{22.094}&0.765&0.169&\markfirst{1.001}&\markfirst{22.744}&\markfirst{0.780}&\markfirst{0.154}\\
    \bottomrule
    \end{tabular}}
    \vskip -0.2cm
    \caption{Quantitative results on the TUM RGB-D dataset. We mark the best two results with \colorfirsttext{first} and \colorsecondtext{second}.} % (Orbeez-SLAM failed with monocular fr1-desk.) We gain competitive localization accuracy in all the scenes. In monocular scenes, our Photo-SLAM shows state-of-the-art photorealistic qualities. In RGB-D scenes, we obtain qualitative results on par with the leading Orbeez-SLAM.
    \label{tab:tum}
    \vskip -0.1cm
\end{table*}

\subsection{Implementation and Experiment Setup}\label{subsec:exp-setup}
    We implemented Photo-SLAM fully in C++ and CUDA, making use of ORB-SLAM3~\cite{campos2021orb3}, 3D Gaussian splatting~\cite{kerbl20233dgaussiansplatting}, and the LibTorch framework. The optimization of photorealistic mapping is performed with the Stochastic Gradient Descent algorithm while we use a fixed learning rate and $\lambda=0.2$. Considering image resolution of the testing dataset, the level of the Gaussian pyramid is set to three, \ie $n=2$ by default.
    The compared baseline includes a SOTA classical SLAM system ORB-SLAM3~\cite{campos2021orb3}, a real-time RGB-D dense reconstruction system BundleFusion~\cite{bundlefusion}, a deep-learning based system DROID-SLAM~\cite{droid}, and recent SLAM systems supporting view synthesis, \ie Nice-SLAM~\cite{niceslam}, Orbeez-SLAM~\cite{chung2023orbeez}, ESLAM~\cite{eslam}, Co-SLAM~\cite{coslam}, and Point-SLAM~\cite{point-slam} and Go-SLAM~\cite{go-slam}. 
    %We implemented Photo-SLAM fully in C++ and CUDA, using the LibTorch framework and several custom CUDA kernels designed for point cloud and stereo vision operations. We also built an viewer with the open-souruce Dear ImGui~\cite{software2023imgui} for visualized and interactive training. Our open-source implementation will be made publicly available.

    \noindent\textbf{Hardware.} We ran Photo-SLAM and all compared methods using their official code in a desktop with an NVIDIA RTX 4090 24 GB GPU, an Intel Core i9-13900K CPU, and 64 GB RAM. We further tested Photo-SLAM on a laptop and a Jetson AGX Orin Developer Kit. The laptop is equipped with an NVIDIA RTX 3080ti 16 GB Laptop GPU, an Intel Core i9-12900HX, and 32 GB RAM.  %the estimated trajectory accuracy and the computed photometric loss an embedded platform 

    \noindent\textbf{Datasets and Metrics.} We performed tests for monocular and RGB-D sensor types on the well-known RGB-D datasets: the Replica dataset~\cite{dataset2019replica, imap} and the TUM RGB-D dataset~\cite{dataset2012tum}. As for stereo tests, we used the EuRoC MAV dataset~\cite{dataset2016euroc}. Besides indoor scenes, we utilize a ZED 2 stereo camera to collect outdoor scenes for extra evaluation. 
    
    Following the convention, we used the Absolute Trajectory Error (ATE) metric~\cite{grupp2017evo} to estimate the accuracy of localization, while the RMSE and STD of ATE are reported. Quantitative measurements in terms of PSNR, SSIM, and LPIPS~\cite{lpips} are adopted to analyze the performance of photorealistic mapping. We also report the requirement of computing resources by showing the tracking FPS, rendering FPS, and GPU memory usage. The evaluation regarding mesh reconstruction is out of the range of this work.
    Moreover, to lower the effect of the nondeterministic nature of multi-threading and machine-learning systems, we ran each sequence five times and reported the average results for each metric. Please refer to the supplementary for details. 

 \subsection{Results and Evaluation}\label{subsec:exp-results-evaluation}
    \noindent\textbf{On Replica}.
    As quantitative comparison demonstrated in \Tref{tab:replica},  Photo-SLAM achieves top performance in terms of mapping quality. With competitive localization accuracy, Photo-SLAM can track the camera poses in real time. Moreover, Photo-SLAM renders hundreds of photorealistic views in a resolution of 1200$\times$680 per second with less GPU memory usage. Even on the embedded platform, the rendering speed of Photo-SLAM is about 100 FPS.
    
    In monocular scenarios, Photo-SLAM significantly suppresses other methods. 
    When we disabled the depth supervision of Nice-SLAM~\cite{niceslam}, its accuracy of localization dramatically decreased while the mapping was of 16.311 PSNR. We conduct a qualitative comparison in \fref{fig:replica-mono}. The mapping results of Photo-SLAM are photorealistic.
    
    In RGB-D scenarios, we ran BundleFusion~\cite{bundlefusion} with RGB-D sequences and then extracted textured mesh. And then we used a mesh render to render corresponding images for comparison. As shown in \fref{fig:replica-rgbd}, the mesh reconstructed by the classical method is likely to be aliasing and hollow. ESLAM~\cite{eslam} and Go-SLAM~\cite{go-slam} have the best localization accuracy, but the mapping lacks high-frequency details. By contrast, Photo-SLAM can render high-fidelity images and the rendering speed is about \textbf{three hundred times faster}.
    
    %Our quantitative analyses of experiment results in the eight scenes of the Replica dataset can be found in Tab.~\ref{tab:replica}. Photo-SLAM had the highest rendering FPS and least GPU memory usage even on the Jetson Orin embedded platform. Our renderding FPS was 885 times greater than the best competitor, with a 6-GB GPU memory usage which is the same as the best competitor. The GPU memory usage can further be lowered to 4GB on the laptop or Jetson. Given this advantage, Photo-SLAM still outperformed the compared methods in terms of reconstruction quality. For example, our PSNR on Jetson is 6 more than the best RTX-4090 competitor. We also had first-class localization accuracy and tracking FPS. Qualitative comparison is provided in Fig.~\ref{fig:replica-mono}, where our method provides higher-quality rendered images.
    %\noindent\textbf{RGB-D Replica.} Quantitative results are provided in Tab.~\ref{tab:replica}. We also show qualitative comparisons in Fig.~\ref{fig:replica-rgbd}. Photo-SLAM had the lowest photometric losses and the best rendering quality, which informs that it had state-of-the-art reconstruction quality. The RMSE, STD and tracking FPS show that our localization accuracy and FPS are in the first class. More importantly, our method rendered free-view images 289 times faster than the best NeRF-based competitor, using just 1 GB more GPU memory. On the Jetson, our method used 4-GB GPU memory, the same as the best competitor, but still rendered 31 times faster than it.
    
\begin{figure}
    \centering
    \def\imw{0.3}
    \subfloat[Ground Truth]{
    \begin{minipage}{\imw\linewidth}
        \includegraphics[width=\linewidth]{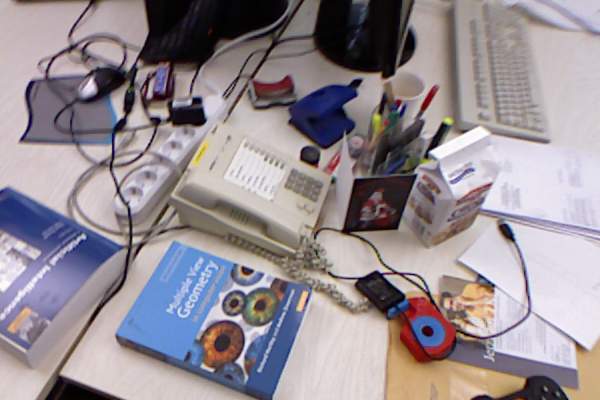}
        \includegraphics[width=\linewidth]{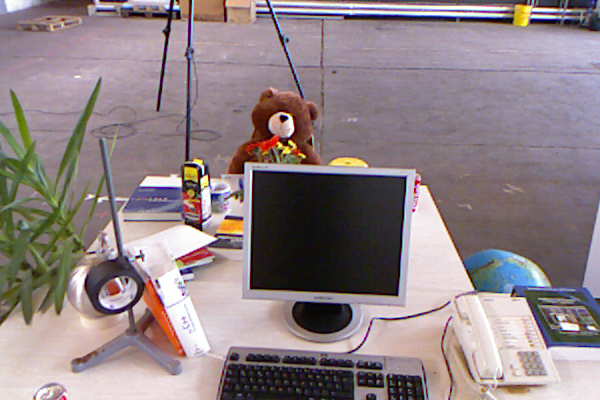}
        \includegraphics[width=\linewidth]{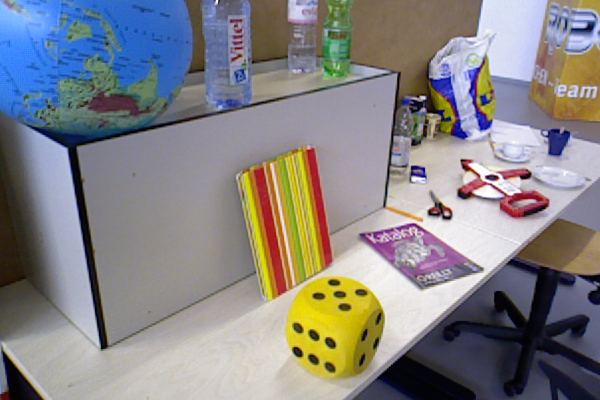}
    \end{minipage}
    }
    \subfloat[Ours (Mono)]{
    \begin{minipage}{\imw\linewidth}
        \includegraphics[width=\linewidth]{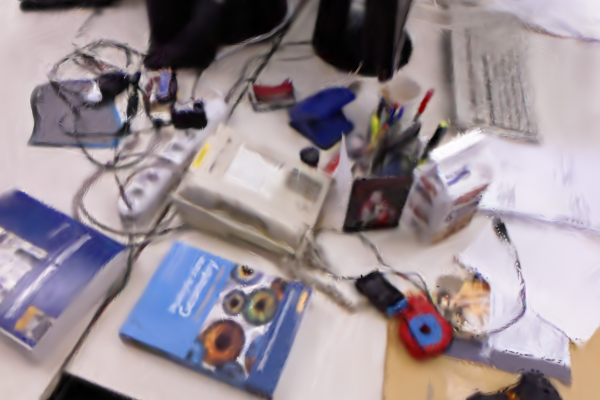}
        \includegraphics[width=\linewidth]{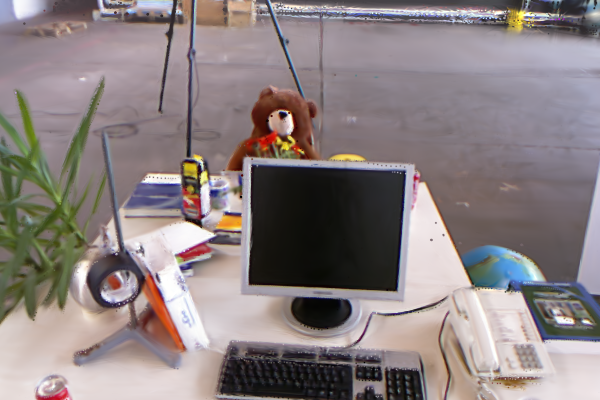}
        \includegraphics[width=\linewidth]{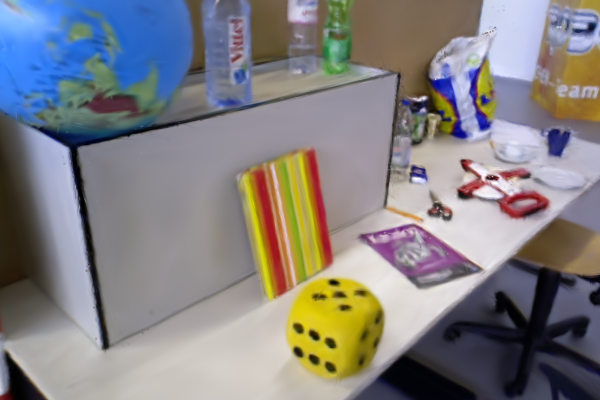}
    \end{minipage}
    }
    \subfloat[Ours (RGB-D)]{
    \begin{minipage}{\imw\linewidth}
        \includegraphics[width=\linewidth]{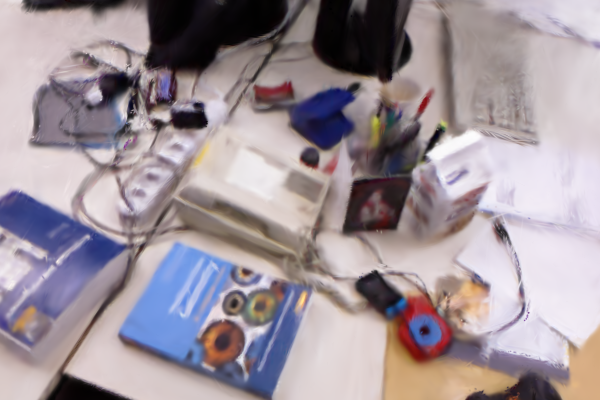}
        \includegraphics[width=\linewidth]{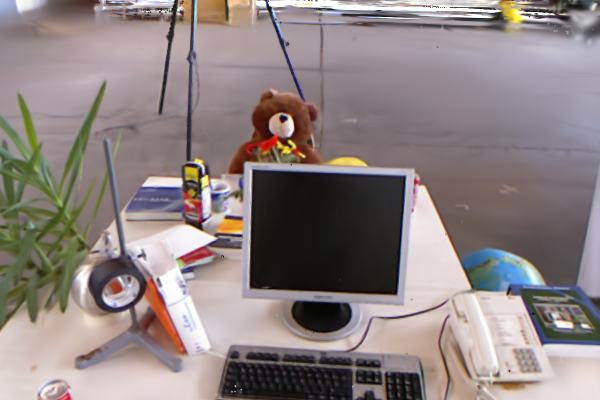}
        \includegraphics[width=\linewidth]{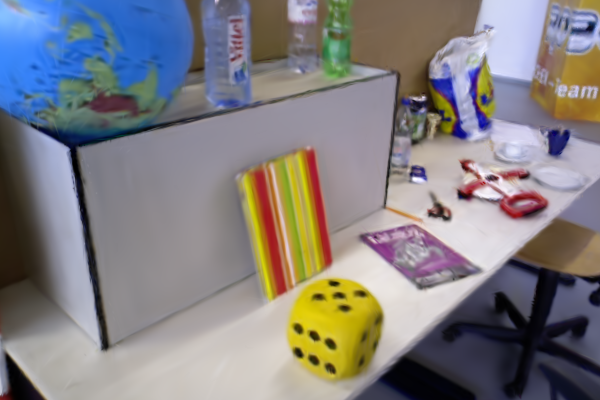}
    \end{minipage}
    }
    \vskip -0.2cm
    \caption{Qualitative results of Photo-SLAM on dataset TUM.}
    \label{fig:tum}
    %\vskip -0.2cm
\end{figure}

    \noindent\textbf{On TUM}. 
    We provide quantitative analyses on the three sequences of the TUM dataset in \Tref{tab:tum}. Compared to learning-based methods, \eg, DROID-SLAM~\cite{droid} and Go-SLAM~\cite{go-slam}, ORB-SLAM3 runs faster without the requirement of GPU and has higher accuracy regarding localization. It is shown that the classical method still has advantages in terms of robustness and generalization. \fref{fig:tum} is a gallery of Photo-SLAM mapping.
    
    %Examples of qualitative results are shown in Fig~\ref{fig:tum}. We gained top three PSNR, SSIM and LPIPS in two out of three different types of scenes, and apparently leading results on fr2-xyz, showing that Photo-SLAM achieved state-of-the-art photorealistic mapping quality. RMSE of the best competitors and that of ours were on the same order of magnitude, indicating our competitive localization accuracy.
    
\begin{figure}
        \centering
        \def\imw{0.48}
        \includegraphics[width=\imw\linewidth]{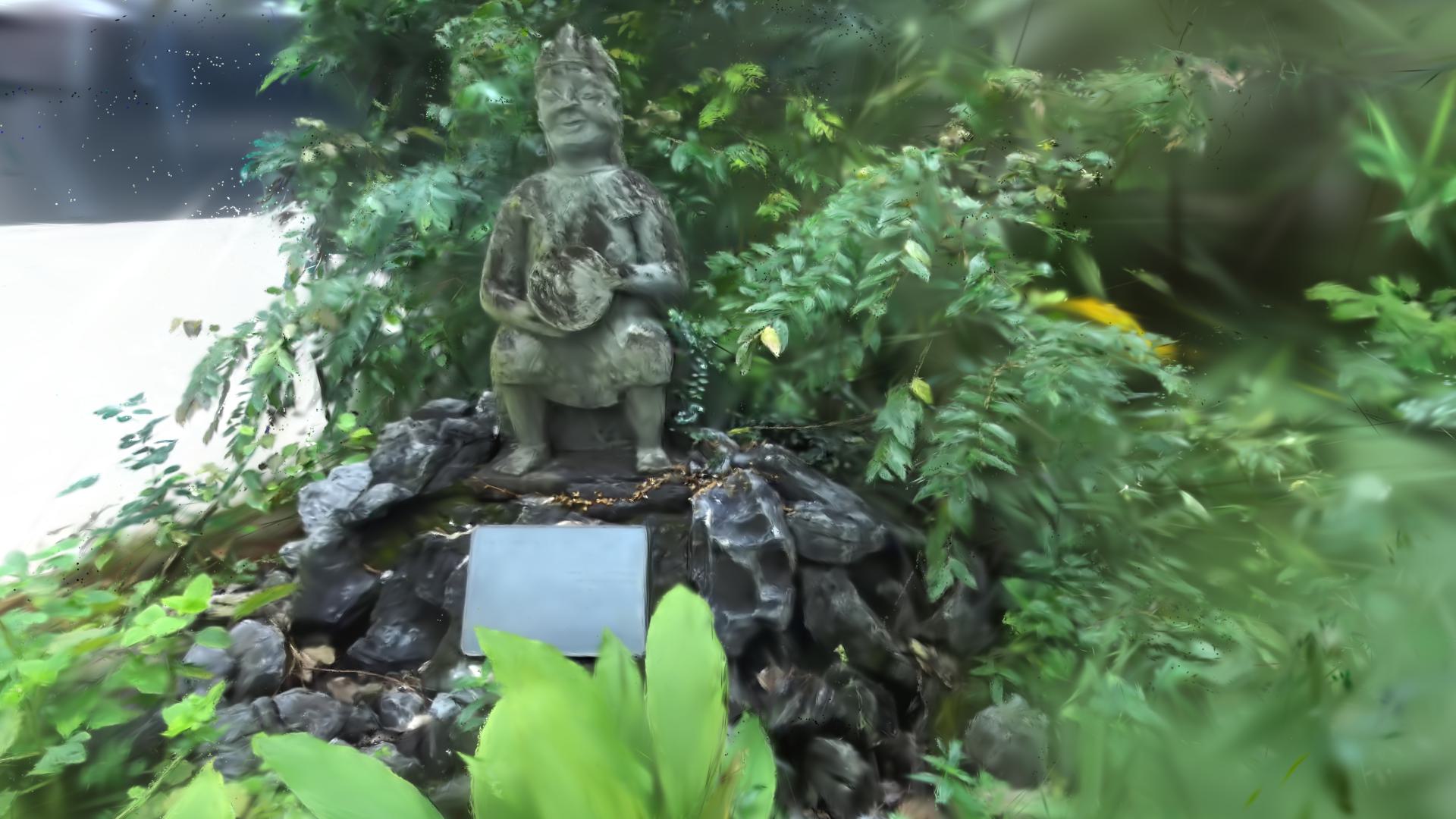}
        \includegraphics[width=\imw\linewidth]{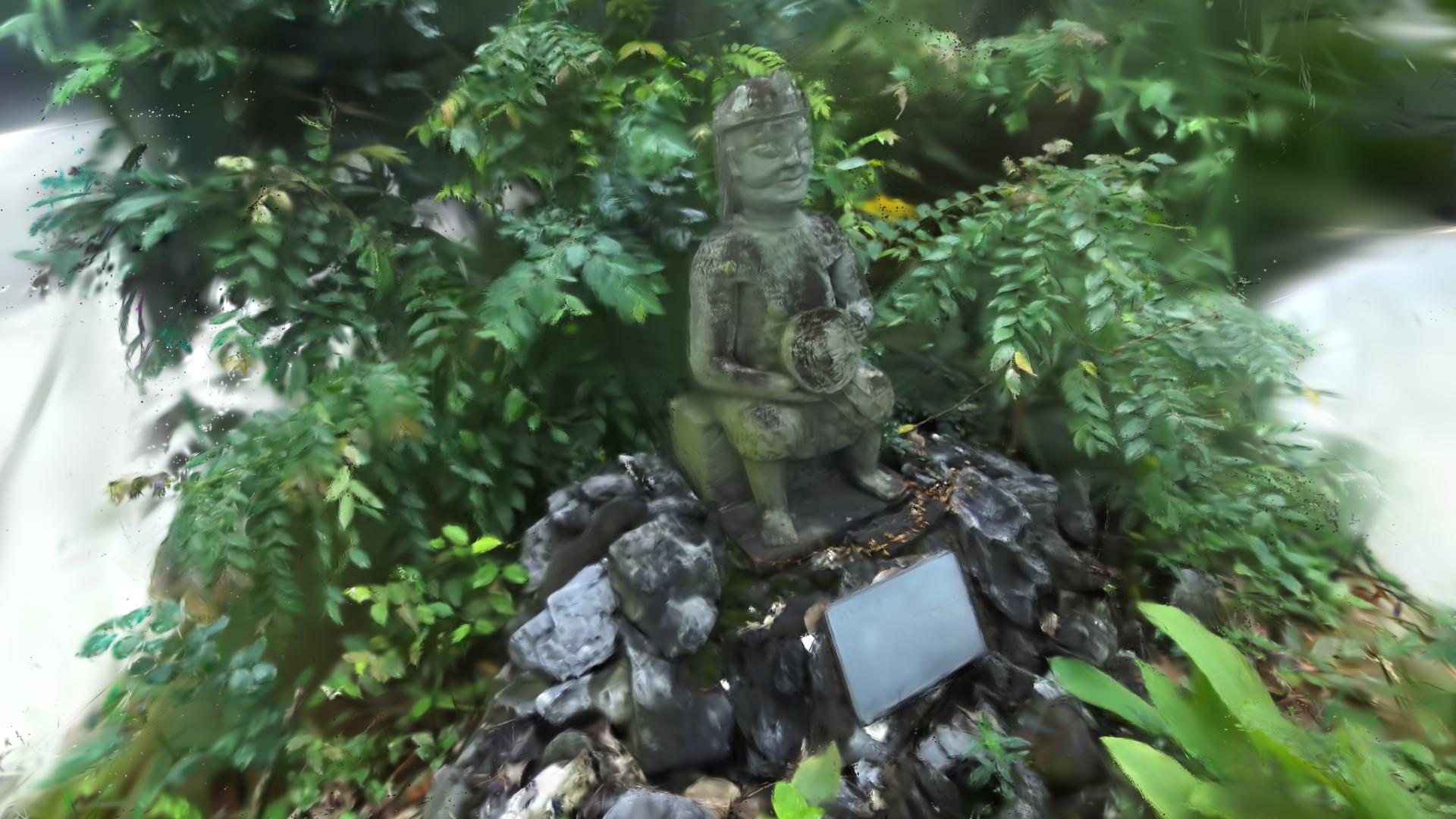}
        \includegraphics[width=\imw\linewidth]{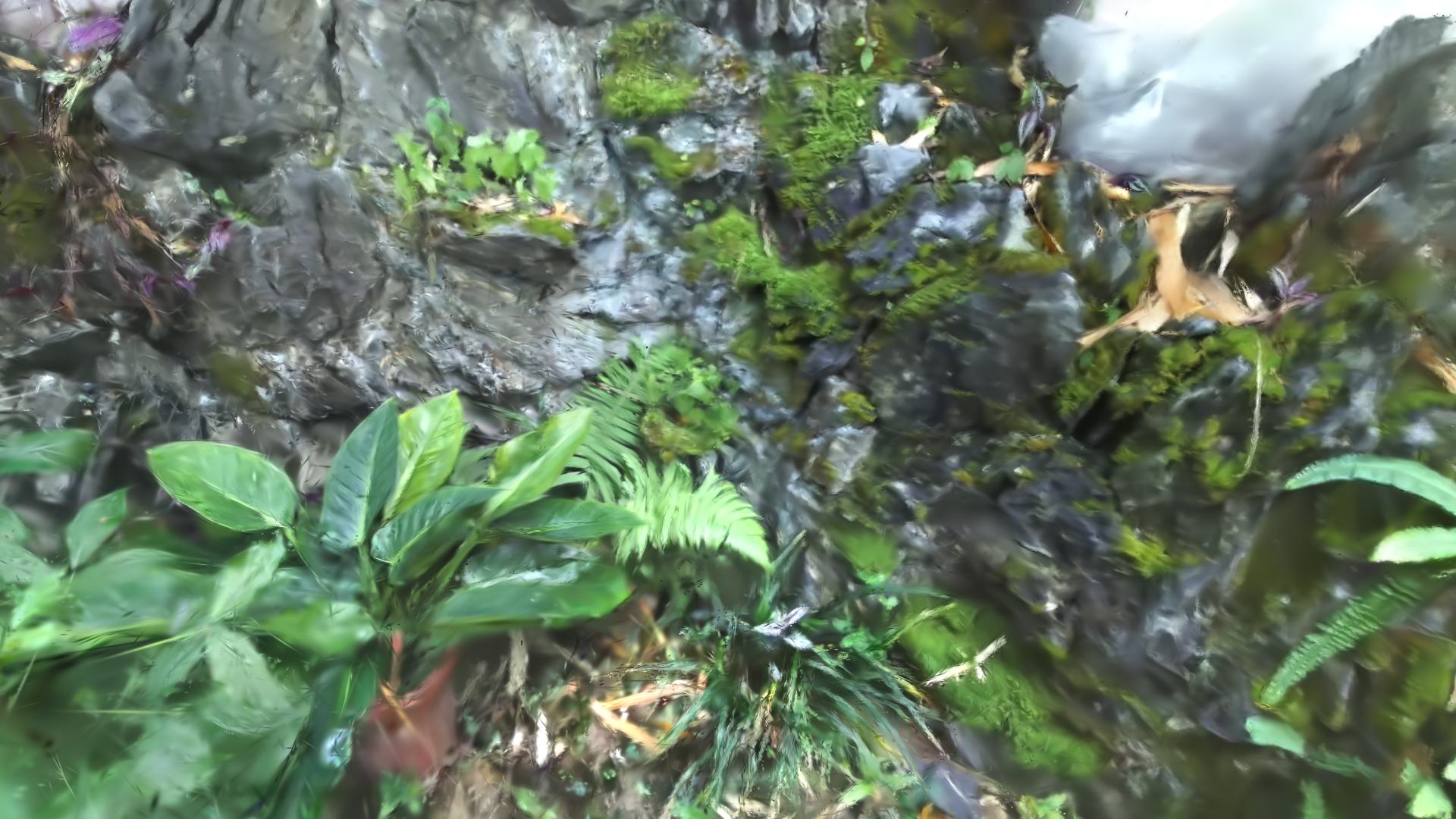}
        \includegraphics[width=\imw\linewidth]{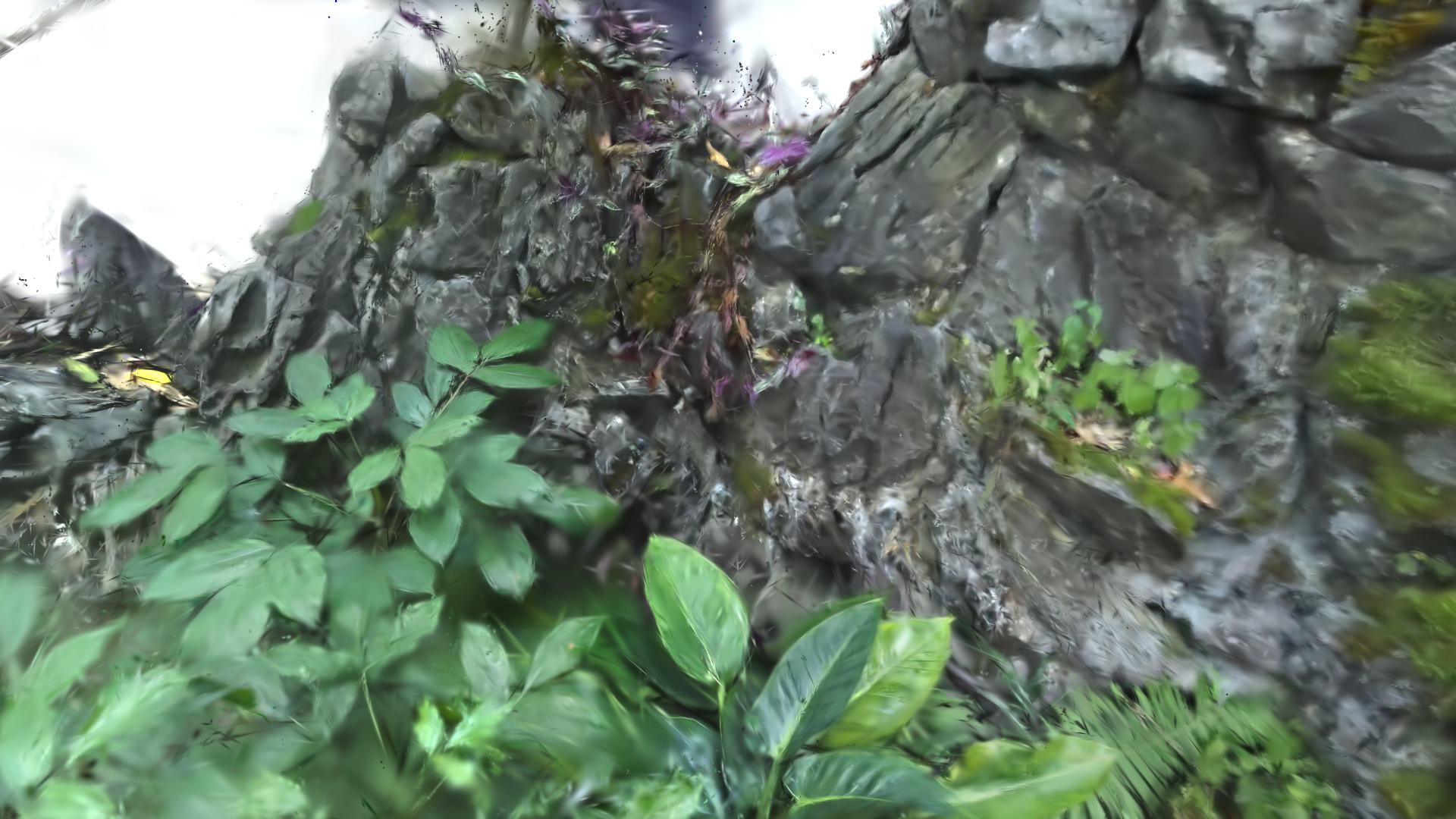}
        %\vskip -0.2cm
        \caption{Mapping results of Photo-SLAM using a hand-held stereo camera in an outdoor unbounding scene.}
        %\vskip -0.1cm
        \label{fig:ust}
    \end{figure}

    %As the results in Tab.~\ref{tab:tum} and Fig.~\ref{fig:tum} indicate, Photo-SLAM still gained top three PSNR, SSIM and LPIPS on fr1-desk, which demands the highest real-time performance because of the shortest duration. On fr2-xyz and fr3-office, we had the top localization accuracy, while gaining first-rate photorealistic mapping quality just on par with the leading Orbeez-SLAM. Our rendering speed still had an absolute advantage on the TUM dataset with both monocular and RGB-D cameras, which we put in the supplementary to save space.
    \noindent\textbf{On Stereo.} Stereo cameras can provide more robust tracking but have hardly been supported by former real-time dense SLAM systems. However, Photo-SLAM has been designed to be compatible with stereo cameras. We provide quantitative results on the EuRoC dataset in \Tref{tab:euroc} and qualitative results in supplementary. The results show that our system could still perform decently in stereo scenes.
    Further, we used a hand-held stereo camera to collect some outdoor scenes, and the mapping results of Photo-SLAM are illustrated in \fref{fig:ust}.

\begin{table}[t]
    \centering
    \footnotesize
    \tabcolsep=0.08cm
    \resizebox{0.8\linewidth}{!}{
    \begin{tabular}{cc|ccccc}
    \toprule
    \multicolumn{2}{c}{On Euroc Stereo}& \makecell{\scriptsize ORB-\\SLAM3} &\makecell{\scriptsize DROID-\\SLAM} &\textbf{\makecell{Ours\\ (Jetson)}} &\textbf{\makecell{Ours\\ (Laptop)}} &\textbf{Ours} \\
    \midrule
    \multirow{4}{*}{{MH-01}}& {\scriptsize RMSE (cm) $\downarrow$} &4.379&39.514&\markthird{4.207}&\markfirst{4.049}&\marksecond{4.109}\\
        &PSNR $\uparrow$ &-&-&\markfirst{13.979}&\cellcolor{firstcolor}\marksecond{13.962}&\markthird{13.952}\\
        &SSIM $\uparrow$ &-&-&\markfirst{0.426}&\marksecond{0.421}&\markthird{0.420}\\
        &LPIPS $\downarrow$ &-&-&\markthird{0.428}&\marksecond{0.378}&\markfirst{0.366}\\
    \midrule
    \multirow{4}{*}{{MH-02}}&{\scriptsize RMSE (cm) $\downarrow$} &\markthird{4.525}&39.265&\markfirst{4.193}&4.731&\marksecond{4.441}\\   
        &PSNR $\uparrow$ &-&-&\marksecond{14.210}&\markfirst{14.254}&\markthird{14.201}\\
        &SSIM $\uparrow$ &-&-&\markfirst{0.436}&\markfirst{0.436}&\markthird{0.430}\\
        &LPIPS $\downarrow$ &-&-&\markthird{0.447}&\marksecond{0.373}&\markfirst{0.356}\\
    \midrule
    \multirow{4}{*}{{V1-01}}&{\scriptsize RMSE (cm) $\downarrow$}  &8.940&21.646&\marksecond{8.830}&\markthird{8.836}&\markfirst{8.821}\\
        &PSNR $\uparrow$ &-&-&\markthird{16.933}&\marksecond{17.025}&\markfirst{17.069}\\
        &SSIM $\uparrow$ &-&-&\markfirst{0.626}&\marksecond{0.622}&\markthird{0.618}\\
        &LPIPS $\downarrow$ &-&-&\markthird{0.321}&\marksecond{0.284}&\markfirst{0.266}\\
    \midrule
    \multirow{4}{*}{{V2-01}}&{\scriptsize RMSE (cm) $\downarrow$}  &26.904&\markfirst{15.344}&\markthird{26.643}&26.736&\marksecond{26.609}\\
        &PSNR $\uparrow$ &-&-&\marksecond{16.038}&\markfirst{16.052}&\markthird{15.677}\\
        &SSIM $\uparrow$ &-&-&\markfirst{0.643}&\marksecond{0.635}&\markthird{0.622}\\
        &LPIPS $\downarrow$ &-&-&\markthird{0.347}&\markfirst{0.314}&\marksecond{0.323}\\
    \bottomrule
    \end{tabular}}
    \vskip -0.2cm
    \caption{Quantitative results on the EuRoC MAV dataset, using stereo inputs. Our Photo-SLAM is the first system to support online photorealistic mapping with stereo cameras.} %, which has hardly been supported by former real-time dense 3D reconstruction SLAM systems. We mark the best two results with \colorfirsttext{first} and \colorsecondtext{second}.  To the best of our knowledge, o
    \label{tab:euroc}
    \vskip -0.2cm
\end{table}

\begin{table}
    \centering
    \footnotesize
    \tabcolsep=0.07cm 
    \resizebox{0.85\linewidth}{!}{
    \begin{tabular}{ccc|cccccc}
    \toprule
    \multicolumn{3}{c}{{On Replica}}&\multicolumn{3}{c}{Mono}&\multicolumn{3}{c}{RGB-D}\\\cmidrule(lr){1-3} \cmidrule(lr){4-6} \cmidrule(lr){7-9}
    \#&Geo&GP&{\scriptsize PSNR $\uparrow$ }&\scriptsize{FPS $\uparrow$} &\scriptsize{MB} &\scriptsize{PSNR $\uparrow$ } &\scriptsize{FPS $\uparrow$ }& {\scriptsize{MB}}\\
    \midrule
    (1)&w/o &n = 2  &31.274 & \textbf{994.2}&10.742 & 33.296  &923.0 &18.199 \\
    (2)&w/ &w/o &20.002 & 353.2 &44.100 &33.696 &860.0 &31.856\\
    (3)&w/o &w/o &22.913 &645.0 &5.782 &32.551 &1010.8 &13.901\\
    (4)&w/ &n = 1 &30.903 & 803.8 &21.819 &34.634  &953.7 &31.552\\
    (5)&w/ &n = 3 &31.563 & 877.6&22.510  &33.305  &946.2&31.039\\
    \midrule
    \makecell{\scriptsize default}&w/& n = 2&\textbf{33.302} &911.3 &31.419 &\textbf{34.958}&\textbf{1084.0} &35.211\\
    \bottomrule
    \end{tabular}
    }
    %\vskip -0.2cm
    \caption{Ablation study on the effect of geometry-based densification (Geo) and Gaussian-Pyramid-based (GP) learning.}
    \label{tab:ablation}
\end{table}
\subsection{Ablation Study}\label{subsec:exp-ablation}
    We proposed geometry-based densification (Geo) and Gaussian-Pyramid-based (GP) learning to 
    boost the system performance of real-time photorealistic mapping. In this section, we constructed an ablation study to measure the efficacy of each algorithm, which can be quantified by PSNR, rendering speed (FPS), and final model size in megabytes (MB). The quantitative results are demonstrated in \Tref{tab:ablation}. 

    \begin{figure}
        \centering
        \def\imw{0.49}
        \subfloat[w/o Geo]{\label{fig:ablation-woGeo}\includegraphics[width=\imw\linewidth]{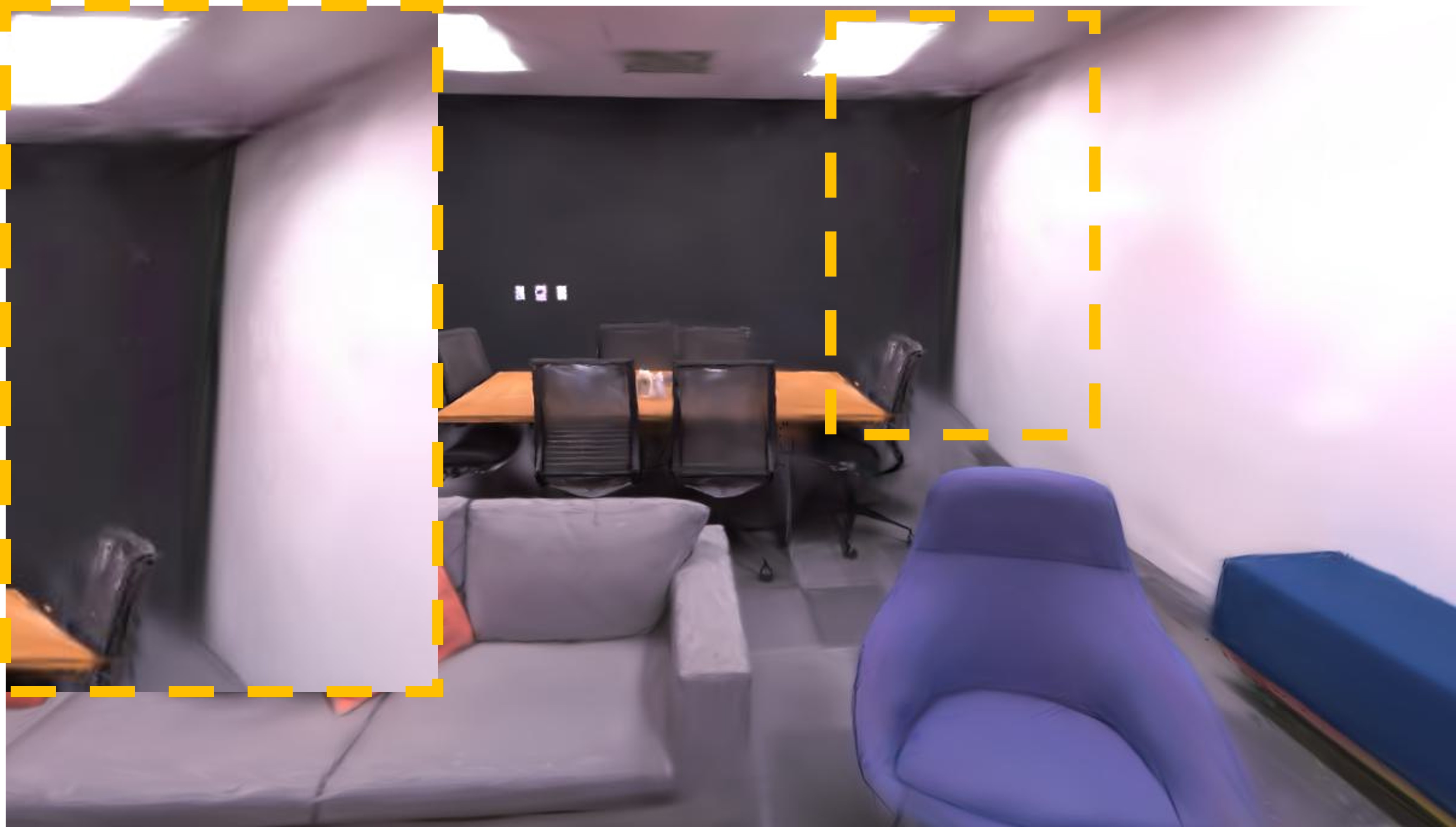}}
        \subfloat[w/o GP]{\label{fig:ablation-woGP}\includegraphics[width=\imw\linewidth]{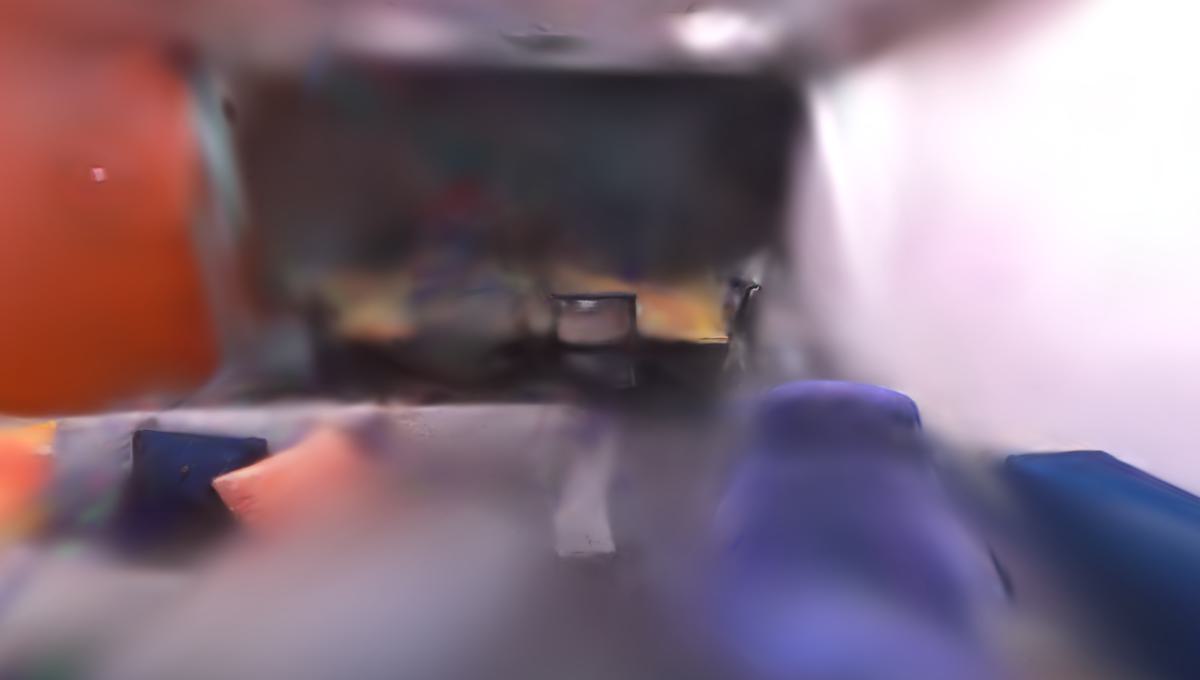}}
        %\vskip -0.2cm
        \caption{Mapping results of different ablated systems on the monocular Replica scene.}
        \label{fig:ablation}
    \end{figure}
    
    The relationship is tangled between rendering quality or speed, and the number of hyper primitives represented by the model size. 
    The models are typically small without actively densifying hyper primitives based on the inactive 2D geometric features (Geo). 
    However, without Geo, PSNR suffers degradation by 2.028 and 1.662 on monocular and RGB-D scenarios respectively, as indicated in \Tref{tab:ablation}(1). Compared to \fref{fig:replica-mono-ours}, the rendering image without Geo (\fref{fig:ablation-woGeo}) exhibits artifacts, such as on the ceiling. In RGB-D scenarios, more hyper primitives can get higher PSNR. However, without Gaussian-Pyramid-based learning, more hyper primitives are densified by Geo, and thus lead to a decrease in mapping quality and rendering speed especially in monocular scenarios, as visualized in \fref{fig:ablation-woGP} and reported in \Tref{tab:ablation}(2) and (3).
    This is because densified hyper primitives without precise depth information have inaccurate positions. Without thorough optimization, inaccurate hyper primitives become encumbrances. 
    It is noticeable that the systems with GP learning can generally perform better, highlighting the effectiveness of GP learning.
    Additionally, increasing the Gaussian pyramid levels improves mapping quality, and our Photo-SLAM with default 3-level GP learning achieves the best results. 
    However, we found that the results of using a Gaussian pyramid with 4 levels deteriorated, as shown in \Tref{tab:ablation}(5), possibly due to overfitting low-level features during incremental mapping.
    Since the image is rendered by splatting hyper primitives, the rendering speed theoretically is correlated to the number of visible hyper primitives in the current view rather than the model size of the whole scene. 
    Although a smaller model does not necessarily imply a higher average rendering speed, a precise reconstructed model should accurately capture its essential details while using concise parameters, which is the premise for high-speed rendering.
    Moreover, reducing the time required for rendering enables the optimization of hyper primitives with more iterations during online mapping, ultimately leading to improved accuracy and quality. % in the reconstructed model.

    In conclusion, the ablation study verifies that the geometry-based densification strategy allows the system to obtain sufficient hyper primitives while the Gaussian-Pyramid-based learning guarantees hyper primitives optimized thoroughly, enhancing online photorealistic mapping performance. There is no doubt that the default Photo-SLAM can reconstruct a map with more appropriate hyper primitives, and achieve better rendering quality and high rendering speed than ablated systems.
        
    %When we disable Gaussian-Pyramid-based learning, the mapping quality and rendering speed of monocular cameras decrease significantly, as illustrated in \fref{fig:ablation-woGP}. 

    %Additionally, the relationship between final model size and rendering quality, or speed is relatively complex. 
    %When the initial positions of primitives generally are proper, i.e., RGB-D scenarios, more hyper primitives represented by model size can get higher PSNR. 

    %It is obvious that the hyper primitives densified by Geo would impact the rendering quality and speed without GP learning, particularly for monocular scenarios
    
    %However, it is

    %and GP learning enable the system to effectively and efficiently learn hyper primitive.
    %densifying the hyper primitives based on geometry.
%The rendering speed is relatively higher due to the sparser hyper primitives map.

    %We isolated the choices we made on our GP learning strategy to test the rationality of it. To be specific, we constructed a series of experiments to measure the effect of different number of pyramid levels on the photorealistic mapping quality, which can be quantified by PSNR, SSIM and LPIPS. The results are provided in Tab.~\ref{tab:ablation}. The effect of GP learning was little in RGB-D scenes. However, when it was completely disabled (i.e. set to level 0), the results deteriorated quickly. When using a pyramid with more or less levels (i.e. using level 1 or 3), the deterioration is still apparent. Altogether, a level-2 GP learning can significantly improve the photorealistic mapping quality in monocular scenes.

\section{Conclusion}\label{sec:conclusion}

In this paper, we have proposed a novel SLAM framework called Photo-SLAM for simultaneous localization and photorealistic mapping. 
Instead of highly relying on resource-intensive implicit representations and neural solvers, we introduced a hyper primitives map. It enables our system to leverage explicit geometric features for localization and implicitly capture the texture information of the scenes. In addition to geometry-based densification, we proposed Gaussian-Pyramid-based learning, a new progressive training method, to further enhance mapping performance. %resulting in high-quality visual reconstruction. 
Extensive experiments have demonstrated that Photo-SLAM significantly outperforms existing SOTA SLAMs for online photorealistic mapping.
Furthermore, our system verifies its practicality by achieving real-time performance on an embedded platform, highlighting its potential for advanced robotics applications in real-world scenarios.

\noindent\textbf{Acknowledgements:}
{This work is partially supported by the Innovation and Technology Support Programme of the Innovation and Technology Fund (Ref: ITS/200/20FP), the Marine Conservation Enhancement Fund (MCEF20107 \& MCEF23EG01), and an internal grant from HKUST (R9429).}
{
    \small
    \bibliographystyle{ieeenat_fullname}
    \bibliography{main}

\begin{thebibliography}{46}
\providecommand{\natexlab}[1]{#1}
\providecommand{\url}[1]{\texttt{#1}}
\expandafter\ifx\csname urlstyle\endcsname\relax
  \providecommand{\doi}[1]{doi: #1}\else
  \providecommand{\doi}{doi: \begingroup \urlstyle{rm}\Url}\fi

\bibitem[Burri et~al.(2016)Burri, Nikolic, Gohl, Schneider, Rehder, Omari,
  Achtelik, and Siegwart]{dataset2016euroc}
Michael Burri, Janosch Nikolic, Pascal Gohl, Thomas Schneider, Joern Rehder,
  Sammy Omari, Markus~W Achtelik, and Roland Siegwart.
\newblock The euroc micro aerial vehicle datasets.
\newblock \emph{The International Journal of Robotics Research}, 2016.

\bibitem[Campos et~al.(2021)Campos, Elvira, Rodr{\'\i}guez, Montiel, and
  Tard{\'o}s]{campos2021orb3}
Carlos Campos, Richard Elvira, Juan J~G{\'o}mez Rodr{\'\i}guez, Jos{\'e}~MM
  Montiel, and Juan~D Tard{\'o}s.
\newblock Orb-slam3: An accurate open-source library for visual,
  visual--inertial, and multimap slam.
\newblock \emph{IEEE Transactions on Robotics}, 37\penalty0 (6):\penalty0
  1874--1890, 2021.

\bibitem[Chen et~al.(2022)Chen, Xu, Geiger, Yu, and Su]{TensoRF}
Anpei Chen, Zexiang Xu, Andreas Geiger, Jingyi Yu, and Hao Su.
\newblock Tensorf: Tensorial radiance fields.
\newblock In \emph{European Conference on Computer Vision (ECCV)}, 2022.

\bibitem[Chung et~al.(2023)Chung, Tseng, Hsu, Shi, Hua, Yeh, Chen, Chen, and
  Hsu]{chung2023orbeez}
Chi-Ming Chung, Yang-Che Tseng, Ya-Ching Hsu, Xiang-Qian Shi, Yun-Hung Hua,
  Jia-Fong Yeh, Wen-Chin Chen, Yi-Ting Chen, and Winston~H Hsu.
\newblock Orbeez-slam: A real-time monocular visual slam with orb features and
  nerf-realized mapping.
\newblock In \emph{2023 IEEE International Conference on Robotics and
  Automation (ICRA)}, pages 9400--9406. IEEE, 2023.

\bibitem[Curless and Levoy(1996)]{tsdf}
Brian Curless and Marc Levoy.
\newblock A volumetric method for building complex models from range images.
\newblock In \emph{Proceedings of the 23rd annual conference on Computer
  graphics and interactive techniques}, pages 303--312, 1996.

\bibitem[Dai et~al.(2017)Dai, Nie{\ss}ner, Zollh{\"o}fer, Izadi, and
  Theobalt]{bundlefusion}
Angela Dai, Matthias Nie{\ss}ner, Michael Zollh{\"o}fer, Shahram Izadi, and
  Christian Theobalt.
\newblock Bundlefusion: Real-time globally consistent 3d reconstruction using
  on-the-fly surface reintegration.
\newblock \emph{ACM Transactions on Graphics (ToG)}, 36\penalty0 (4):\penalty0
  1, 2017.

\bibitem[Engel et~al.(2014)Engel, Sch{\"o}ps, and Cremers]{lsd-slam}
Jakob Engel, Thomas Sch{\"o}ps, and Daniel Cremers.
\newblock Lsd-slam: Large-scale direct monocular slam.
\newblock In \emph{European conference on computer vision}, pages 834--849.
  Springer, 2014.

\bibitem[Engel et~al.(2017)Engel, Koltun, and Cremers]{dso}
Jakob Engel, Vladlen Koltun, and Daniel Cremers.
\newblock Direct sparse odometry.
\newblock \emph{IEEE transactions on pattern analysis and machine
  intelligence}, 40\penalty0 (3):\penalty0 611--625, 2017.

\bibitem[Forster et~al.(2014)Forster, Pizzoli, and Scaramuzza]{svo}
Christian Forster, Matia Pizzoli, and Davide Scaramuzza.
\newblock Svo: Fast semi-direct monocular visual odometry.
\newblock In \emph{2014 IEEE international conference on robotics and
  automation (ICRA)}, pages 15--22. IEEE, 2014.

\bibitem[Fridovich-Keil et~al.(2022)Fridovich-Keil, Yu, Tancik, Chen, Recht,
  and Kanazawa]{fridovich2022plenoxels}
Sara Fridovich-Keil, Alex Yu, Matthew Tancik, Qinhong Chen, Benjamin Recht, and
  Angjoo Kanazawa.
\newblock Plenoxels: Radiance fields without neural networks.
\newblock In \emph{Proceedings of the IEEE/CVF Conference on Computer Vision
  and Pattern Recognition}, pages 5501--5510, 2022.

\bibitem[G\'alvez-L\'opez and Tard\'os(2012)]{GalvezTRO2012DBOW2}
Dorian G\'alvez-L\'opez and J.~D. Tard\'os.
\newblock Bags of binary words for fast place recognition in image sequences.
\newblock \emph{IEEE Transactions on Robotics}, 28\penalty0 (5):\penalty0
  1188--1197, 2012.

\bibitem[Godard et~al.(2017)Godard, Mac~Aodha, and Brostow]{monodepth}
Cl{\'e}ment Godard, Oisin Mac~Aodha, and Gabriel~J Brostow.
\newblock Unsupervised monocular depth estimation with left-right consistency.
\newblock In \emph{Proceedings of the IEEE conference on computer vision and
  pattern recognition}, pages 270--279, 2017.

\bibitem[Grupp(2017)]{grupp2017evo}
Michael Grupp.
\newblock evo: Python package for the evaluation of odometry and slam.
\newblock \url{https://github.com/MichaelGrupp/evo}, 2017.

\bibitem[Huang et~al.(2022)Huang, Chen, Zhang, and Yeung]{360roam}
Huajian Huang, Yingshu Chen, Tianjian Zhang, and Sai-Kit Yeung.
\newblock 360roam: Real-time indoor roaming using geometry-aware ${360^\circ}$
  radiance fields.
\newblock \emph{arXiv preprint arXiv:2208.02705}, 2022.

\bibitem[Izadi et~al.(2011)Izadi, Kim, Hilliges, Molyneaux, Newcombe, Kohli,
  Shotton, Hodges, Freeman, Davison, et~al.]{kinectfusion}
Shahram Izadi, David Kim, Otmar Hilliges, David Molyneaux, Richard Newcombe,
  Pushmeet Kohli, Jamie Shotton, Steve Hodges, Dustin Freeman, Andrew Davison,
  et~al.
\newblock Kinectfusion: real-time 3d reconstruction and interaction using a
  moving depth camera.
\newblock In \emph{Proceedings of the 24th annual ACM symposium on User
  interface software and technology}, pages 559--568, 2011.

\bibitem[Johari et~al.(2023)Johari, Carta, and Fleuret]{eslam}
Mohammad~Mahdi Johari, Camilla Carta, and Fran{\c{c}}ois Fleuret.
\newblock Eslam: Efficient dense slam system based on hybrid representation of
  signed distance fields.
\newblock In \emph{Proceedings of the IEEE/CVF Conference on Computer Vision
  and Pattern Recognition}, pages 17408--17419, 2023.

\bibitem[Kendall et~al.(2015)Kendall, Grimes, and Cipolla]{posenet}
Alex Kendall, Matthew Grimes, and Roberto Cipolla.
\newblock Posenet: A convolutional network for real-time 6-dof camera
  relocalization.
\newblock In \emph{Proceedings of the IEEE international conference on computer
  vision}, pages 2938--2946, 2015.

\bibitem[Kerbl et~al.(2023)Kerbl, Kopanas, Leimk{\"u}hler, and
  Drettakis]{kerbl20233dgaussiansplatting}
Bernhard Kerbl, Georgios Kopanas, Thomas Leimk{\"u}hler, and George Drettakis.
\newblock 3d gaussian splatting for real-time radiance field rendering.
\newblock \emph{ACM Transactions on Graphics (ToG)}, 42\penalty0 (4):\penalty0
  1--14, 2023.

\bibitem[Li et~al.(2023)Li, M{\"u}ller, Evans, Taylor, Unberath, Liu, and
  Lin]{neuralangelo}
Zhaoshuo Li, Thomas M{\"u}ller, Alex Evans, Russell~H Taylor, Mathias Unberath,
  Ming-Yu Liu, and Chen-Hsuan Lin.
\newblock Neuralangelo: High-fidelity neural surface reconstruction.
\newblock In \emph{Proceedings of the IEEE/CVF Conference on Computer Vision
  and Pattern Recognition}, pages 8456--8465, 2023.

\bibitem[Liu et~al.(2020)Liu, Gu, Zaw~Lin, Chua, and Theobalt]{nsvf}
Lingjie Liu, Jiatao Gu, Kyaw Zaw~Lin, Tat-Seng Chua, and Christian Theobalt.
\newblock Neural sparse voxel fields.
\newblock \emph{Advances in Neural Information Processing Systems},
  33:\penalty0 15651--15663, 2020.

\bibitem[Mildenhall et~al.(2020)Mildenhall, Srinivasan, Tancik, Barron,
  Ramamoorthi, and Ng]{nerf}
Ben Mildenhall, Pratul~P Srinivasan, Matthew Tancik, Jonathan~T Barron, Ravi
  Ramamoorthi, and Ren Ng.
\newblock Nerf: Representing scenes as neural radiance fields for view
  synthesis.
\newblock In \emph{European conference on computer vision}, pages 405--421.
  Springer, 2020.

\bibitem[M{\"u}ller et~al.(2022)M{\"u}ller, Evans, Schied, and
  Keller]{muller2022instant}
Thomas M{\"u}ller, Alex Evans, Christoph Schied, and Alexander Keller.
\newblock Instant neural graphics primitives with a multiresolution hash
  encoding.
\newblock \emph{ACM Transactions on Graphics (ToG)}, 41\penalty0 (4):\penalty0
  1--15, 2022.

\bibitem[Mur-Artal and Tard{\'o}s(2017)]{mur2017orb2}
Raul Mur-Artal and Juan~D Tard{\'o}s.
\newblock Orb-slam2: An open-source slam system for monocular, stereo, and
  rgb-d cameras.
\newblock \emph{IEEE transactions on robotics}, 33\penalty0 (5):\penalty0
  1255--1262, 2017.

\bibitem[Mur-Artal et~al.(2015)Mur-Artal, Montiel, and Tardos]{mur2015orb}
Raul Mur-Artal, Jose Maria~Martinez Montiel, and Juan~D Tardos.
\newblock Orb-slam: a versatile and accurate monocular slam system.
\newblock \emph{IEEE transactions on robotics}, 31\penalty0 (5):\penalty0
  1147--1163, 2015.

\bibitem[Prisacariu et~al.(2014)Prisacariu, Kahler, Cheng, Ren, Valentin, Torr,
  Reid, and Murray]{infinitam}
V.~A. Prisacariu, O. Kahler, M.~M. Cheng, C.~Y. Ren, J. Valentin, P.~H.~S.
  Torr, I.~D. Reid, and D.~W. Murray.
\newblock {A Framework for the Volumetric Integration of Depth Images}.
\newblock \emph{ArXiv e-prints}, 2014.

\bibitem[Rublee et~al.(2011)Rublee, Rabaud, Konolige, and Bradski]{ORB}
E. Rublee, V. Rabaud, K. Konolige, and G. Bradski.
\newblock Orb: An efficient alternative to sift or surf.
\newblock \emph{IEEE International Conference on Computer Vision}, 58\penalty0
  (11):\penalty0 2564--2571, 2011.

\bibitem[Sandström et~al.(2023)Sandström, Li, Van~Gool, and
  R.~Oswald]{point-slam}
Erik Sandström, Yue Li, Luc Van~Gool, and Martin R.~Oswald.
\newblock Point-slam: Dense neural point cloud-based slam.
\newblock In \emph{Proceedings of the IEEE/CVF International Conference on
  Computer Vision (ICCV)}, 2023.

\bibitem[Straub et~al.(2019)Straub, Whelan, Ma, Chen, Wijmans, Green, Engel,
  Mur-Artal, Ren, Verma, Clarkson, Yan, Budge, Yan, Pan, Yon, Zou, Leon,
  Carter, Briales, Gillingham, Mueggler, Pesqueira, Savva, Batra, Strasdat,
  Nardi, Goesele, Lovegrove, and Newcombe]{dataset2019replica}
Julian Straub, Thomas Whelan, Lingni Ma, Yufan Chen, Erik Wijmans, Simon Green,
  Jakob~J. Engel, Raul Mur-Artal, Carl Ren, Shobhit Verma, Anton Clarkson,
  Mingfei Yan, Brian Budge, Yajie Yan, Xiaqing Pan, June Yon, Yuyang Zou,
  Kimberly Leon, Nigel Carter, Jesus Briales, Tyler Gillingham, Elias Mueggler,
  Luis Pesqueira, Manolis Savva, Dhruv Batra, Hauke~M. Strasdat, Renzo~De
  Nardi, Michael Goesele, Steven Lovegrove, and Richard Newcombe.
\newblock The {R}eplica dataset: A digital replica of indoor spaces.
\newblock \emph{arXiv preprint arXiv:1906.05797}, 2019.

\bibitem[Sturm et~al.(2012)Sturm, Engelhard, Endres, Burgard, and
  Cremers]{dataset2012tum}
J. Sturm, N. Engelhard, F. Endres, W. Burgard, and D. Cremers.
\newblock A benchmark for the evaluation of rgb-d slam systems.
\newblock In \emph{Proc. of the International Conference on Intelligent Robot
  Systems (IROS)}, 2012.

\bibitem[Sucar et~al.(2021)Sucar, Liu, Ortiz, and Davison]{imap}
Edgar Sucar, Shikun Liu, Joseph Ortiz, and Andrew~J Davison.
\newblock imap: Implicit mapping and positioning in real-time.
\newblock In \emph{Proceedings of the IEEE/CVF International Conference on
  Computer Vision}, pages 6229--6238, 2021.

\bibitem[Sun et~al.(2022)Sun, Sun, and Chen]{dvgo}
Cheng Sun, Min Sun, and Hwann-Tzong Chen.
\newblock Direct voxel grid optimization: Super-fast convergence for radiance
  fields reconstruction.
\newblock In \emph{Proceedings of the IEEE/CVF Conference on Computer Vision
  and Pattern Recognition}, pages 5459--5469, 2022.

\bibitem[Takikawa et~al.(2021)Takikawa, Litalien, Yin, Kreis, Loop,
  Nowrouzezahrai, Jacobson, McGuire, and Fidler]{nglod}
Towaki Takikawa, Joey Litalien, Kangxue Yin, Karsten Kreis, Charles Loop, Derek
  Nowrouzezahrai, Alec Jacobson, Morgan McGuire, and Sanja Fidler.
\newblock Neural geometric level of detail: Real-time rendering with implicit
  3d shapes.
\newblock In \emph{Proceedings of the IEEE/CVF Conference on Computer Vision
  and Pattern Recognition}, pages 11358--11367, 2021.

\bibitem[Teed and Deng(2020)]{raft}
Zachary Teed and Jia Deng.
\newblock Raft: Recurrent all-pairs field transforms for optical flow.
\newblock In \emph{Computer Vision--ECCV 2020: 16th European Conference,
  Glasgow, UK, August 23--28, 2020, Proceedings, Part II 16}, pages 402--419.
  Springer, 2020.

\bibitem[Teed and Deng(2021)]{droid}
Zachary Teed and Jia Deng.
\newblock Droid-slam: Deep visual slam for monocular, stereo, and rgb-d
  cameras.
\newblock \emph{Advances in neural information processing systems},
  34:\penalty0 16558--16569, 2021.

\bibitem[Tewari et~al.(2022)Tewari, Thies, Mildenhall, Srinivasan, Tretschk,
  Yifan, Lassner, Sitzmann, Martin-Brualla, Lombardi,
  et~al.]{advancesNueralRendering}
Ayush Tewari, Justus Thies, Ben Mildenhall, Pratul Srinivasan, Edgar Tretschk,
  Wang Yifan, Christoph Lassner, Vincent Sitzmann, Ricardo Martin-Brualla,
  Stephen Lombardi, et~al.
\newblock Advances in neural rendering.
\newblock In \emph{Computer Graphics Forum}, pages 703--735. Wiley Online
  Library, 2022.

\bibitem[Wang et~al.(2023)Wang, Wang, and Agapito]{coslam}
Hengyi Wang, Jingwen Wang, and Lourdes Agapito.
\newblock Co-slam: Joint coordinate and sparse parametric encodings for neural
  real-time slam.
\newblock In \emph{Proceedings of the IEEE/CVF Conference on Computer Vision
  and Pattern Recognition}, pages 13293--13302, 2023.

\bibitem[Wang et~al.(2020)Wang, Zhu, Wang, Hu, Qiu, Wang, Hu, Kapoor, and
  Scherer]{tartanair}
Wenshan Wang, Delong Zhu, Xiangwei Wang, Yaoyu Hu, Yuheng Qiu, Chen Wang, Yafei
  Hu, Ashish Kapoor, and Sebastian Scherer.
\newblock Tartanair: A dataset to push the limits of visual slam.
\newblock In \emph{2020 IEEE/RSJ International Conference on Intelligent Robots
  and Systems (IROS)}, pages 4909--4916. IEEE, 2020.

\bibitem[Wizadwongsa et~al.(2021)Wizadwongsa, Phongthawee, Yenphraphai, and
  Suwajanakorn]{nex}
Suttisak Wizadwongsa, Pakkapon Phongthawee, Jiraphon Yenphraphai, and Supasorn
  Suwajanakorn.
\newblock Nex: Real-time view synthesis with neural basis expansion.
\newblock In \emph{Proceedings of the IEEE/CVF Conference on Computer Vision
  and Pattern Recognition}, pages 8534--8543, 2021.

\bibitem[Xiangli et~al.(2022)Xiangli, Xu, Pan, Zhao, Rao, Theobalt, Dai, and
  Lin]{bungeenerf}
Yuanbo Xiangli, Linning Xu, Xingang Pan, Nanxuan Zhao, Anyi Rao, Christian
  Theobalt, Bo Dai, and Dahua Lin.
\newblock Bungeenerf: Progressive neural radiance field for extreme multi-scale
  scene rendering.
\newblock In \emph{European conference on computer vision}, pages 106--122.
  Springer, 2022.

\bibitem[Xie et~al.(2022)Xie, Takikawa, Saito, Litany, Yan, Khan, Tombari,
  Tompkin, Sitzmann, and Sridhar]{neuralrendering}
Yiheng Xie, Towaki Takikawa, Shunsuke Saito, Or Litany, Shiqin Yan, Numair
  Khan, Federico Tombari, James Tompkin, Vincent Sitzmann, and Srinath Sridhar.
\newblock Neural fields in visual computing and beyond.
\newblock In \emph{Computer Graphics Forum}, pages 641--676. Wiley Online
  Library, 2022.

\bibitem[Yang et~al.(2020)Yang, Stumberg, Wang, and Cremers]{d3vo}
Nan Yang, Lukas~von Stumberg, Rui Wang, and Daniel Cremers.
\newblock D3vo: Deep depth, deep pose and deep uncertainty for monocular visual
  odometry.
\newblock In \emph{Proceedings of the IEEE/CVF conference on computer vision
  and pattern recognition}, pages 1281--1292, 2020.

\bibitem[Yu et~al.(2021)Yu, Li, Tancik, Li, Ng, and Kanazawa]{plenoctrees}
Alex Yu, Ruilong Li, Matthew Tancik, Hao Li, Ren Ng, and Angjoo Kanazawa.
\newblock Plenoctrees for real-time rendering of neural radiance fields.
\newblock In \emph{Proceedings of the IEEE/CVF International Conference on
  Computer Vision}, pages 5752--5761, 2021.

\bibitem[Zhang et~al.(2018)Zhang, Isola, Efros, Shechtman, and Wang]{lpips}
Richard Zhang, Phillip Isola, Alexei~A Efros, Eli Shechtman, and Oliver Wang.
\newblock The unreasonable effectiveness of deep features as a perceptual
  metric.
\newblock In \emph{Proceedings of the IEEE conference on computer vision and
  pattern recognition}, pages 586--595, 2018.

\bibitem[Zhang et~al.(2023)Zhang, Tosi, Mattoccia, and Poggi]{go-slam}
Youmin Zhang, Fabio Tosi, Stefano Mattoccia, and Matteo Poggi.
\newblock Go-slam: Global optimization for consistent 3d instant
  reconstruction.
\newblock In \emph{Proceedings of the IEEE/CVF International Conference on
  Computer Vision}, pages 3727--3737, 2023.

\bibitem[Zhou et~al.(2018)Zhou, Ummenhofer, and Brox]{deeptam}
Huizhong Zhou, Benjamin Ummenhofer, and Thomas Brox.
\newblock Deeptam: Deep tracking and mapping.
\newblock In \emph{Proceedings of the European conference on computer vision
  (ECCV)}, pages 822--838, 2018.

\bibitem[Zhu et~al.(2022)Zhu, Peng, Larsson, Xu, Bao, Cui, Oswald, and
  Pollefeys]{niceslam}
Zihan Zhu, Songyou Peng, Viktor Larsson, Weiwei Xu, Hujun Bao, Zhaopeng Cui,
  Martin~R Oswald, and Marc Pollefeys.
\newblock Nice-slam: Neural implicit scalable encoding for slam.
\newblock In \emph{Proceedings of the IEEE/CVF Conference on Computer Vision
  and Pattern Recognition}, pages 12786--12796, 2022.

\end{thebibliography}
}

\clearpage
\setcounter{page}{1}
\maketitlesupplementary

\begin{figure}
    \centering
    \includegraphics[width=\linewidth]{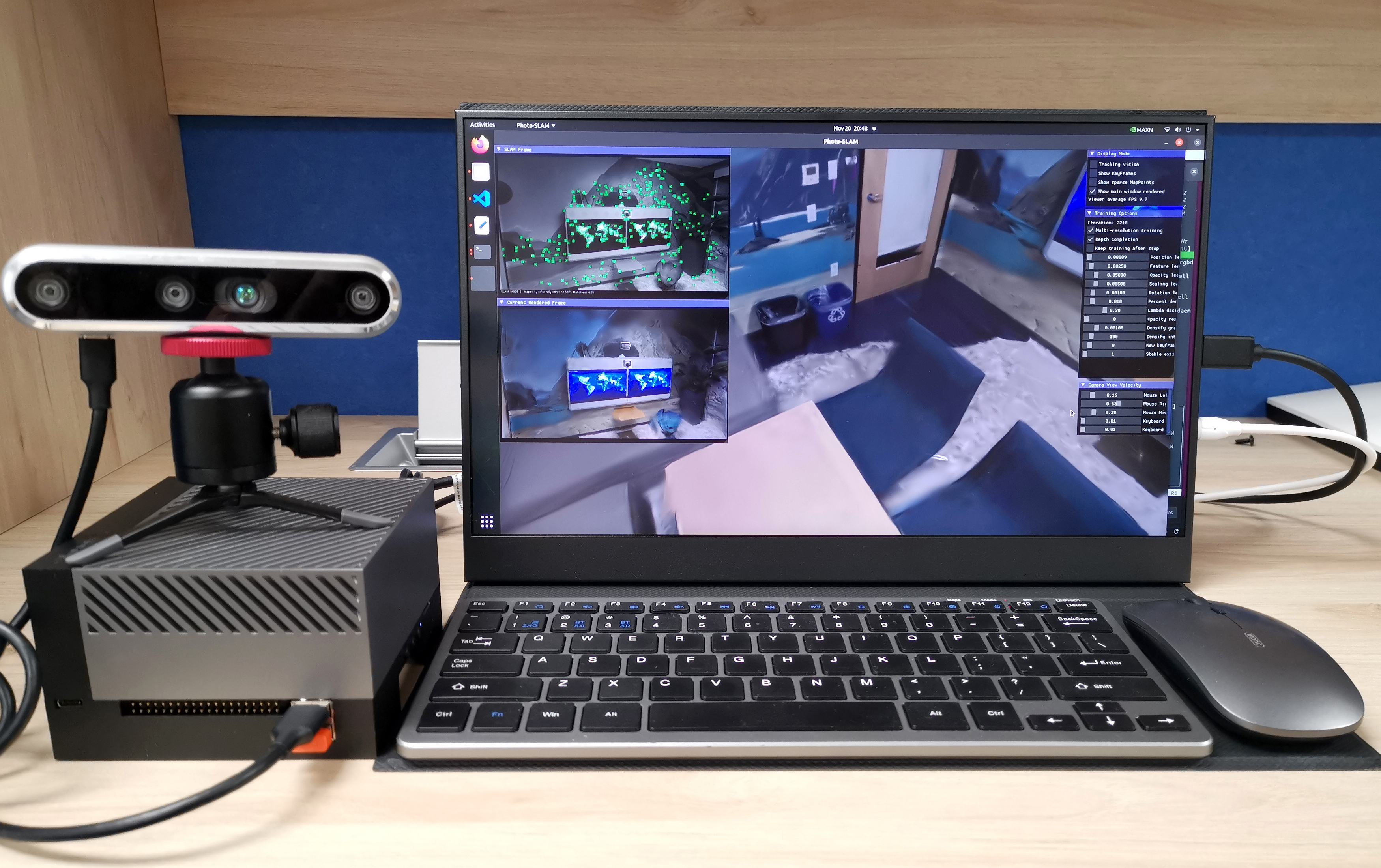}
    \caption{The proposed system has real-time performance on embedded platforms, such as Jetson AGX Orin Developer Kit.}
    \label{fig:jetson}
\end{figure}

Photo-SLAM is a novel system for simultaneous localization and photorealistic mapping, which can even run on embedded platforms at real-time speed, as demonstrated in \fref{fig:jetson}. In this supplementary, we provide additional results regarding localization and mapping performance. 

\begin{figure}
    \centering
    \includegraphics[width=0.9\linewidth]{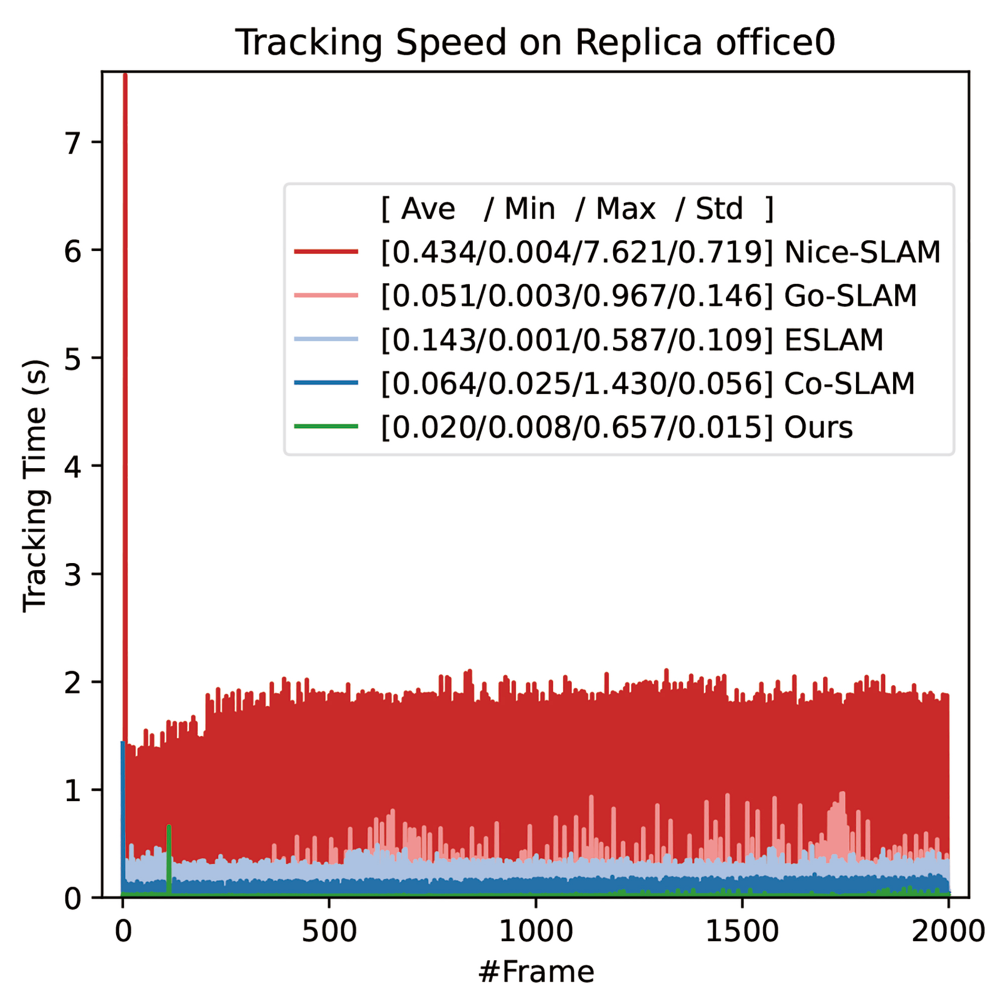}
    \vspace{-0.2cm}
    \caption{Tracking speed comparisons on scene office0 using an RGB-D camera. The vertical axis denotes the processing time of each frame while the horizontal axis denotes the frame number. {[Ave/Min/Max/Std]} represent the average, minimum, and maximum tracking time and its standard deviation respectively.}
    \label{fig:tracking}
\end{figure}

\section{Localication}
\label{sec:Localication}
\noindent\textbf{Stability}. As online systems, SLAMs are required to process the incoming frames and estimate current camera poses in time. Therefore, tracking stability regarding latency and the average processing time is an important factor in evaluating system performance in addition to pose estimation accuracy. As reported in Table 1 %\Tref{tab:replica} 
of the main paper, Photo-SLAM is capable of processing more than 40 frames per second with accurate pose estimation. The average tracking speed is about six times faster than ESLAM~\cite{eslam} and three times faster than Co-SLAM~\cite{coslam}. Here, we provide additional analysis on tracking stability while an example plotted in \fref{fig:tracking}.

Although the average tracking time of Go-SLAM~\cite{go-slam} is less than Co-SLAM and ESLAM, the processing latency is high due to frequently conducting expensive global optimization. As shown in \fref{fig:tracking}, Go-SLAM often takes about 1 second to process the frame and estimate the pose. Moreover, both Nice-SLAM~\cite{niceslam} and Co-SLAM need a longer time to accurately initialize the tracking. Obviously, our system can rapidly and stably process the incoming frames, having minimum average tracking time and standard deviation. The peak processing time of our system occurs when loop closure is detected for correcting pose estimation drift. 

\begin{figure}
    \centering
    \def\imgw{0.9}
    \subfloat[Replica office0]{\includegraphics[width=\imgw\linewidth]{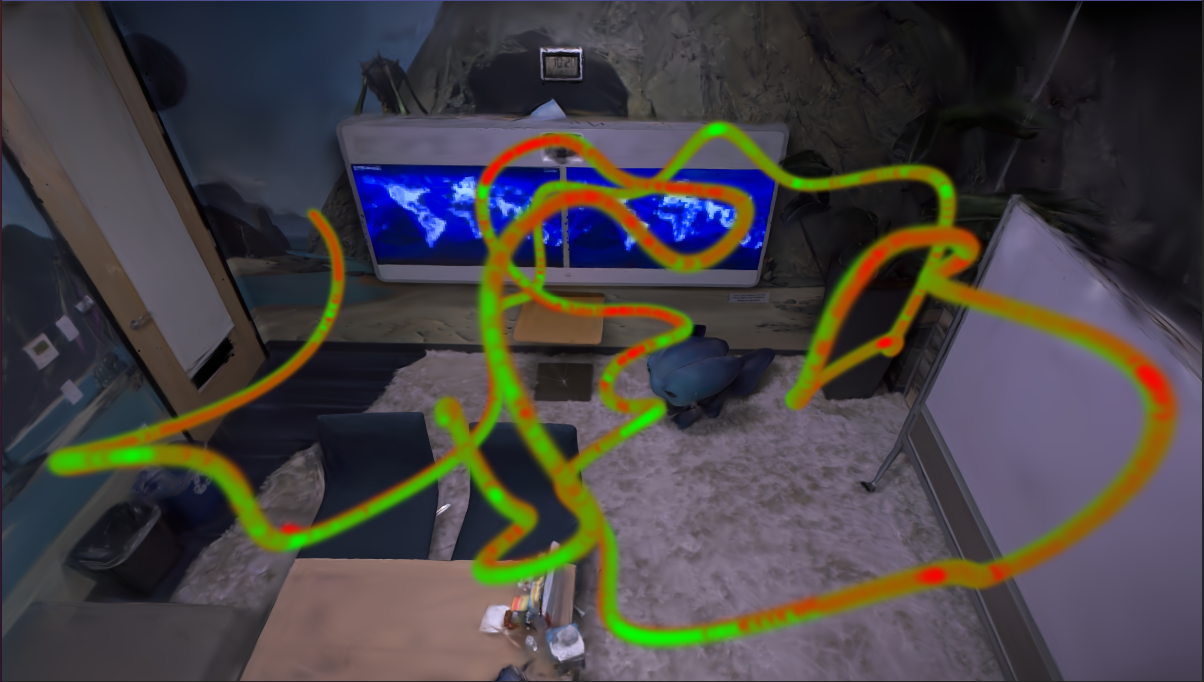}}\\
    %\subfloat[Replica office4]{\includegraphics[width=\imgw\linewidth]{figure/traj/office4.png}}\\
    \subfloat[Replica room0]{\includegraphics[width=\imgw\linewidth]{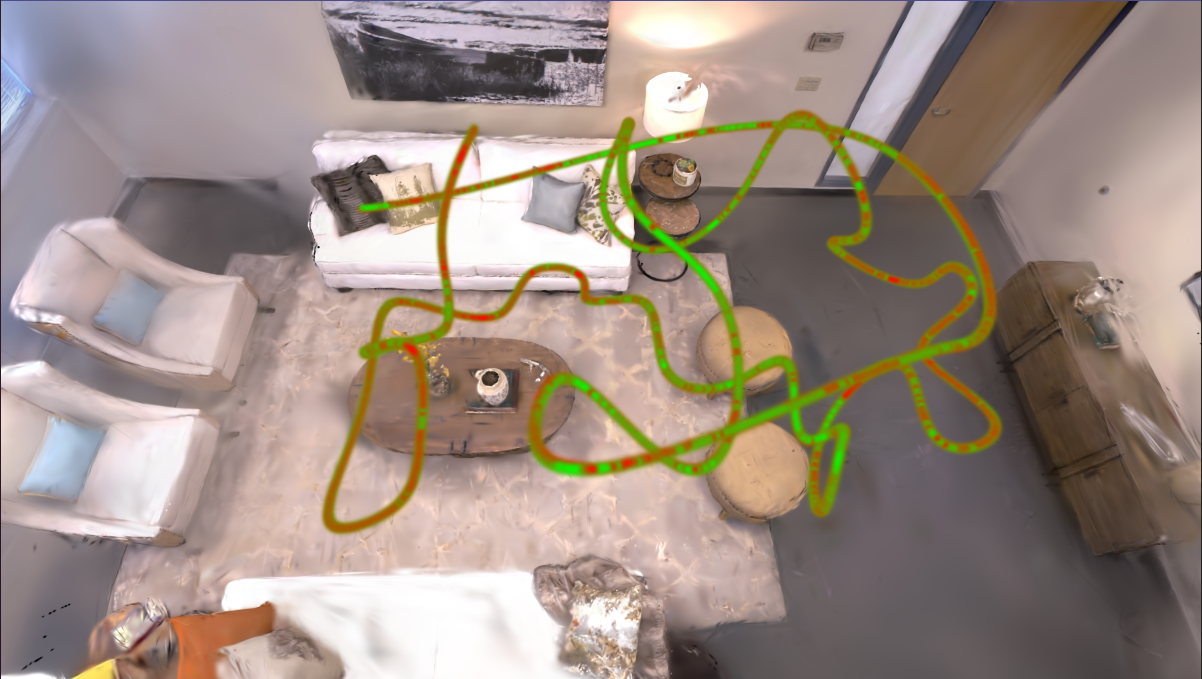}}\\
    \subfloat[TUM fr3-office]{\includegraphics[width=\imgw\linewidth]{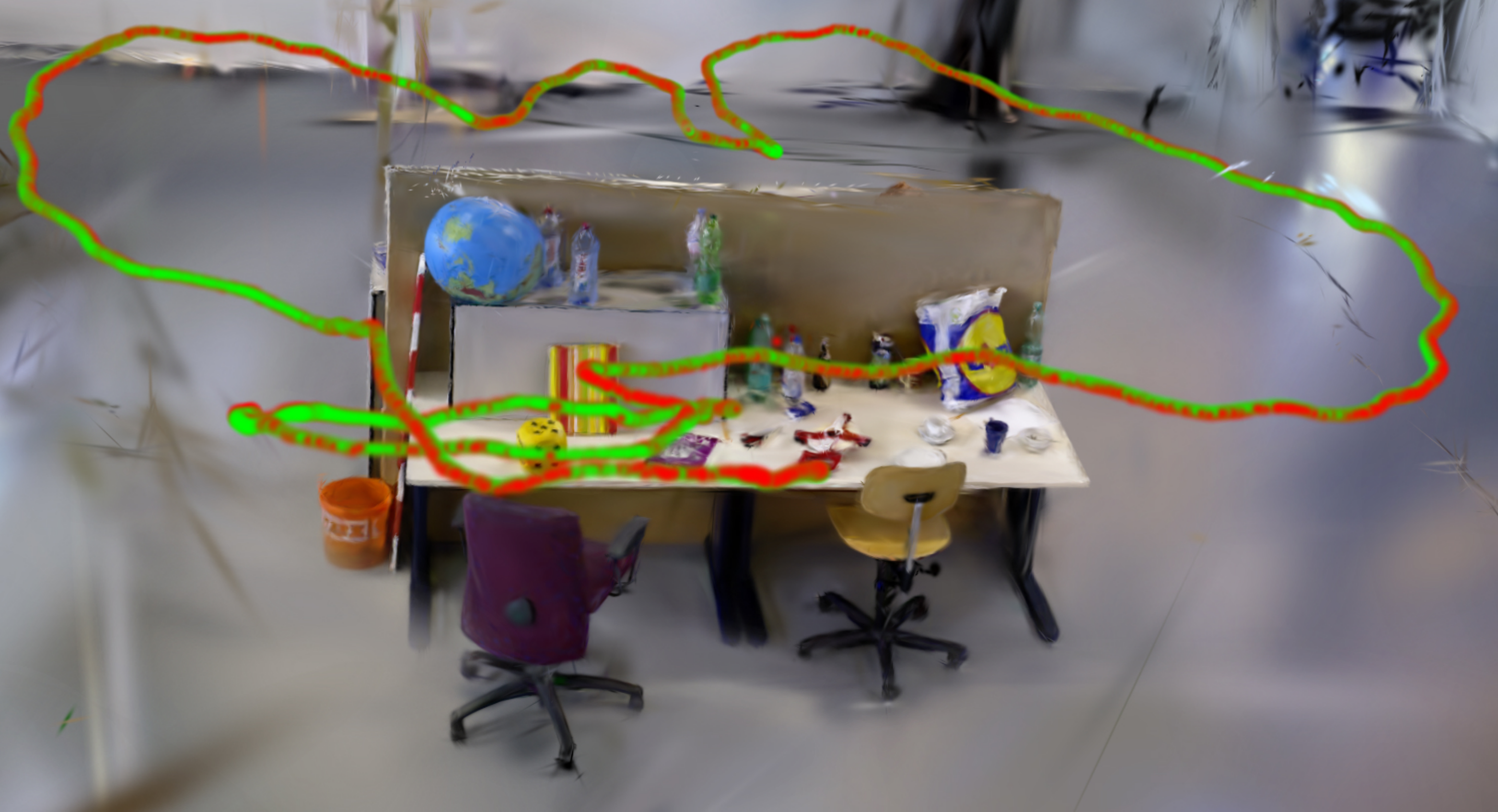}}
    %\subfloat[MH-02]{\includegraphics[width=\imgw\linewidth]{figure/traj/MH_02_easy.png}}
    \caption{Trajectory in the reconstructed map. Green points denote ground truth trajectory while red denotes the estimated trajectory of Photo-SLAM.}
    \label{fig:traj_map}
\end{figure}

\noindent\textbf{Accuracy}. Some qualitative tracking results of Photo-SLAM are demonstrated in \fref{fig:traj_map}.

\section{Discussion}
\label{sec:vsmapping}

\noindent\textbf{Online Mapping vs Offline Mapping}.
For online mapping, the mapping process occurs simultaneously with the localization process. Therefore it requires continuous and prompt updates with each new observation as the robot or camera moves and observes its surroundings. In general, online photorealistic mapping is more challenging than offline photorealistic mapping, since it is crucial to balance the trade-off between computational efficiency and rendering quality.
As mentioned in the main paper, we proposed a geometry-based densification strategy and a Gaussian-Pyramid-based (GP) learning method to achieve high-quality online mapping. To further support this statement, we compared the photorealistic mapping performance between our Photo-SLAM and 3D Gaussian splatting (3DG)~\cite{kerbl20233dgaussiansplatting}. 3DG is the SOTA offline method which takes a set of images with known poses and a sparse point cloud as input to learn a radiance field for view synthesis. During the experiments, 3DG used the keyframe poses estimated by Photo-SLAM and performed training for the same duration as Photo-SLAM. The required point cloud input is initialized in three different ways: 1) randomly initializing 100 points; 2) randomly initializing 10,000 points; and 3) initializing from the hyper primitives map of Photo-SLAM. The results are reported in \Tref{tab:3dg}. Without inputting fine-grained point clouds, 3DG needs more time for optimization such that the rendering quality decreases. In addition, to enhance rendering quality, 3DG tends to densify point clouds leading to larger model size and slower rendering speed. Whether using monocular cameras or RGB-D cameras, Photo-SLAM consistently delivers compelling rendering quality and faster rendering speeds, owing to the effectiveness of the proposed algorithms.

%This versatile performance makes it a valuable tool for various applications, such as augmented reality, virtual tours, and visual mapping, where both rendering quality and speed are crucial factors.
%Our system Photo-SLAM achieves competitive rendering quality while higher rendering speeds whether using monocular cameras or RGB-D cameras.

%and verify the effectiveness of the proposed algorithms
%In our system, online photorealistic mapping optimized the hyper primitives by minimizing the loss between the original images and rendering images.
%The system incrementally builds a map of the environment in real time as the robot or camera moves and observes its surroundings. 
%In contrast to offline mapping, online mapping 
%As mentioned in the main paper, while the introduction of a 3D Gaussian splatting (3DG)~\cite{kerbl20233dgaussiansplatting} renderer helps in reducing the costs associated with rendering to some extent, it does not inherently enable the generation of high-fidelity view reconstruction for online incremental mapping. 
%real-Time Radiance Field Rendering
\begin{table}
    \centering
    \footnotesize
    \tabcolsep=0.05cm 
    \begin{tabular}{c|ccccc}
    \toprule
    method & {PSNR $\uparrow$} &SSIM $\uparrow$ & LPIPS $\downarrow$ & \makecell{Rendering \\FPS} $\uparrow$ &\makecell{Model Size \\(MB)}\\
    \midrule
    1) 3DG &27.844 &0.861 &0.213 &745.480 &36.141\\
    %2) 3DG &29.051 &0.869 &0.199 &693.976 &43.026\\
    %3) 3DG &34.153 &0.927 &0.077 &576.166 &92.122\\
    2) 3DG &34.555 &0.942 &0.065 &483.904 &144.196\\
    3) 3DG &37.055 &0.962 &0.032 &448.109 &219.470\\
    Ours using Mono &33.302 &0.926 &0.078 &911.262 &31.419 \\
    Ours using RGB-D &34.958 &0.942 &0.059 &1084.017 &35.211\\
    %1) 3DG &28.167 &0.891 & 0.142 &728.976 &27.068\\
    %2) 3DG &32.592 &0.932 & 0.083 &612.783 &74.253\\
    %3) 3DG &34.123 &0.949 & 0.051 &538.971 &126.199 \\
    %Ours using Mono &31.622 &0.920 &0.086 &1131.96 &26.653\\
    %Ours using RGB-D &33.789 &0.938 &0.066 &1125.18 &25.226 \\
    \bottomrule
    \end{tabular}
    \vspace{-0.1cm}
    \caption{Comparison of mapping performance between 3D Gaussian splatting (3DG) and our system Photo-SLAM with different settings on the Replica dataset.}
    \label{tab:3dg}
\end{table}

%As we stated in the main paper, although the introduction of a 3D Gaussian splatting renderer can reduce view reconstruction costs, it does not enable the generation of high-fidelity rendering for online incremental mapping,

%The map is continuously updated and refined with each new observation, allowing the system to maintain an up-to-date representation of the environment. 

%although the introduction of a 3D Gaussian splatting renderer can reduce view reconstruction costs, it does not enable the generation of high-fidelity rendering for online incremental mapping,

%\noindent\textbf{On Larger-scale Scenes}

\begin{table}[t]
    \centering
    %\tiny
    \footnotesize
    \tabcolsep=0.1cm 
    \resizebox{0.85\linewidth}{!}{
    \begin{tabular}{ccc|ccc}
    \toprule
	\multicolumn{3}{c}{On TUM Dataset}& \multicolumn{3}{c}{Resources}\\\cmidrule(lr){1-3} \cmidrule(lr){4-6}
	Scene&Cam&Method& \makecell{Tracking \\FPS} $\uparrow$  & \makecell{Rendering \\FPS} $\uparrow$ &\makecell{Model Size \\(MB)}\\
	\midrule      
    \multirow{6}{*}{\rotatebox{90}{fr1-desk}}&\multirow{3}{*}{Mono}&\textbf{Ours (Jetson)}&28.267&340.507&4.610\\ 
    &&\textbf{Ours (Laptop)}&28.330&1105.062&7.421\\ 
    &&\textbf{Ours}&57.781&2016.690&10.027\\ 
    &\multirow{3}{*}{RGB-D}&\textbf{Ours (Jetson)}&27.970&380.622&5.743\\ 
    &&\textbf{Ours (Laptop)}&28.930&1061.040&8.432\\ 
    &&\textbf{Ours}&58.378&2083.896&9.963\\ 
    \midrule      
    \multirow{6}{*}{\rotatebox{90}{fr2-xyz}}&\multirow{3}{*}{Mono}&\textbf{Ours (Jetson)}&24.005&169.321&14.286\\ 
    &&\textbf{Ours (Laptop)}&24.922&619.554&16.102\\ 
    &&\textbf{Ours}&58.241&1405.797&20.380\\ 
    &\multirow{3}{*}{RGB-D}&\textbf{Ours (Jetson)}&21.032&274.718&6.319\\ 
    &&\textbf{Ours (Laptop)}&22.665&701.590&13.850\\ 
    &&\textbf{Ours}&52.904&1790.120&21.399\\ 
    \midrule      
    \multirow{6}{*}{\rotatebox{90}{fr3-office}}&\multirow{3}{*}{Mono}&\textbf{Ours (Jetson)}&36.700&291.398&10.669\\ 
    &&\textbf{Ours (Laptop)}&38.929&824.658&16.249\\ 
    &&\textbf{Ours}&81.575&1522.120&19.211\\ 
    &\multirow{3}{*}{RGB-D}&\textbf{Ours (Jetson)}&18.039&291.907&12.726\\ 
    &&\textbf{Ours (Laptop)}&19.636&764.342&15.349\\ 
    &&\textbf{Ours}&43.650&1540.757&17.009\\  
    \bottomrule
    \end{tabular}}
    \vspace{-0.2cm}
    \caption{Additional results of Photo-SLAM with different platforms on the TUM dataset.} 
    \label{tab:tum-detailed} %in terms of tracking and rendering speed and model size
\end{table}

\begin{table}[t]
    \centering
    %\tiny
    \footnotesize
    \tabcolsep=0.1cm 
    \resizebox{0.85\linewidth}{!}{
    \begin{tabular}{cc|ccc}
    \toprule
	\multicolumn{2}{c}{On EuRoC Dataset}& \multicolumn{3}{c}{Resources}\\\cmidrule(lr){1-2} \cmidrule(lr){3-5}
	Scene&Method& \makecell{Tracking \\FPS} $\uparrow$  & \makecell{Rendering \\FPS} $\uparrow$ &\makecell{Model Size \\(MB)}\\
	\midrule      
    \multirow{3}{*}{{MH-01}}&\textbf{Ours (Jetson)}&21.359&93.762&43.385\\ 
    &\textbf{Ours (Laptop)}&25.019&316.403&89.700\\ 
    &\textbf{Ours}&44.977&613.958&123.528\\ 
    \midrule      
    \multirow{3}{*}{{MH-02}}&\textbf{Ours (Jetson)}&22.355&101.021&36.263\\ 
    &\textbf{Ours (Laptop)}&26.189&332.174&81.569\\ 
    &\textbf{Ours}&46.556&675.508&113.116\\ 
    \midrule      
    \multirow{3}{*}{{V1-01}}&\textbf{Ours (Jetson)}&21.332&106.008&28.444\\ 
    &\textbf{Ours (Laptop)}&25.403&367.903&55.263\\ 
    &\textbf{Ours}&44.763&835.119&74.457\\ 
    \midrule      
    \multirow{3}{*}{{V2-01}}&\textbf{Ours (Jetson)}&23.872&99.988&27.840\\ 
    &\textbf{Ours (Laptop)}&27.556&307.025&62.588\\ 
    &\textbf{Ours}&48.911&595.234&82.600\\ 
    \bottomrule
    \end{tabular}}
    \vspace{-0.2cm}
    \caption{Additional results of Photo-SLAM with different platforms on the EuRoC MAV stereo dataset.} 
    \label{tab:euroc-detailed}
\end{table}

\begin{table*}
    \centering
    %\tiny
    \footnotesize
    \tabcolsep=0.1cm 
    \resizebox{0.85\textwidth}{!}{
    \begin{tabular}{ccc|cccccccc}
    \toprule
	\multicolumn{3}{c}{On Replica Dataset}& \multicolumn{2}{c}{Localization}& \multicolumn{3}{c}{Mapping}& \multicolumn{3}{c}{Resources}\\\cmidrule(lr){1-3} \cmidrule(lr){4-5} \cmidrule(lr){6-8} \cmidrule(lr){9-11}
	Scene&Cam&Method&\makecell{Trajectory \\(RMSE$_{cm}$)} $\downarrow$ &\makecell{Rotation \\(RMSE)} $\downarrow$& {PSNR $\uparrow$} &SSIM $\uparrow$ & LPIPS $\downarrow$ & \makecell{Tracking \\FPS} $\uparrow$  & \makecell{Rendering \\FPS} $\uparrow$ &\makecell{Model Size \\(MB)}\\
	\midrule      
    \multirow{6}{*}{office0}&\multirow{3}{*}{Mono}&\textbf{Ours (Jetson)}&0.467&0.00334&34.415&0.940&0.949&18.328&123.551&17.703\\ 
    &&\textbf{Ours (Laptop)}&0.587&0.00343&36.227&0.954&0.071&20.422&413.645&21.703\\ 
    &&\textbf{Ours}&0.575&0.00369&36.989&0.955&0.061&42.487&930.598&24.975\\ 
    &\multirow{3}{*}{RGB-D}&\textbf{Ours (Jetson)}&0.499&0.00356&35.447&0.949&0.086&19.076&154.087&18.281\\ 
    &&\textbf{Ours (Laptop)}&0.519&0.00317&38.219&0.965&0.053&22.446&497.917&18.826\\ 
    &&\textbf{Ours}&0.522&0.00307&38.477&0.964&0.050&48.588&1447.887&19.740\\ 
    \midrule      
    \multirow{6}{*}{office1}&\multirow{3}{*}{Mono}&\textbf{Ours (Jetson)}&5.586&0.31140&32.382&0.904&0.113&18.312&80.048&22.904\\ 
    &&\textbf{Ours (Laptop)}&0.379&0.00463&37.970&0.954&0.060&19.968&322.160&21.224\\ 
    &&\textbf{Ours}&0.315&0.00383&37.592&0.950&0.062&42.296&857.324&26.982\\ 
    &\multirow{3}{*}{RGB-D}&\textbf{Ours (Jetson)}&0.402&0.00517&37.510&0.953&0.065&19.194&118.584&20.656\\ 
    &&\textbf{Ours (Laptop)}&0.440&0.00543&39.109&0.962&0.049&22.349&496.644&19.548\\ 
    &&\textbf{Ours}&0.436&0.00477&39.089&0.961&0.047&47.333&1263.343&21.193\\ 
    \midrule      
    \multirow{6}{*}{office2}&\multirow{3}{*}{Mono}&\textbf{Ours (Jetson)}&1.402&0.01452&28.083&0.900&0.131&17.502&90.181&19.704\\ 
    &&\textbf{Ours (Laptop)}&2.087&0.02154&31.202&0.927&0.098&18.975&343.662&27.332\\ 
    &&\textbf{Ours}&5.031&0.04696&31.794&0.929&0.091&39.604&930.777&31.558\\ 
    &\multirow{3}{*}{RGB-D}&\textbf{Ours (Jetson)}&1.209&0.00964&29.755&0.919&0.110&17.860&124.420&27.560\\ 
    &&\textbf{Ours (Laptop)}&1.188&0.00972&32.720&0.940&0.080&21.507&425.452&31.711\\ 
    &&\textbf{Ours}&1.276&0.01094&33.034&0.938&0.077&44.062&904.249&34.065\\ 
    \midrule      
    \multirow{6}{*}{office3}&\multirow{3}{*}{Mono}&\textbf{Ours (Jetson)}&0.429&0.00232&28.058&0.886&0.132&17.881&96.872&15.505\\ 
    &&\textbf{Ours (Laptop)}&0.409&0.00239&32.012&0.924&0.090&19.518&368.530&20.475\\ 
    &&\textbf{Ours}&0.472&0.00227&31.622&0.920&0.086&40.870&1131.957&26.653\\ 
    &\multirow{3}{*}{RGB-D}&\textbf{Ours (Jetson)}&0.718&0.00222&30.954&0.917&0.103&17.889&120.118&20.270\\ 
    &&\textbf{Ours (Laptop)}&0.747&0.00233&33.594&0.939&0.072&20.051&388.624&23.617\\ 
    &&\textbf{Ours}&0.782&0.00233&33.789&0.938&0.066&40.603&1125.175&25.226\\ 
    \midrule      
    \multirow{6}{*}{office4}&\multirow{3}{*}{Mono}&\textbf{Ours (Jetson)}&0.579&0.00305&30.399&0.921&0.109&18.755&102.949&15.201\\ 
    &&\textbf{Ours (Laptop)}&0.616&0.00279&33.656&0.940&0.078&20.311&375.033&21.444\\ 
    &&\textbf{Ours}&0.583&0.00272&34.168&0.941&0.072&42.262&849.305&26.154\\ 
    &\multirow{3}{*}{RGB-D}&\textbf{Ours (Jetson)}&0.661&0.00367&32.219&0.931&0.091&17.107&92.237&32.405\\ 
    &&\textbf{Ours (Laptop)}&0.629&0.00446&35.534&0.951&0.059&19.361&333.874&32.270\\ 
    &&\textbf{Ours}&0.582&0.00423&36.020&0.952&0.054&39.870&1061.749&35.421\\ 
    \midrule      
    \multirow{6}{*}{room0}&\multirow{3}{*}{Mono}&\textbf{Ours (Jetson)}&0.369&0.00321&26.423&0.787&0.221&17.987&87.246&17.121\\ 
    &&\textbf{Ours (Laptop)}&0.349&0.00294&29.899&0.868&0.125&19.521&332.127&33.151\\ 
    &&\textbf{Ours}&0.345&0.00299&29.772&0.871&0.106&41.020&754.729&44.333\\ 
    &\multirow{3}{*}{RGB-D}&\textbf{Ours (Jetson)}&0.514&0.00265&27.867&0.833&0.165&17.424&104.248&31.196\\ 
    &&\textbf{Ours (Laptop)}&0.521&0.00257&31.288&0.914&0.075&19.119&322.585&52.266\\ 
    &&\textbf{Ours}&0.541&0.00270&30.716&0.899&0.075&39.825&897.870&55.397\\ 
    \midrule      
    \multirow{6}{*}{room1}&\multirow{3}{*}{Mono}&\textbf{Ours (Jetson)}&0.803&0.00670&27.076&0.841&0.177&19.834&99.038&19.699\\ 
    &&\textbf{Ours (Laptop)}&1.046&0.00868&30.459&0.902&0.092&21.580&333.430&32.959\\ 
    &&\textbf{Ours}&1.183&0.00772&31.302&0.910&0.083&44.316&782.326&43.865\\ 
    &\multirow{3}{*}{RGB-D}&\textbf{Ours (Jetson)}&0.381&0.00299&30.191&0.895&0.108&18.881&121.986&29.503\\ 
    &&\textbf{Ours (Laptop)}&0.399&0.00277&33.071&0.931&0.062&21.782&367.455&43.568\\ 
    &&\textbf{Ours}&0.394&0.00320&33.511&0.934&0.057&43.352&1018.111&49.617\\ 
    \midrule      
    \multirow{6}{*}{room2}&\multirow{3}{*}{Mono}&\textbf{Ours (Jetson)}&0.241&0.00280&27.432&0.889&0.138&17.918&80.568&16.303\\ 
    &&\textbf{Ours (Laptop)}&0.235&0.00263&32.970&0.935&0.075&19.499&339.442&22.790\\ 
    &&\textbf{Ours}&0.225&0.00258&33.181&0.934&0.067&40.313&1053.078&26.834\\ 
    &\multirow{3}{*}{RGB-D}&\textbf{Ours (Jetson)}&0.260&0.00257&31.883&0.928&0.078&15.982&95.480&33.592\\ 
    &&\textbf{Ours (Laptop)}&0.275&0.00250&35.295&0.954&0.045&18.158&336.103&38.307\\ 
    &&\textbf{Ours}&0.305&0.00257&35.028&0.951&0.043&36.244&953.755&41.032\\ 
    \bottomrule
    \end{tabular}}
    \vspace{-0.5em}
    \caption{Detailed results of Photo-SLAM with different platforms on the Replica dataset.} 
    \label{tab:replica-detailed}
\end{table*}

\section{More Results}
\begin{figure*}[t]
    \centering
    \def\imw{0.24}
     \subfloat[fr1-desk (Mono)]{
    \includegraphics[width=\imw\linewidth]{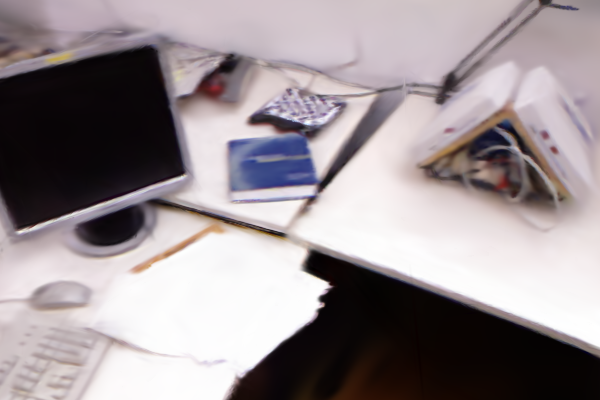}
    \includegraphics[width=\imw\linewidth]{figure/compare/tum_mono/1305031457.291656.png}
    \includegraphics[width=\imw\linewidth]{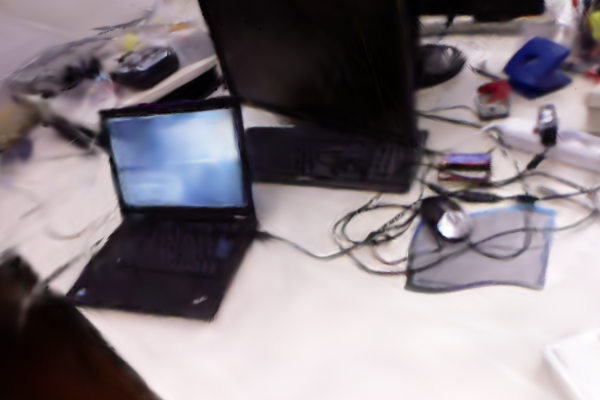}
    \includegraphics[width=\imw\linewidth]{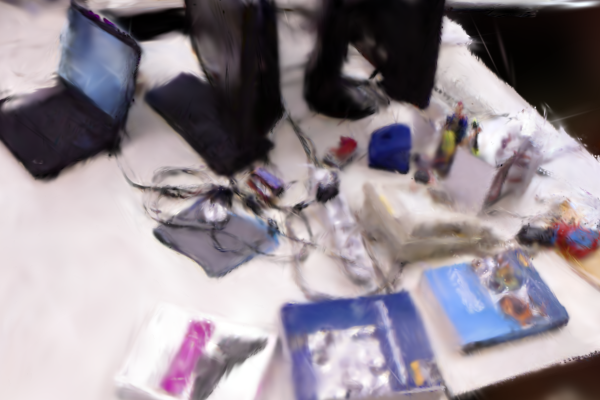}
    }\\
    \subfloat[fr2-xyz (Mono)]{
    \includegraphics[width=\imw\linewidth]{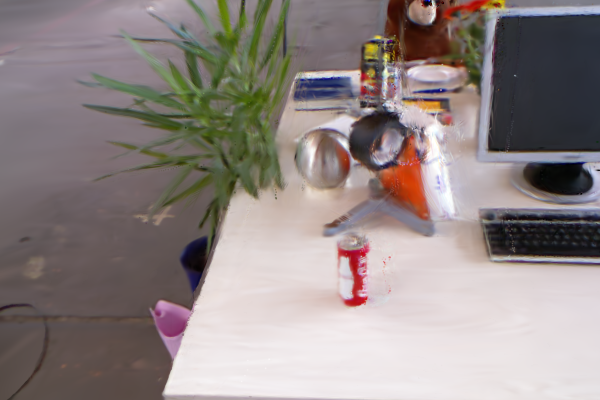}
    \includegraphics[width=\imw\linewidth]{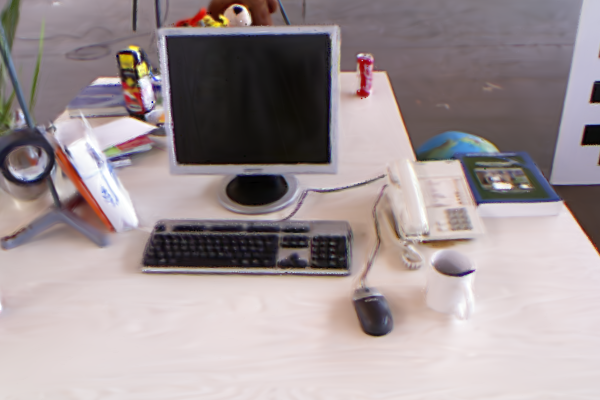}
    \includegraphics[width=\imw\linewidth]{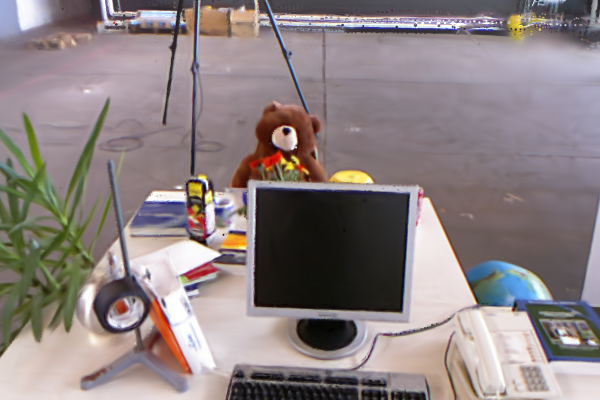}
    \includegraphics[width=\imw\linewidth]{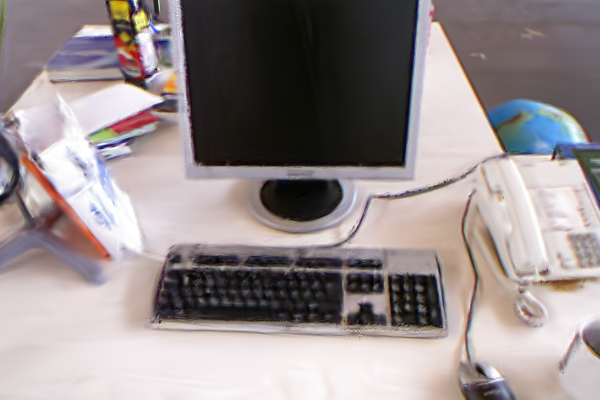}
    }\\
    \subfloat[fr3-office (Mono)]{
    \includegraphics[width=\imw\linewidth]{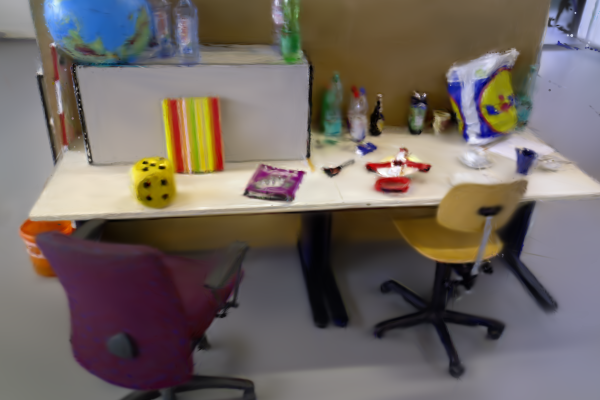}
    \includegraphics[width=\imw\linewidth]{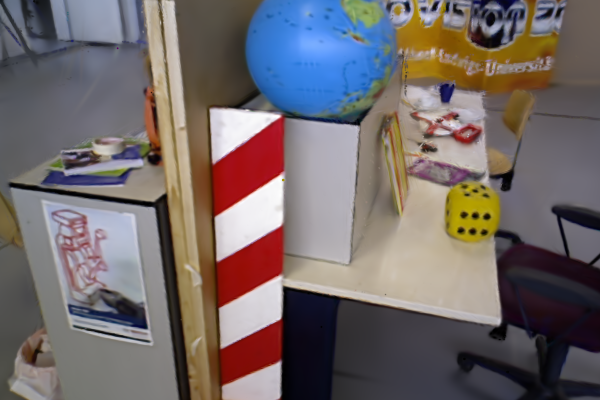}
    \includegraphics[width=\imw\linewidth]{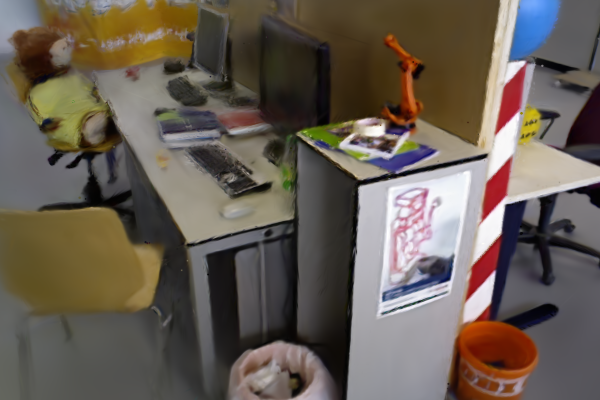}
    \includegraphics[width=\imw\linewidth]{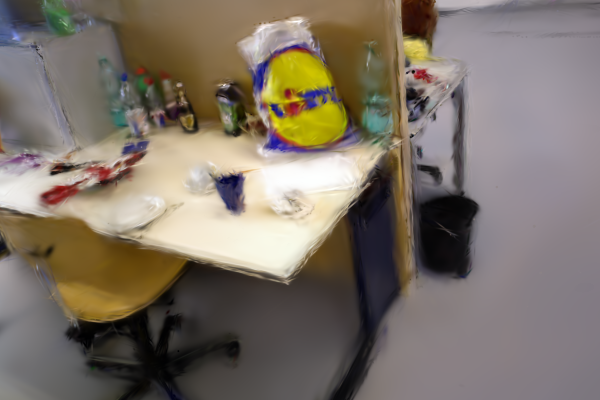}
    }
    %\vskip -0.1cm
    \caption{Qualitative results of Photo-SLAM on TUM using monocular cameras.}
    \label{fig:tum-mono}
    %\vskip -0.1cm
\end{figure*}

\begin{figure*}[t]
    \centering
    \def\imw{0.24}
     \subfloat[fr1-desk (RGB-D)]{
    \includegraphics[width=\imw\linewidth]{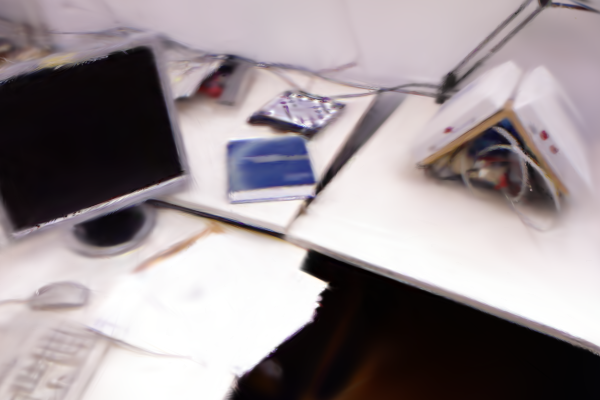}
    \includegraphics[width=\imw\linewidth]{figure/compare/tum_rgbd/1305031457.291656.png}
    \includegraphics[width=\imw\linewidth]{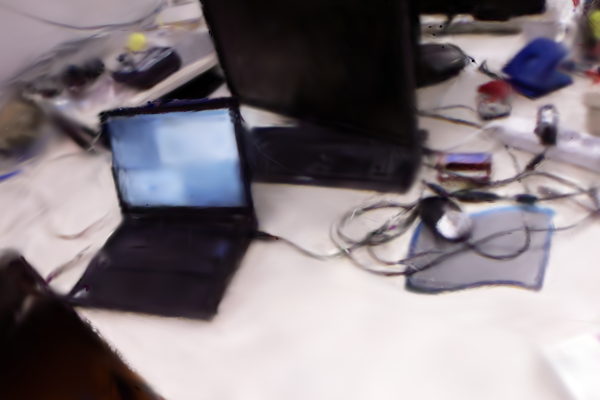}
    \includegraphics[width=\imw\linewidth]{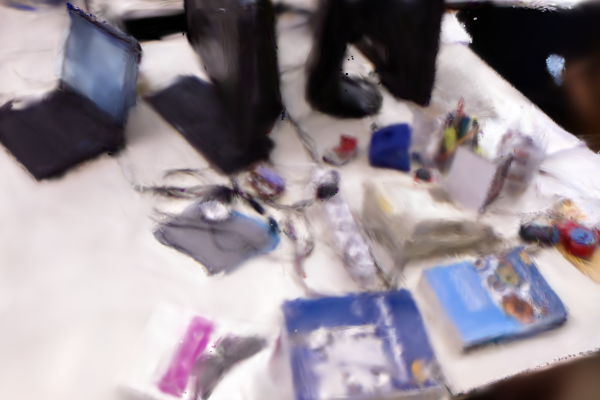}
    }\\
    \subfloat[fr2-xyz (RGB-D)]{
    \includegraphics[width=\imw\linewidth]{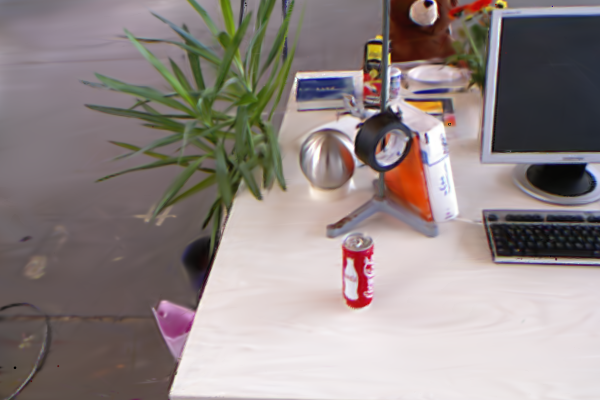}
    \includegraphics[width=\imw\linewidth]{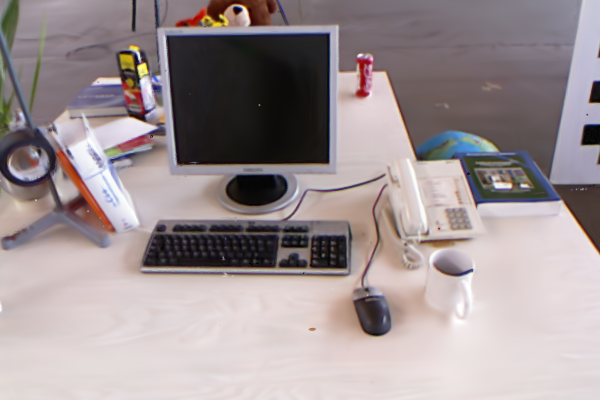}
    \includegraphics[width=\imw\linewidth]{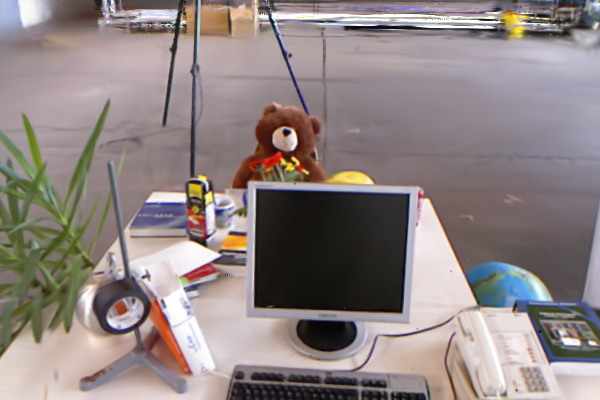}
    \includegraphics[width=\imw\linewidth]{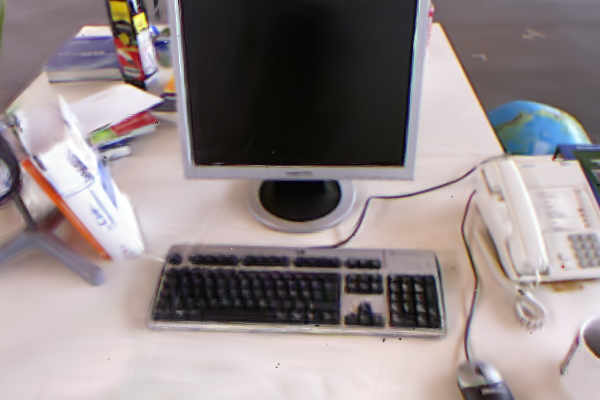}
    }\\
    \subfloat[fr3-office (RGB-D)]{
    \includegraphics[width=\imw\linewidth]{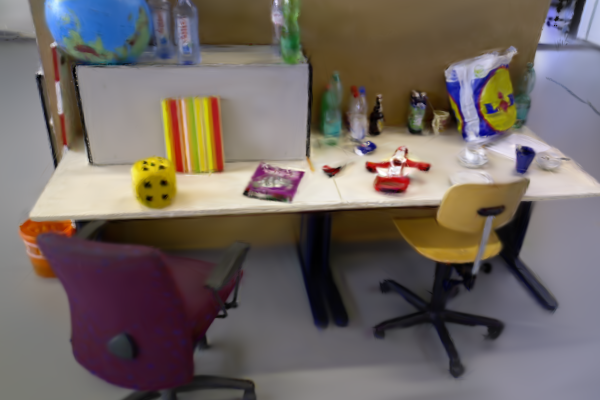}
    \includegraphics[width=\imw\linewidth]{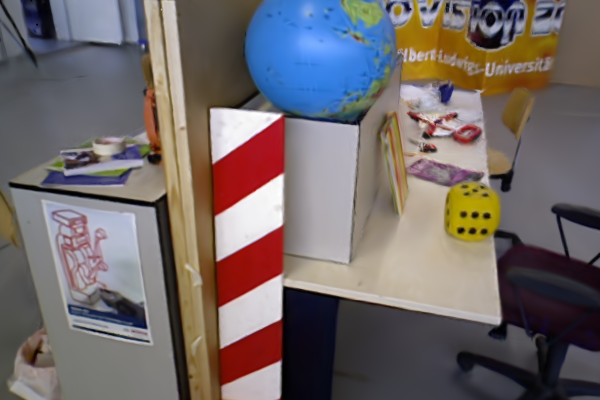}
    \includegraphics[width=\imw\linewidth]{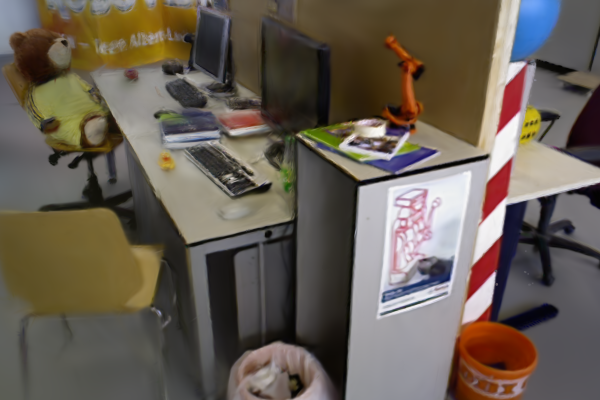}
    \includegraphics[width=\imw\linewidth]{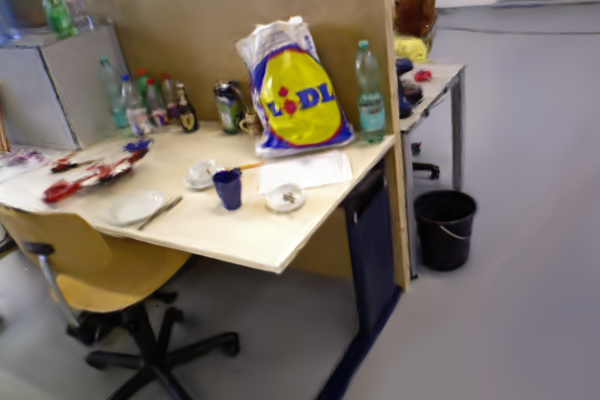}
    }
    %\vskip -0.1cm
    \caption{Qualitative results of Photo-SLAM on TUM using RGB-D cameras.}
    \label{fig:tum-rgbd}
    %\vskip -0.1cm
\end{figure*}

\begin{figure*}[t]
    \centering
    \def\imw{0.31}
     \subfloat[EuRoC MH-01]{
    \includegraphics[width=\imw\linewidth]{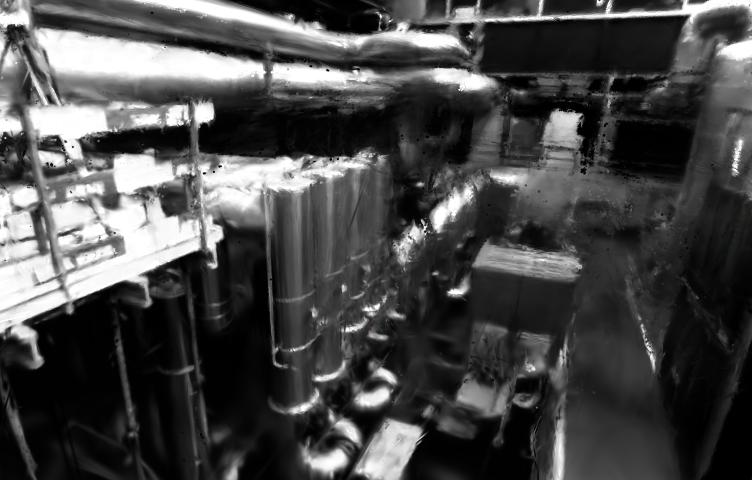}
    \includegraphics[width=\imw\linewidth]{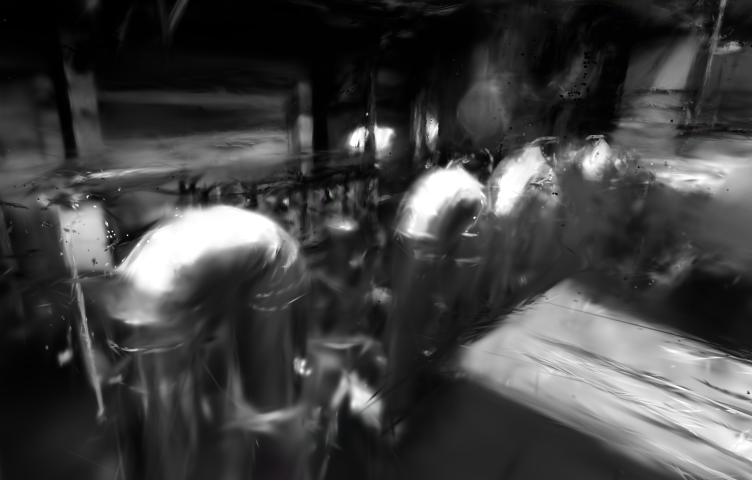}
    \includegraphics[width=\imw\linewidth]{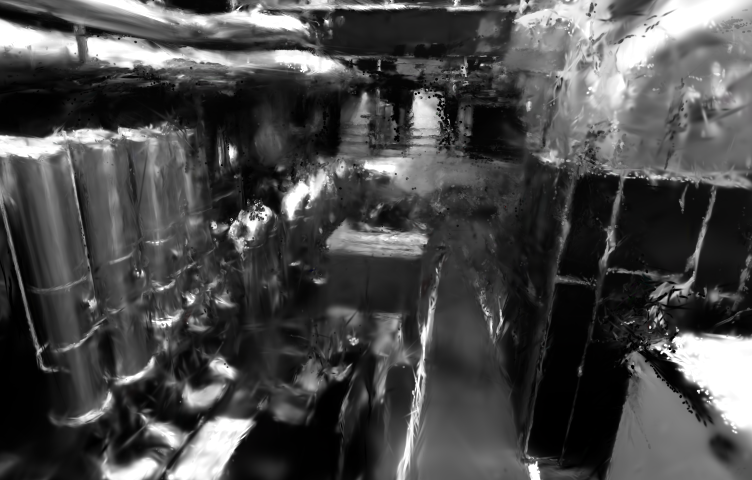}
    }\\
    \subfloat[EuRoC MH-02]{
    \includegraphics[width=\imw\linewidth]{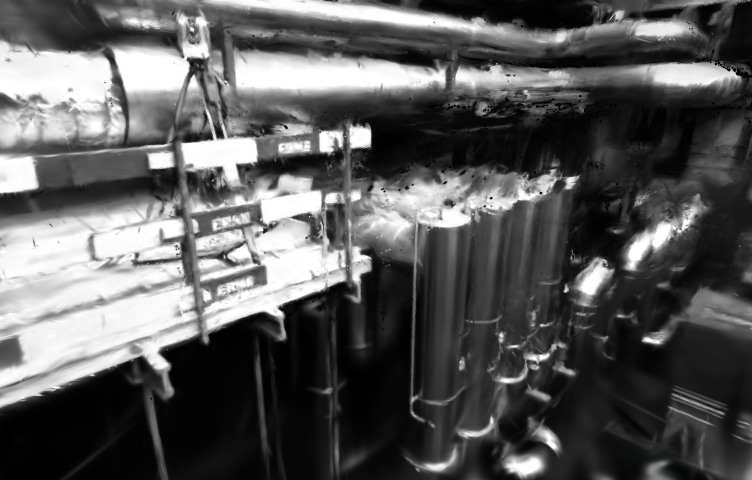}
    \includegraphics[width=\imw\linewidth]{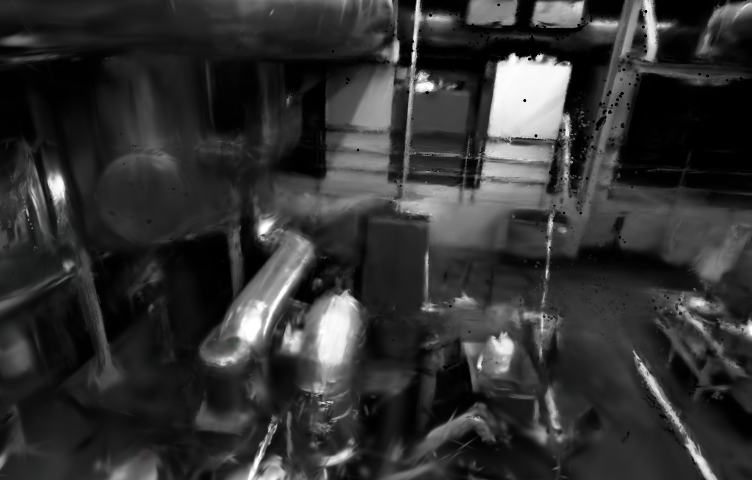}
    \includegraphics[width=\imw\linewidth]{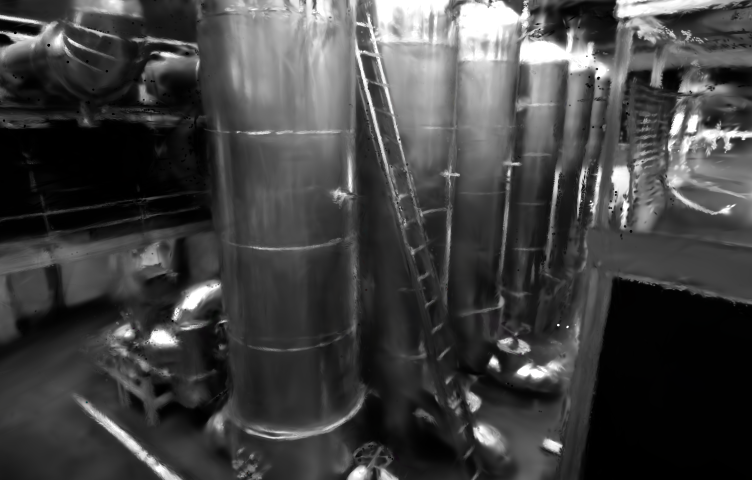}
    }\\
    \subfloat[EuRoC V1-01]{
    \includegraphics[width=\imw\linewidth]{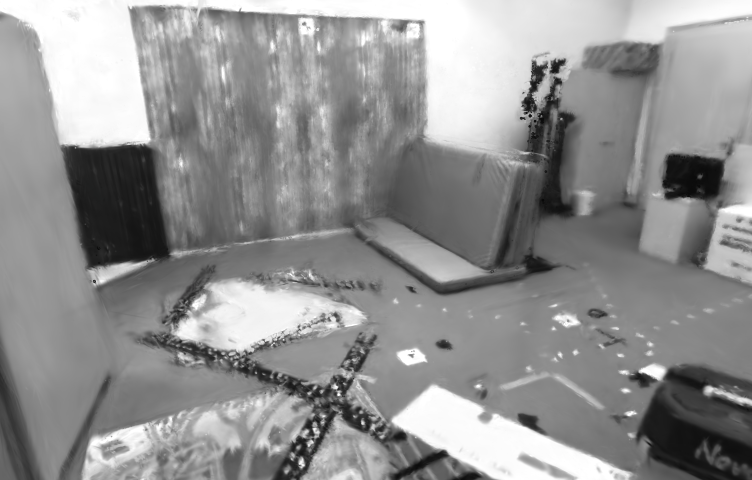}
    \includegraphics[width=\imw\linewidth]{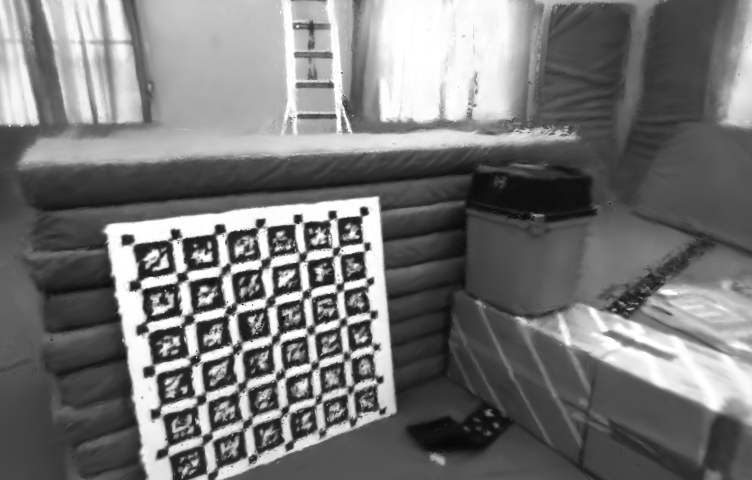}
    \includegraphics[width=\imw\linewidth]{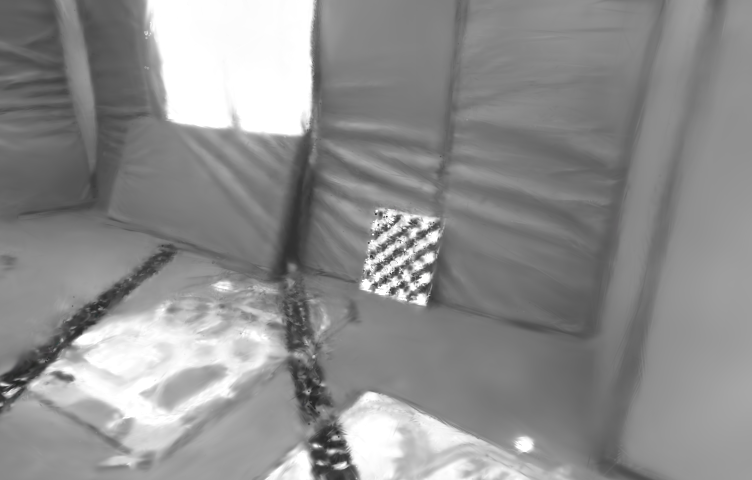}
    }\\
    \subfloat[EuRoC V2-01]{
    \includegraphics[width=\imw\linewidth]{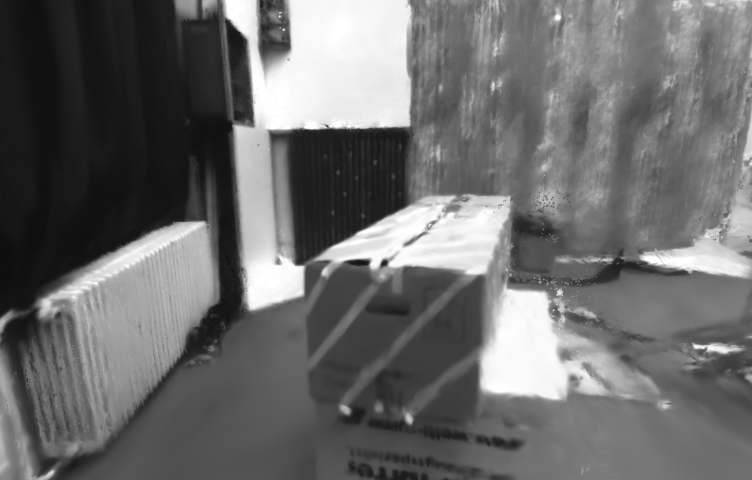}
    \includegraphics[width=\imw\linewidth]{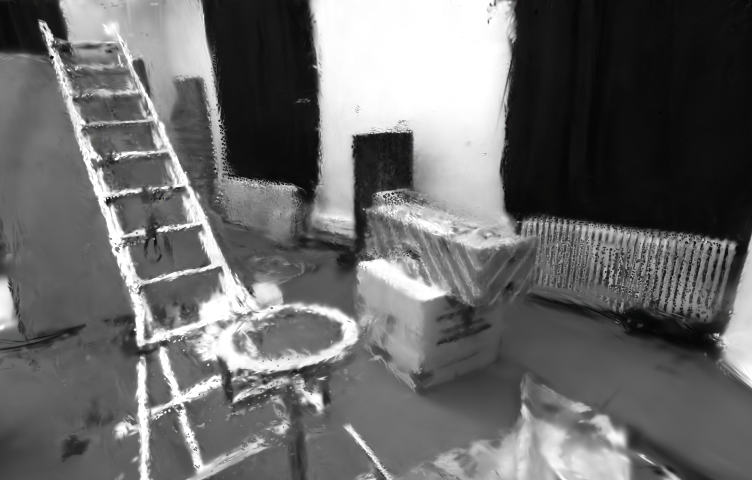}
    \includegraphics[width=\imw\linewidth]{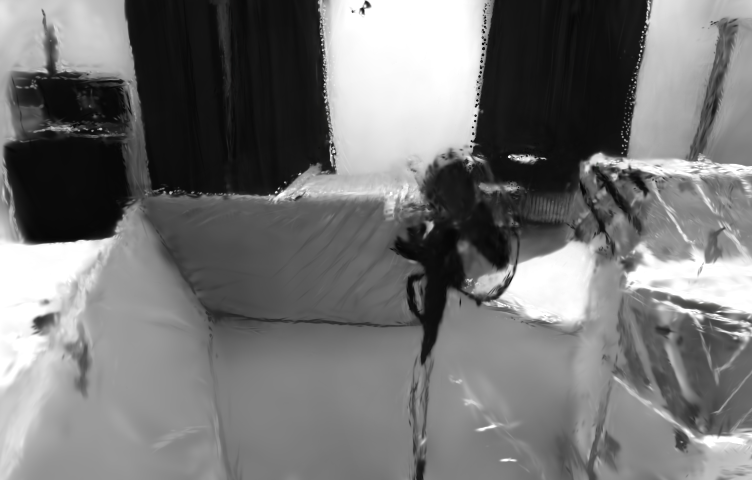}
    }\\
    %\vskip -0.1cm
    \caption{Qualitative results of Photo-SLAM on stereo EuRoC.}
    \label{fig:euroc}
    %\vskip -0.1cm
\end{figure*}

The results of each scene of the replica dataset are detailed in \Tref{tab:replica-detailed}
Additional qualitative results on the TUM dataset are demonstrated on \fref{fig:tum-mono} and \fref{fig:tum-rgbd}, while \fref{fig:euroc} illustrates qualitative results of Photo-SLAM on EuReC Stereo Dataset. %The live demos are in the supplementary video.

%\section{Qualitative Results on EuReC Stereo Dataset}

\end{document}